%% file: main.tex
\documentclass[]{youtu} 

\usepackage{mathpazo}
\usepackage{graphicx}
\usepackage{natbib}
\usepackage{multirow}
\usepackage{booktabs}
\usepackage{CJKutf8}
\usepackage[table]{xcolor}
\usepackage{subcaption}
\usepackage{graphicx}
\usepackage{soul}

\title{Youtu-VL: Unleashing Visual Potential via Unified Vision-Language Supervision}
\author{Youtu-VL Team$^*$}
\sourcecode{https://github.com/TencentCloudADP/youtu-vl}
\model{https://huggingface.co/collections/tencent/youtu}

\begin{document}

\abstract{
Despite the significant advancements represented by Vision-Language Models (VLMs), current architectures often exhibit limitations in retaining fine-grained visual information, leading to coarse-grained multimodal comprehension. We attribute this deficiency to a suboptimal training paradigm inherent in prevailing VLMs, which exhibits a text-dominant optimization bias by conceptualizing visual signals merely as passive conditional inputs rather than supervisory targets. To mitigate this, we introduce \textbf{Youtu-VL}, a framework leveraging the \textit{Vision-Language Unified Autoregressive Supervision (VLUAS)} paradigm, which fundamentally shifts the optimization objective from ``vision-as-input'' to ``vision-as-target.'' By integrating visual tokens directly into the prediction stream, Youtu-VL applies unified autoregressive supervision to both visual details and linguistic content. Furthermore, we extend this paradigm to encompass vision-centric tasks, enabling a standard VLM to perform vision-centric tasks without task-specific additions. Extensive empirical evaluations demonstrate that Youtu-VL achieves competitive performance on both general multimodal tasks and vision-centric tasks, establishing a robust foundation for the development of comprehensive generalist visual agents.
}
\maketitle

\renewcommand{\thefootnote}{*}
\footnotetext{Full author list in contributions.}
\renewcommand{\thefootnote}{\arabic{footnote}}

\vspace{-.1em}


\begin{figure*}[htbp]
  \centering
  \includegraphics[width=0.98\linewidth]{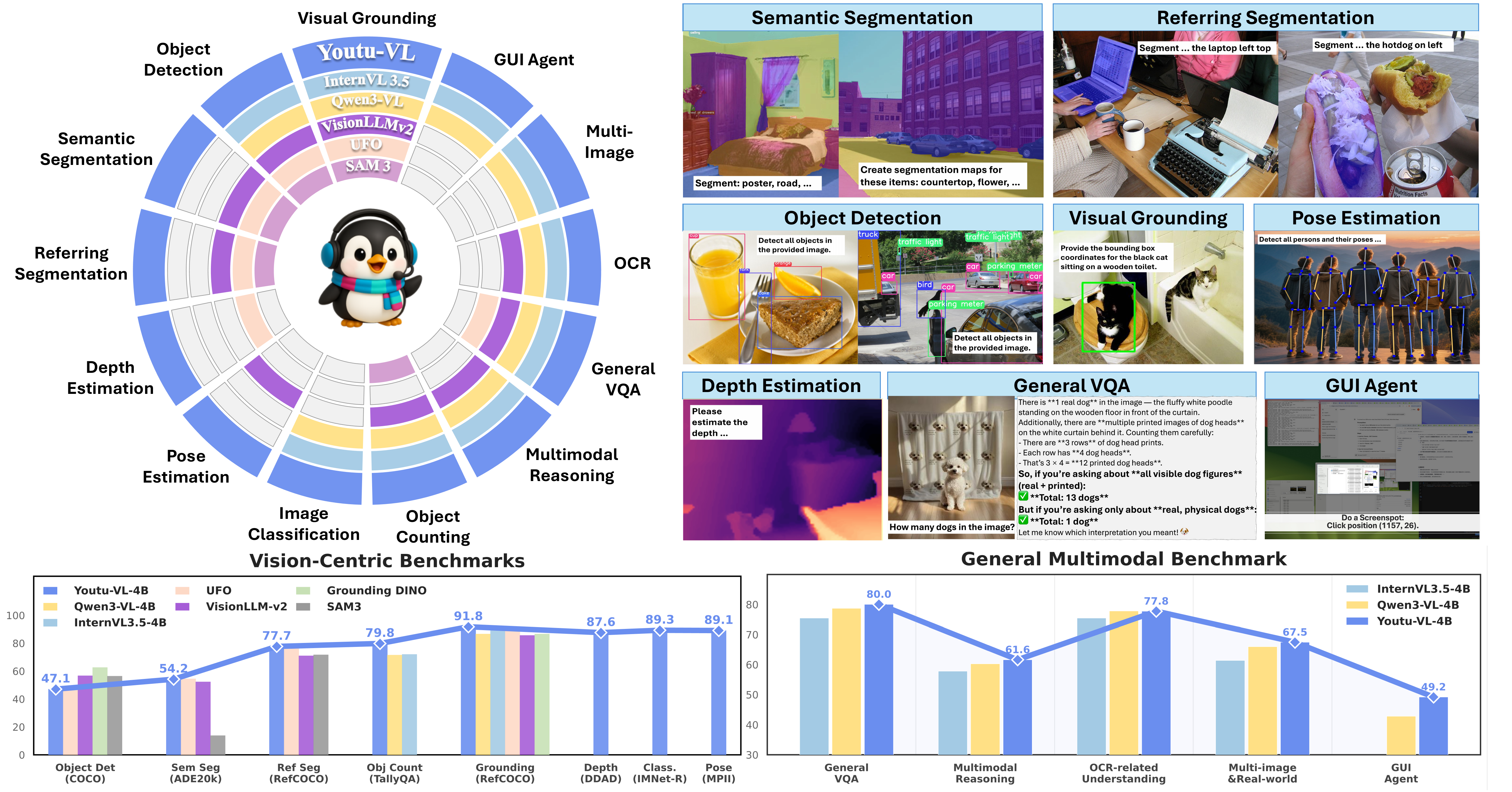}
  \caption{\textbf{Youtu-VL achieves competitive performance on both general multimodal tasks and vision-centric tasks.} The concentric rings illustrate the capability scope of different models across various tasks. Colored regions indicate that the model supports the corresponding task, while white regions denote a lack of support. Unlike prior models that exhibit functional gaps, Youtu-VL accommodates a comprehensive range of vision-centric and multimodal tasks via a standard architecture, achieving competitive performance without relying on task-specific modules.}
  \label{fig:abstract}
\end{figure*}

\input{sec/1_introduction}
\input{sec/2_architecture}
\input{sec/3_pre-training}
\input{sec/4_post-training}
\input{sec/5_evaluation}
\input{sec/6_conclusion}
\input{sec/7_contributions}

\setcitestyle{numbers,square}
\setcitestyle{square,numbers,comma}
\bibliography{youtu_bib}

\newpage
\appendix
\input{sec/b_evaluation_details}

\input{sec/c_extended_expriments}

\input{sec/a_qualitative_examples}

\end{document}

%% file: sec/1_introduction.tex
\section{Introduction}

\begin{figure*}[htbp]
  \centering
  \includegraphics[width=0.7\linewidth]{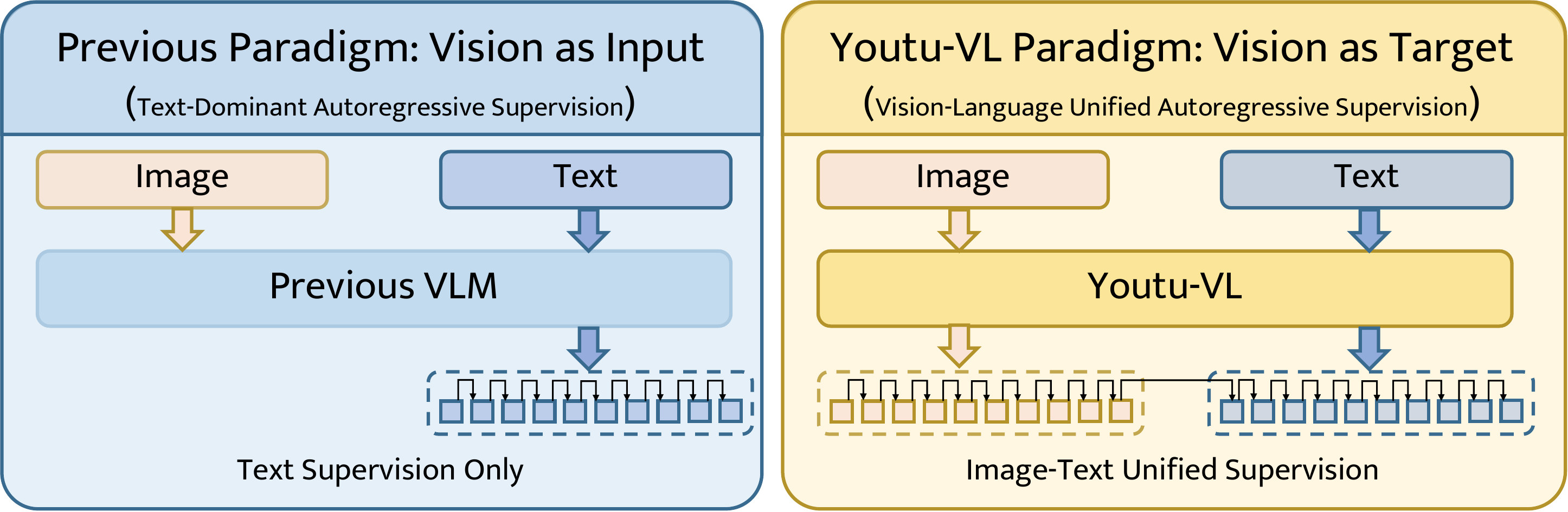}
  \caption{\textbf{Comparison between the previous "vision as input" paradigm and the Youtu-VL "vision as target" paradigm.} The left panel shows the previous text-dominant VLM, which relies solely on text supervision. The right panel illustrates the Youtu-VL paradigm, which incorporates Vision-Language Unified Autoregressive Supervision (VLUAS), treating vision as a target to achieve unified supervision for both image and text.}
  \label{fig:intro_compare}
\end{figure*}

Vision-Language Models (VLMs) have achieved significant proficiency in multimodal tasks by integrating the understanding capabilities of Large Language Models (LLMs) with pre-trained visual encoders. By aligning visual features to linguistic semantics, these architectures have become the standard for generalist applications ranging from image captioning to visual reasoning~\citep{Qwen3-VL, wang2025internvl35advancingopensourcemultimodal}.

However, a fundamental limitation persists in current VLM research: the retention of fine-grained visual information. We attribute this deficiency to a text-dominant optimization bias inherent in prevailing training paradigms. In typical architectures, visual signals are conceptualized merely as passive conditional inputs, while the optimization process is driven solely by autoregressive text generation objectives~\citep{llava, assran2023selfsupervisedlearningimagesjointembedding}. Consequently, the model is implicitly encouraged to discard visual details deemed redundant for coarse-grained text generation, inevitably creating an information bottleneck that hinders dense perception capabilities.

To address the aforementioned optimization bottleneck, we introduce \textbf{Youtu-VL}, a framework built upon the \textit{Vision-Language Unified Autoregressive Supervision (VLUAS)} paradigm. As illustrated in Figure~\ref{fig:intro_compare}, diverging from the prevailing ``vision-as-input'' strategies, we expand the textual lexicon into a unified multimodal vocabulary $\mathcal{V}_{\text{unified}}$ via a learned visual codebook. Central to this design is our Synergistic Vision Tokenizer, which fuses high-level semantic concepts with low-level geometric structures to produce discrete codes, thereby providing dense semantic visual supervision. This enables the model to treat visual signals as supervisory targets rather than passive conditions. By integrating these visual tokens into the prediction stream, Youtu-VL enforces the unified autoregressive supervision of visual details and linguistic content, ensuring the preservation of fine-grained information typically discarded by standard text-generation objectives.

Furthermore, we extend this unified paradigm to encompass a comprehensive suite of vision-centric tasks employing a standard architecture without task-specific modules. We categorize these capabilities into two distinct streams: text-based prediction and dense prediction. For text-based prediction tasks such as object detection and visual grounding, we implement an axis-specific vocabulary with absolute pixel coordinates, enabling the model to generate precise bounding boxes directly as textual tokens without normalization artifacts. Conversely, targeting pixel-level tasks like semantic segmentation and depth estimation, we utilize the model's native logit representations. Through multi-task autoregressive vision supervision, our approach enables high-quality dense prediction directly from these raw logits without extra task-specific heads or embeddings. This design unifies the inference pipeline, allowing a standard VLM to seamlessly transition between high-level reasoning and low-level dense perception without structural additions.

Our contributions are summarized as follows:

\begin{itemize}
    \item \textbf{Vision-Language Unified Autoregressive Supervision (VLUAS).} We propose a paradigm shift from text-only supervision to generative unification. By treating visual tokens as optimization targets, we mitigate text-dominant bias and enforce the retention of fine-grained visual details.
    
    \item \textbf{Vision-Centric Predictions from the Standard Architecture.} We treat image and text tokens with equivalent autoregressive status, empowering Youtu-VL to perform vision-centric tasks for both dense vision prediction (e.g., segmentation, depth) and text-based prediction (e.g., grounding, detection) within a standard VLM architecture, eliminating the need for task-specific additions.
    
    \item \textbf{Empirical Performance.} Extensive evaluations demonstrate that Youtu-VL achieves competitive performance on both general multimodal tasks and vision-centric tasks, establishing a robust baseline for generalist visual agents.
\end{itemize}

%% file: sec/2_architecture.tex
\section{Architecture and Methodology}

\begin{figure*}[t!]
  \centering
  \includegraphics[width=1\linewidth]{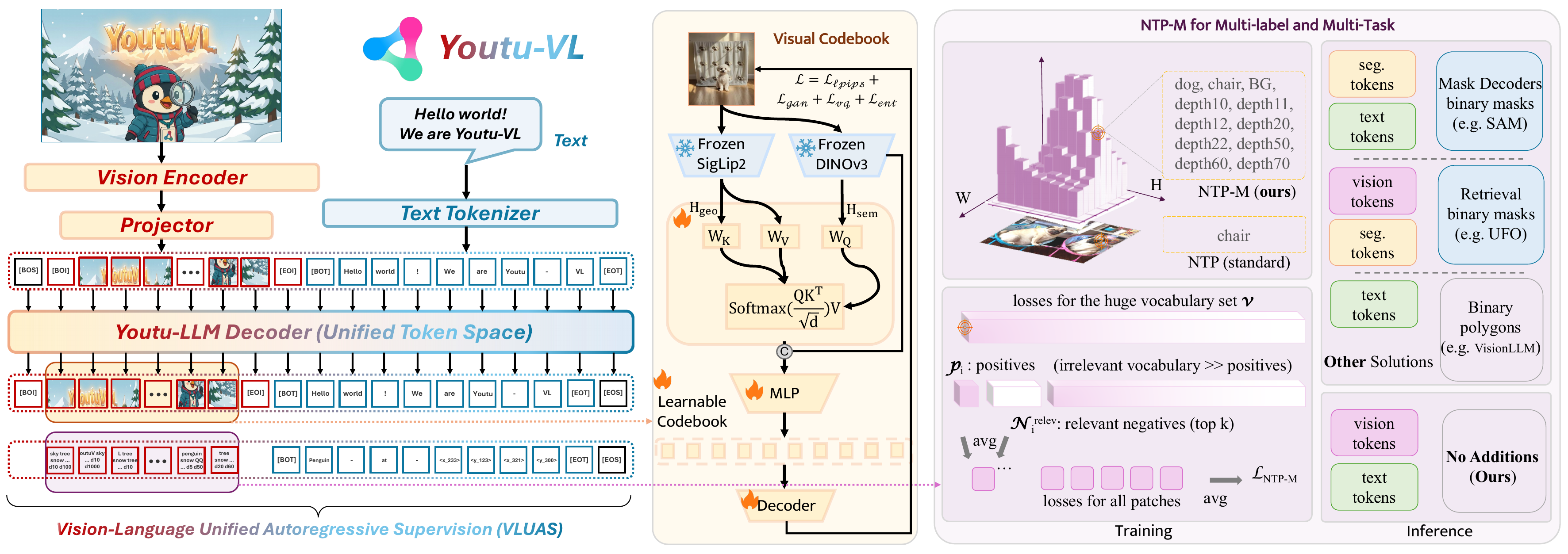}
  \caption{\textbf{Overview of the Youtu-VL Framework.}
Left: The architecture integrates a Vision Encoder and Youtu-LLM via a Spatial Merge Projector, operating under the proposed VLUAS paradigm for unified autoregressive modeling.
Middle: The Synergistic Vision Tokenizer. We construct a unified vocabulary by fusing semantic and geometric features via cross-attention, optimized with perception and adversarial losses.
Right: Dense prediction mechanism. Our proposed NTP-M enables robust multi-label supervision with a relevant negative sampling. Unlike conventional approaches, Youtu-VL achieves direct dense prediction without auxiliary decoders or task-specific tokens.}
  \label{fig:YoutuVL_pipeline}
\end{figure*}

We introduce Youtu-VL, a novel framework built on our proposed Vision-Language Unified Autoregressive Supervision (VLUAS) and vision-centric predictions from the standard architecture. As illustrated in the left panel of Figure~\ref{fig:YoutuVL_pipeline}, Youtu-VL comprises three core components: a vision encoder, a vision-language projector designed to map visual features into a unified vision-language token space, and a LLM.

\textbf{Large Language Model (LLM).} The LLM employed in Youtu-VL is a self-developed model built upon the architecture of Youtu-LLM~\citep{lu2026youtullmunlockingnativeagentic}. Building upon this foundation, we introduce the VLUAS paradigm, which extends the original text vocabulary with a dedicated image-token vocabulary. To achieve this, we employ a multi-stage training pipeline. In Stage~3, our Vision Tokenizer is trained to construct the visual codebook, forming the visual portion of $\mathcal{V}_{\text{unified}}$ (Figure~\ref{fig:YoutuVL_pipeline}, Middle). In Stage~4, we apply the proposed multi-label NTP objective directly to the LLM’s autoregressive outputs over vision tokens, enabling dense prediction supervision without introducing auxiliary decoders or task-specific tokens (Figure~\ref{fig:YoutuVL_pipeline}, Right).

\textbf{Vision Encoder.} Our vision encoder is built upon SigLIP-2~\citep{tschannen2025siglip2multilingualvisionlanguage}. Specifically, we adopt the \texttt{siglip2-so400m-patch16-naflex} variant. This architecture incorporates 2D Rotary Position Embedding (RoPE)~\citep{su2023roformerenhancedtransformerrotary} according to spatial shapes for positional encoding. It further employs window attention for efficiency, with global attention inserted every 8 layers. In addition, it leverages FlashAttention~\citep{dao2022flashattentionfastmemoryefficientexact} through the cumulative sequence length mechanism to handle variable-length sequences within a batch. 
Importantly, our design allows the encoder to process images at \textit{native resolution}, which provides flexibility for high-resolution inputs without requiring resizing to a fixed scale.

\textbf{Vision-Language Projector.} The Vision-Language Projector employs a Spatial Merge operation to concatenate adjacent $2 \times 2$ patch features~\citep{bai2023qwenvlversatilevisionlanguagemodel}. This effectively reduces the token count to one-quarter ($1/4$) of its original size, thereby shortening the input sequence length for the subsequent LLM. Finally, a two-layer Multi-Layer Perceptron (MLP) projects these compressed features into the LLM's input space.

\subsection{Vision-Language Unified Autoregressive Supervision Paradigm}
\label{sec:VL}
The VLUAS paradigm bridges the modality gap by extending the conventional textual lexicon into a Unified Image-Text Vocabulary ($\mathcal{V}_{\text{unified}}$). Under this formulation, visual signals are represented as tokens and trained with the same next-token prediction objective as language, which enables a unified autoregressive modeling interface and supports direct token-level supervision for dense prediction. This section details the construction of this vocabulary via a specialized tokenizer, followed by the formulation of our unified autoregressive objective.

\paragraph{Construction of Unified Image-Text Vocabulary.}
To realize the VLUAS paradigm, we expand the conventional textual lexicon into a unified image-text vocabulary ($\mathcal{V}_{\text{unified}}$) by introducing a vision tokenizer that maps an image to a sequence of discrete indices through vector quantization. We use these discrete indices as prediction targets, which casts visual learning into the same next-token prediction form as language modeling and enables direct token-level dense supervision via cross-entropy. In parallel, the conditioning pathway remains continuous: the input image is encoded into continuous embeddings by a vision encoder, projected to the LLM hidden dimension, and concatenated with text embeddings as context. This preserves fine-grained visual cues in the context and avoids propagating quantization error into the conditioning signal. A purely continuous alternative would require regressing high-dimensional vectors, departing from the token-based autoregressive interface and making supervision less aligned with standard LLM training. This asymmetric design therefore combines information-preserving continuous conditioning with stable, vocabulary-based autoregressive supervision.

For the vision tokenizer representation, it is important to preserve both high-level semantics and fine spatial structure. Semantic encoders tend to be spatially coarse, whereas pixel-level objectives can retain excessive high-frequency redundancy and encourage texture shortcuts. We address this by designing a synergistic vision tokenizer that forms dense vision-language aligned features before discretization. Specifically, we leverage two frozen foundational encoders with complementary properties: SigLIP-2~\citep{tschannen2025siglip2multilingualvisionlanguage} provides rich language-aligned semantics, while DINOv3~\citep{siméoni2025dinov3} offers boundary-consistent local correspondences via self-distillation, which helps maintain spatial structure. Although DINOv3 contains dense semantics as well, it is not explicitly aligned to language; combining it with SigLIP-2 allows us to bind language-aligned semantic content to geometry-aware spatial layout.

Concretely, we employ a cross-attention fusion mechanism that probes semantic features under structural constraints. We project the feature maps into a shared manifold to define Query ($\mathbf{Q}$), Key ($\mathbf{K}$), and Value ($\mathbf{V}$) matrices:
\begin{equation}
    \mathbf{Q} = \mathbf{H}_{\text{geo}} \mathbf{W}_Q,\quad
    \mathbf{K} = \mathbf{H}_{\text{sem}} \mathbf{W}_K,\quad
    \mathbf{V} = \mathbf{H}_{\text{sem}} \mathbf{W}_V,
\end{equation}
where $\mathbf{H}_{\text{geo}}$ and $\mathbf{H}_{\text{sem}}$ denote the hidden states from the structural (DINOv3) and semantic (SigLIP-2) encoders, respectively. The fused synergistic representation $\mathbf{Z}_{\text{syn}}$ is computed via cross-attention:
\begin{equation}
    \mathbf{Z}_{\text{syn}} =
    \text{Softmax}\!\left(\frac{\mathbf{Q}\mathbf{K}^{\top}}{\sqrt{d_k}}\right)\mathbf{V}.
\end{equation}
Prior to quantization, we concatenate $\mathbf{Z}_{\text{syn}}$ with the original structural features $\mathbf{H}_{\text{geo}}$ along the channel dimension, project the composite representation via an MLP, and discretize it using Index Backpropagation Quantization (IBQ)~\citep{shi2025scalableimagetokenizationindex}. This process maps continuous vectors to the nearest prototype in a learnable codebook $\mathcal{C}=\{c_k\}_{k=1}^K$, configured with a vocabulary size of $K=150{,}000$ and embedding dimension $D=768$.

The tokenizer is optimized end-to-end to reconstruct input images from these discrete codes. We employ a compound objective function $\mathcal{L}_{\text{tok}}$ that balances perceptual fidelity with codebook usage:
\begin{equation}
    \mathcal{L}_{\text{tok}} = \underbrace{\lambda_{p}\mathcal{L}_{\text{lpips}} + \lambda_{g}\mathcal{L}_{\text{gan}}}_{\text{Perceptual \& Adversarial Fidelity}} + \underbrace{\mathcal{L}_{\text{vq}} + \lambda_{e}\mathcal{L}_{\text{ent}}}_{\text{Codebook Optimization}}
\end{equation}
where $\mathcal{L}_{\text{lpips}}$ enforces textural realism via perceptual similarity, and $\mathcal{L}_{\text{gan}}$ denotes the adversarial discriminator loss. To preclude codebook collapse, we integrate a vector quantization loss $\mathcal{L}_{\text{vq}}$ alongside an entropy regularization term $\mathcal{L}_{\text{ent}}$. We set the loss weights to $\lambda_p=1$, $\lambda_g=1$, and $\lambda_e=0.1$. A pivotal design choice in our strategy is the deliberate exclusion of standard pixel-wise $\ell_1$ reconstruction loss. We posit that $\ell_1$ minimization creates a ``texture bias'', encouraging the model to memorize high-frequency noise as a shortcut, thereby bypassing high-level semantic abstraction. By relying exclusively on perceptual and adversarial constraints, we compel the codebook to encode structural semantics rather than mere pixel statistics. Empirically, this approach yields a codebook utilization rate of $97.74\%$ on our composite dataset. The learned indices form the visual vocabulary $\mathcal{V}_{\text{img}}$, which, when merged with the textual vocabulary, establishes the unified foundation $\mathcal{V}_{\text{unified}} = \mathcal{V}_{\text{text}} \cup \mathcal{V}_{\text{img}}$ for our VLUAS paradigm.

\paragraph{Vision-Language Unified Autoregressive Supervision (VLUAS) Paradigm.} 
The core contribution of Youtu-VL is the transition from a discriminative ``vision-as-input'' paradigm to a generative ``vision-as-target'' paradigm. We reformulate the training objective to model the joint probability of vision and language sequences autoregressively. To maximize performance, we employ an asymmetric representation strategy for inputs and targets. For the input context, we utilize continuous visual embeddings extracted by the SigLIP-2 encoder. These features are linearly projected to the LLM dimension and concatenated with textual embeddings, retaining maximum signal fidelity by eliminating quantization errors at the input stage. 

Formally, given a multimodal sequence $\mathcal{S}$ containing textual tokens $T = \{t_1, \dots, t_N\}$ and visual tokens $V = \{v_1, \dots, v_M\}$, the model optimizes the probability of the next token $s_i$ given the context $s_{<i}$, regardless of modality. The unified objective function $\mathcal{L}_{\text{VLUAS}}$ is a weighted summation of textual and visual prediction losses:
\begin{equation}
    \mathcal{L}_{\text{VLUAS}} = \mathcal{L}_{\text{text}} + \lambda \mathcal{L}_{\text{image}}
\end{equation}
where $\lambda$ is a hyperparameter balancing task gradients (empirically set to $0.5$). The individual loss components are calculated using standard cross-entropy:
\begin{equation}
    \mathcal{L}_{\text{text}} = - \sum_{i \in \mathcal{I}_t} \log P(t_i \mid s_{<i}; \theta), \quad 
    \mathcal{L}_{\text{image}} = - \sum_{j \in \mathcal{I}_v} \log P(v_j \mid s_{<j}; \theta)
\end{equation}
Here, $\mathcal{I}_t$ and $\mathcal{I}_v$ denote the indices of text and image tokens in the flattened sequence, respectively. By enforcing $\mathcal{L}_{\text{image}}$, Youtu-VL is compelled to simultaneously reconstruct visual details and linguistic content. This resolves the text-dominant optimization bias, preserves fine-grained visual information typically lost in discriminative approaches.

\subsection{Vision-Centric Predictions from the standard VLM}
\label{sec:cv_archi}

\noindent\textbf{Text Prediction for Vision-Centric Tasks.} We categorize vision-centric tasks into two paradigms: text prediction and dense prediction. The text prediction paradigm encompasses tasks including image classification, object counting, visual grounding, object detection, pose estimation, and polygon-based segmentation. Central to the text prediction paradigm is the precise localization of objects via coordinate prediction. We realize it with three designs. (1) Axis-specific vocabulary: we expanded the tokenizer’s vocabulary by introducing 2048 coordinates for both the X-axis and the Y-axis (e.g., <x\_0>). This design significantly reduces the sequence length required for coordinate representation (wo. x \& y indicators, commas, and long numbers). It also effectively mitigates ambiguity between X and Y values, particularly in multi-coordinate scenarios. (2) Absolute pixel coordinates: the model operates directly on absolute pixel coordinates rather than normalized relative coordinates. This design eliminates the need for coordinate renormalization during training and inference, thereby preventing potential interference caused by the model attempting to learn implicit scaling mappings. (3) Parsing tokens: for precise results parsing, some parsing tokens have been integrated into the vocabulary (refer to Appendix \ref{sec:app_eval_cv}). 

Building upon this vocabulary design, we formulate specific vision tasks as follows. (1) Visual grounding is presented as the prediction of four coordinate tokens XYXY defining the bounding box. (2) Object detection extends the grounding formulation by appending a text category token before the coordinate sequence. The framework supports flexible query modes, ranging from single and multiple target classes, also supporting "detect all" commands for open-world scenarios. (3) Human pose estimation is formulated as the sequential prediction of multiple keypoint coordinates. (4) Polygon-based segmentation can be implemented by predicting a sequence of points to outline object boundaries (compressed within 20 points to avoid very long cases). (5) Object counting: we supported direct regression of a numerical token, or a ``detect-then-count'' approach (see Figure~\ref{fig:case_dnc}) for verifiable accuracy.

\noindent\textbf{Dense Prediction from the standard VLM.} Our goal is to empower standard VLMs to perform mainstream vision-centric tasks, thereby reconciling general-purpose VLMs with these critical yet often overlooked capabilities. It is evident that text generation alone cannot encompass all modalities, particularly pixel or patch-level dense prediction tasks such as semantic segmentation and depth estimation. Conventional approaches typically resort to incorporating additional task-specific decoders~\citep{rasheed2024glamm,wu2024visionllm, liu2025unipixel} and corresponding task embedding tokens~\citep{tang2025ufo}. However, such add-ons inevitably fragment the unified architecture and complicate both training and inference pipelines, thereby undermining practicality. Inspired by the token activation map~\citep{Li_2025_ICCV}, we contend that a standard VLM is inherently a dense predictor capable of yielding dense predictions directly from the output vision tokens without auxiliary modules. Our inference scheme relies on two core mechanisms: (1) Predict categories directly within the unified vocabulary. (2) Take the vision token logits corresponding to the predicted category token indices, with the argmax operation to obtain the results.

For semantic segmentation, categories are standard text tokens. For depth estimation, we discretize the depth range into bins (e.g., 1–1000), supporting both linear quantization for multiple closed sets and logarithmic quantization for open-world tasks (e.g., depth anything scenario). Note that we do not predict each bin name; we apply argmax across all bin vocabularies. In comparison, predicting classes in semantic segmentation is necessary since the general text vocabulary is too large to argmax. To handle categories represented by multiple tokens (e.g., sub-word units), we aggregate their alignment scores by computing the mean of the raw logits $\mathbf{Z}_v$ associated with each category $k$ ($\mathcal{S}_k$ denotes the set of token indices corresponding to the k-th semantic category). These aggregated logits are reshaped ($\mathcal{R}$) into a spatial grid and upsampled via bilinear interpolation ($\mathcal{I}$) to recover pixel-level granularity. Finally, the dense prediction map $\mathbf{M}$ is derived by taking the argmax ($\mathcal{A}$) over the predicted categories:
\begin{equation}
\label{eq:logits_to_mask}
\mathbf{M} = \mathcal{A} \Big( \mathcal{I} \Big( \mathcal{R} \Big( \bar{\mathbf{Z}} \Big) \Big) \Big), \quad \text{where} \quad \hat{\mathbf{Z}}_k = \frac{1}{|\mathcal{S}_k|} \sum_{t \in \mathcal{S}_k} \mathbf{Z}_{v, t} \ .
\end{equation}
To obtain high-quality pixel-level outputs, a Dense CRF \citep{krahenbuhl2011efficient} can be optionally employed as a post-processing step following interpolation. Additionally, a softmax operation with temperature $\tau$ can be applied to the aggregated logits $\hat{\mathbf{Z}}_k$ to regulate the distribution sharpness, thereby controlling the relative contribution of logits to the color values. This operation is optional for semantic segmentation, while depth estimation is not suggested. Another effective strategy to boost performance is image zooming, which leverages the model's native resolution capability to process finer details. Finally, the mask is converted to a string and appended to the conversation via an open-ended language interface for both input and output.

\noindent\textbf{Autoregressive Vision Loss for Multi-Label and Multi-Tasks.} A core contribution of our work is the autoregressive supervision of vision tokens. We extend the next-token prediction (NTP) paradigm to vision tokens to facilitate a tighter alignment between image and text representations, ensuring robust dense prediction performance rather than restricting supervision solely to textual tokens. Acknowledging that a single image patch often encapsulates multiple semantic labels and task targets (see Figure \ref{fig:YoutuVL_pipeline}), we employ a multi-label supervision objective for vision tokens. We term this NTP-M, a variant of the standard NTP adapted for multi-label and multi-task scenarios. Specifically, we transition from a single-label to a multi-label framework by constructing a multi-hot target vector for each image patch, where the token IDs corresponding to all relevant objects, tasks, and potential granularities are set to 1, while the remaining vocabulary indices are assigned 0. Unlike the standard NTP, which models a categorical distribution via Softmax, we model the probability of each token in the vocabulary independently. Consequently, we redefine the generative objective $p(\mathbf{Y} | \mathbf{X}_v, \mathbf{X}_{\text{instruct}})$ as the joint probability of independent Bernoulli trials over the vocabulary:
\begin{equation}
\label{eq:bce_prob}
p(\mathbf{Y} | \mathbf{X}_v, \mathbf{X}_{\text{instruct}}) = \prod_{i=1}^{L} \prod_{v=1}^{|\mathcal{V}|} \sigma(\mathbf{z}_{i,v})^{y_{i,v}} \cdot (1 - \sigma(\mathbf{z}_{i,v}))^{(1 - y_{i,v})} \ ,
\end{equation}
where $\mathbf{X}_v, \mathbf{X}_{\text{instruct}}$ are the vision token input and instruction (prompt) input, respectively. $L$ denotes the sequence length of vision tokens, $|\mathcal{V}|$ is the vocabulary size, $\sigma(\mathbf{z}_{i,v})$ represents the sigmoid-activated probability for token $v$ at vision patch $i$, and $y_{i,v} \in \{0, 1\}$ indicates the ground truth presence of the corresponding target. Note that $\mathbf{X}_{\text{instruct}}$ can influence the dense prediction process; for example, the prompt before the image guides depth estimation for the depth range of a certain camera.

Given the extensive vocabulary size inherent to VLMs, significant class imbalance poses a challenge to effective supervision. To address this, we implement a robust multi-label next-token prediction loss $\mathcal{L}_{\text{NTP-M}}$ that replaces the naive averaging strategy to ensure stable convergence. Our core insight lies in independent positive-negative averaging coupled with relevant negative sampling. Specifically, we decouple the processing of positive and negative samples. We first calculate the mean loss for all valid positive samples. For negative samples, we rank them based on their predicted probabilities $p$ (post-Sigmoid) in Eq. \ref{eq:bce_prob} and compute the average loss only for the top-$k$ relevant negatives. This approach prevents the gradient dilution often caused by averaging an overwhelming number of irrelevant, low-response negative samples. Distinct from standard online hard example mining~\citep{shrivastava2016training}, which ranks positive and negative samples jointly, our method prevents positive samples from being overshadowed or excluded by the sheer volume of hard negatives, thereby enhancing convergence efficiency. Furthermore, to accommodate incomplete annotations (e.g., instances with semantic labels but missing depth information), we apply a validity mask to exclude the corresponding indices from the averaging process. We formally define the sets for positive samples and relevant negative samples to reflect the sorting and masking logic. Let $p_{i,j} = \sigma(z_{i,j})$ be the Sigmoid-activated probability for the $j$-th token at step $i$. Let $\mathcal{M}_i$ be the set of valid indices for the current task (masking out missing annotations). The set of positive indices is defined as $\mathcal{P}_i = \{j \in \mathcal{M}_i \mid y_{i,j} = 1\}$. For the negative samples, let the candidate set be $\mathcal{C}_i = \{j \in \mathcal{M}_i \mid y_{i,j} = 0\}$. We define the relevant negative set $\mathcal{N}_i^{\text{hard}}$ as the subset of $\mathcal{C}_i$ with size $k$ that maximizes the sum of predicted probabilities (i.e., the top-$k$ sorting operation):
\begin{equation}
\mathcal{N}_i^{\text{hard}} = \underset{\mathcal{S} \subset \mathcal{C}_i, |\mathcal{S}|=k}{\arg\max} \sum_{j \in \mathcal{S}} p_{i,j} \ .
\end{equation}
The final loss $\mathcal{L}_{\text{NTP-M}}$ is the summation of the independent averages over these sets:
\begin{equation}
\mathcal{L}_{\text{NTP-M}} = \sum_{i=1}^{L} \left[ -\frac{1}{|\mathcal{P}_i|} \sum_{j \in \mathcal{P}_i} \log(p_{i,j}) - \frac{1}{k} \sum_{j \in \mathcal{N}_i^{\text{hard}}} \log(1 - p_{i,j}) \right] \ .
\end{equation}

%% file: sec/3_pre-training.tex
\section{Pre-training}

\subsection{Training Recipe}

\begin{figure}[htbp]
\centering
\includegraphics[width=\textwidth]{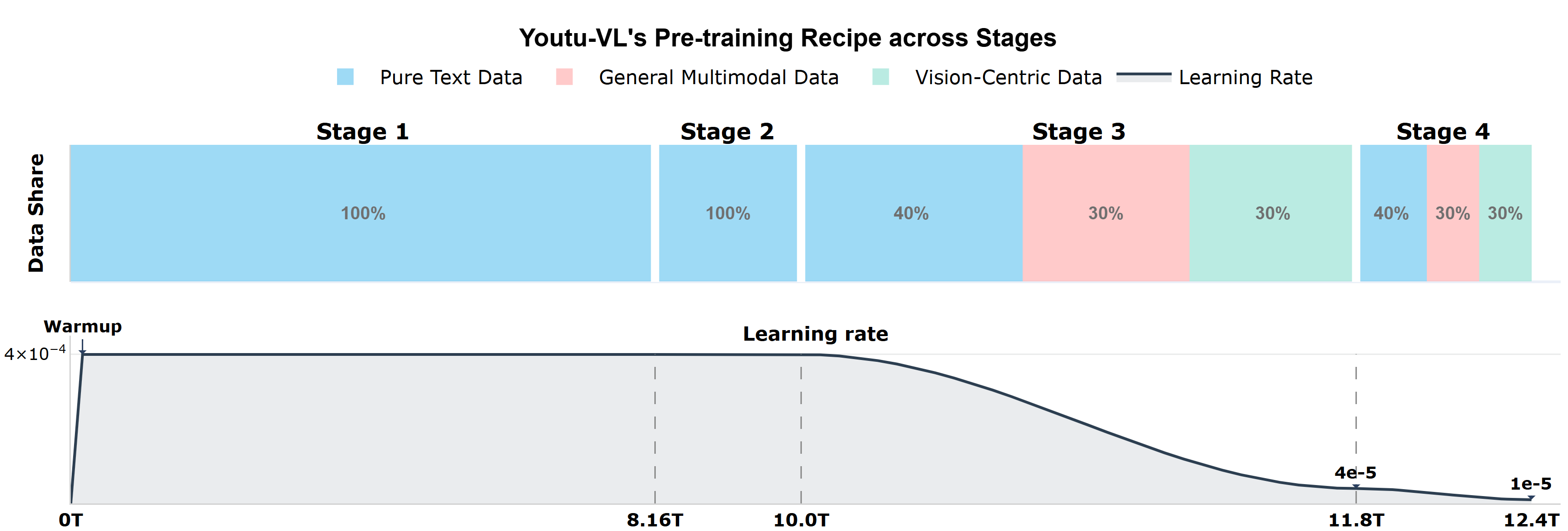}
\caption{\textbf{The pre-training recipe for Youtu-VL}. The top panel illustrates the evolution of the data mixture from Stage 1 to Stage 4: Stages 1 and 2 exclusively utilize pure text data to establish a strong linguistic foundation, while Stages 3 and 4 {progressively enhance multimodal capabilities}. The bottom panel presents the learning rate schedule aligned with the training stages.}
\label{fig:recipe}
\end{figure}

The training of Youtu-VL follows a progressive, multi-stage training recipe. This pipeline is structured into four sequential phases, evolving from establishing a robust language foundation (Stages 1 and 2) to multimodal foundation pre-training. (Stage 3), and culminating in versatile task adaptation (Stage 4). The details of this training recipe are described in Figure~\ref{fig:recipe}. During Stages 3 and 4, we deploy a dual-stream supervision strategy to enforce comprehensive visual supervision. For general multimodal data, we impose the autoregressive visual reconstruction loss $\mathcal{L}_{\text{image}}$ (defined in Section \ref{sec:VL}), compelling the model to predict visual tokens as intrinsic generation targets alongside text. Complementarily, for vision-centric data, we incorporate the specialized loss $\mathcal{L}_{\text{NTP-M}}$ (detailed in Section \ref{sec:cv_archi}), which is essential for facilitating fine-grained dense perception capabilities.

\textbf{{Stages 1 and 2}: Language Backbone Pre-training.} The Youtu-VL language backbone was trained through two distinct phases: Commonsense Pre-training (Stage 1) and STEM- and Coding-centric Pre-training (Stage 2). Collectively, these stages encompassed approximately 10T tokens of pure text data, empowering the model with exceptional proficiency in general reasoning, STEM, and coding tasks. Detailed specifications of the data mixture and training recipe are provided in the Youtu-LLM technical report~\citep{lu2026youtullmunlockingnativeagentic}.

\textbf{Stage 3: {Multimodal Foundation Pre-training.}} To empower the model with robust visual perception and multimodal comprehension, we utilize a comprehensive multimodal dataset comprising image-caption pairs and data for vision-centric tasks. This diverse mixture facilitates the acquisition of broad world knowledge, cross-modal contextual understanding, and proficiency in various visual perception tasks.
During this stage, all components of Youtu-VL—including the LLM backbone, the Vision Encoder, and the Projector—are fully trainable in an end-to-end manner. To preserve and further enhance the linguistic capabilities of the backbone, we incorporate the high-quality text corpus from the General Mid-Training phase of Youtu-LLM. By mixing this text data with multimodal samples, the model is trained on a total of approximately 1.8T tokens, balancing visual and linguistic capabilities.

\textbf{Stage 4: Versatile Task Adaptation.} This phase specializes the model with capabilities across a wide range of tasks, including General VQA, OCR, STEM, GUI, Detection, Segmentation, Grounding, and Pose Estimation. The entire model undergoes end-to-end training on a diverse, high-quality multimodal instruction dataset comprising approximately 0.6T tokens, enabling it to generalize effectively to diverse user instructions. Furthermore, we incorporate a substantial volume of high-quality, synthetic short Chain-of-Thought (CoT) data. This stage empowers the model with long-context understanding and generation, alongside fundamental logical reasoning skills.

\subsection{Pre-Training Data}

\subsubsection{Vision-Centric Data}

\noindent\textbf{Text Data.} Most of the vision-centric tasks described in Section \ref{sec:cv_archi} are achieved through text predictions. The majority of the data is constructed based on open-source datasets, combined with some internal data and synthetic data. We have elaborated on the data details according to the tasks. (1) Visual grounding: Beyond leveraging widely adopted open-source grounding datasets, we curated a diverse set of grounding queries derived from object detection and scene graph datasets. Furthermore, we employed an automated annotation pipeline to synthesize additional grounding data from large-scale image-text pairs. (2) Object detection: We utilized multiple open-source object detection datasets and formulated detection tasks with varying requirements based on bounding box annotations, enabling the model to recognize a broad spectrum of object categories. Additionally, we synthesized a denser and more diverse dataset to further bolster the model's detection capabilities. (3) Polygon-based segmentation targets referring expression and instance tasks. We compress points within 20 for efficient training and avoid out of the window size.

For concept understanding and keypoint perception, the data include: (1) Object counting: In addition to open-source counting datasets, we augmented our training set by synthesizing counting data derived from visual grounding and detection data. Two inference variants are supported: directly counting and a detect-then-count manner. (2) Image classification: We leveraged open-source classification datasets to establish the model's classification proficiency. This enables the recognition of not only common objects and scenes but also fine-grained biological taxonomic concepts. (3) Human pose estimation: We formulate pose estimation as a unified keypoint prediction problem, directly regressing joint coordinates for all persons in an image within a single forward pass. 

\noindent\textbf{Dense Labeled Data.} Dense labeled data is primarily derived from open-source datasets and synthetic data. We have elaborated on the data processing strategies according to the specific dense prediction tasks. (1) Semantic segmentation: We structured the training data by mapping numeric mask labels to text descriptions. During the dataloading phase, labels are converted into token indices, where valid category indices are assigned positive labels, and unlabeled or invalid regions are explicitly ignored during loss calculation. (2) Instance and referring segmentation: We integrated binary segmentation with detection and grounding tasks to support high-quality scenarios. We employed visual prompts via random-colored bounding boxes and assigned foreground (\texttt{<FG>}) and background (\texttt{<BG>}) vocabularies to enable high-quality box-prompted segmentation. Furthermore, we applied diverse augmentation techniques, including padding, cropping, and resizing, to enhance model robustness. (3) Depth estimation: We implemented a linear quantization pipeline to map valid depth ranges into discrete bins (1–1000) for a known camera, with label 0 for ignore labels. Prompts are processed before the image to allow the model to adapt to varying camera parameters, and augmentation involves color jitter and random cutout, using original values for high-resolution data and fixed-ratio resizing (e.g., 3x) for low-resolution sources. The predictions are then dequantized to the real depth for evaluation.

\begin{figure}[hbtp]
  \centering
  \includegraphics[width=0.9\linewidth]{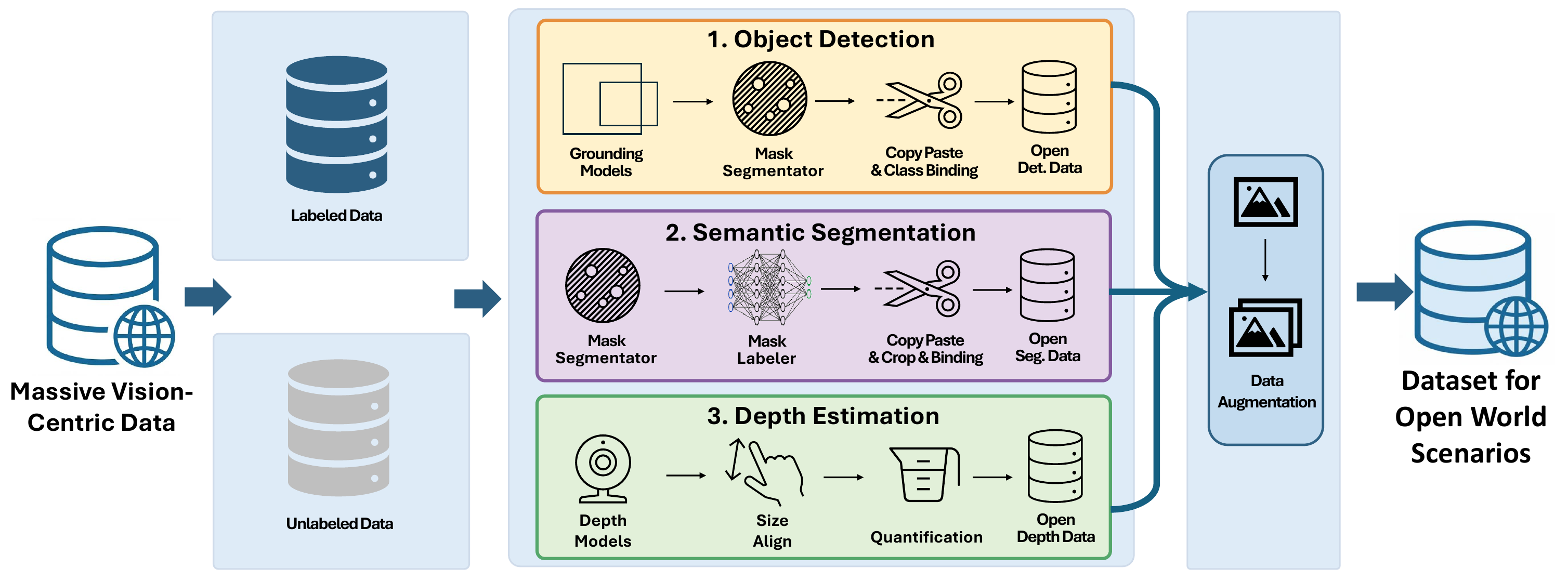}
  \caption{\textbf{Data synthesis pipeline for open-world scenarios}. The framework processes massive vision-centric data through two parallel branches: (top) object detection and semantic segmentation, utilizing grounding models for raw data and data binding strategies for labeled data; and (bottom) depth estimation, employing depth models and quantization. The pipeline applies specific augmentations to generate a comprehensive dataset for open-world tasks.}
  \label{fig:caption_pipeline}
\end{figure}

\noindent\textbf{Data Pipeline for Open World Scenarios.} The open-scene dataset is constructed by processing massive vision-centric data through two primary branches to support object detection, segmentation, and depth estimation. For object detection and semantic segmentation, the pipeline handles both raw and labeled data. Raw data is processed via grounding models and segmentators, while labeled data undergoes data binding to handle single, multiple, or sets of classes. This synthesis addresses four requirements: specifying positive categories, mixing positive and negative categories (for closed-set scenarios), and generating dense scenes. These inputs result in class-merged and class-sampled outputs. Specifically, the "arbitrary category" scenario employs a Copy-Paste strategy where transparent objects undergo random resizing and rotation before being densely placed on backgrounds. For depth estimation, the pipeline similarly splits processing. Raw data is passed through depth models to generate pseudo-labels using open-source models. Labeled data undergoes resize and quantization to simulate stable camera parameters with a fixed focal length of 2000 pixels. We utilize log-uniform quantization with flexible prompt placement to define a valid depth range of 0.5m to 100m; consequently, input images with differing focal lengths represent relative depth and require scaling. Finally, the outputs from both branches pass through task-specific augmentations (detection, segmentation, and depth) and cropped binary segmentation to yield the final dataset for open world scenarios.

\subsubsection{Image Caption and Knowledge Data}

Our data acquisition pipeline begins with a massive aggregation of open-source image-text pairs. To ensure the integrity and robust foundation of this corpus, we implemented a rigorous multi-stage filtration protocol. This initial phase involved strict image-text alignment filtering via CLIP scores~\citep{radford2021learning,wei2025hq,fang2023data}, the excision of NSFW content, and resolution constraints. Following the pruning of unavailable links and corrupted data, this preliminary cleaning yielded a raw corpus comprising approximately 5 trillion tokens.

Building upon this foundation, we sought to maximize information density and training efficiency through a three-pronged strategy: Concept-Balanced Sampling, Rare Class Mining, and Knowledge-Injected Recaptioning.

\begin{figure}[hbtp]
  \centering
  \includegraphics[width=0.9\linewidth]{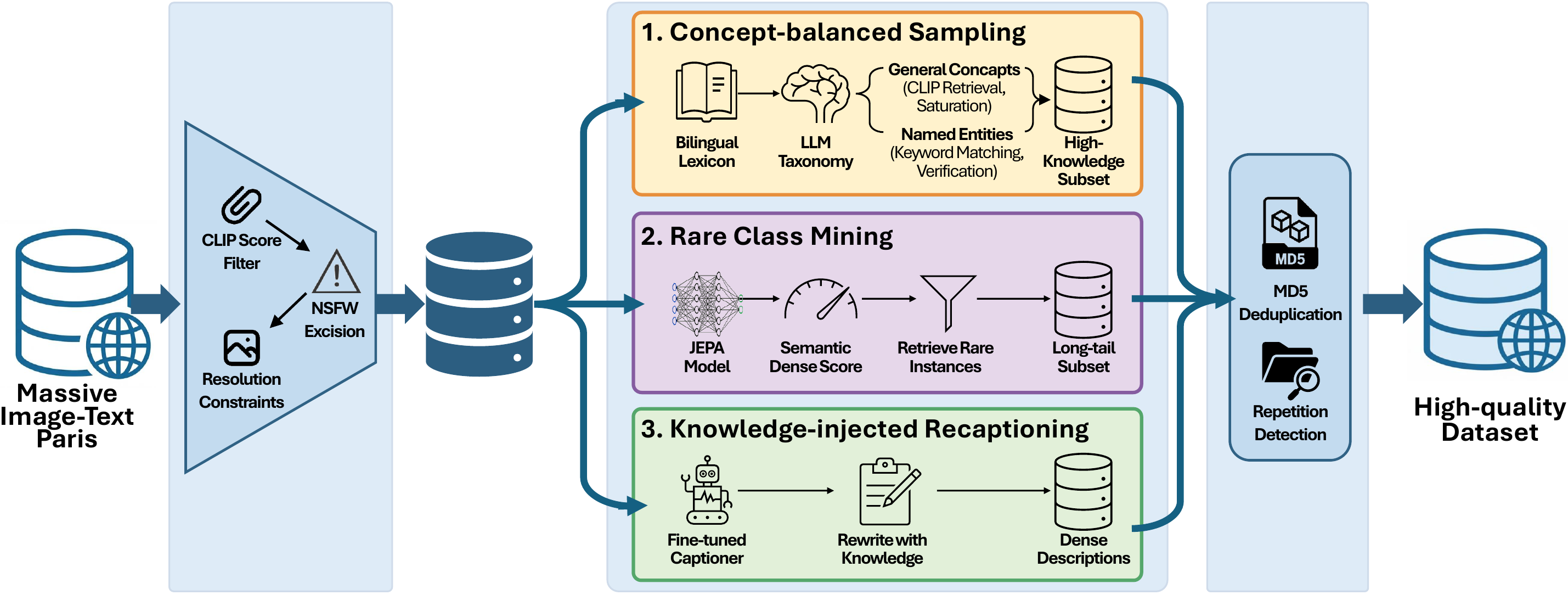}
  \caption{\textbf{The workflow for synthesizing knowledge-dense image caption and knowledge data}. Starting from massive raw image-text pairs, the pipeline proceeds through three main stages: (1) a multi-stage filtration protocol to ensure basic quality; (2) a core enhancement phase featuring {Concept-based Sampling}, {Rare Class Mining}, and {Knowledge-Injected Recaptioning} to maximize information density and diversity; and (3) a final purification stage for deduplication.}
  \label{fig:caption_pipeline}
\end{figure}

\begin{itemize} 
\item \textbf{Concept-Balanced Sampling.} Inspired by the methodologies proposed in MetaCLIP~\citep{chuang2025meta}, we moved beyond random sampling to a rigorous, ontology-driven approach. We curated a comprehensive bilingual lexicon (Chinese and English) derived from multiple encyclopedic sources. LLMs were leveraged to taxonomize each entry into \textit{general concepts} (e.g., "soccer," "cinematography") or \textit{named entities} (e.g., specific movie titles, historical figures). For general concepts, we employed CLIP-score retrieval to mine the training data, continuing until a pre-defined saturation threshold was met for each concept. For named entities, where factual precision is paramount, we utilized stringent keyword matching and entity verification to ensure correct association. This balanced sampling strategy allowed us to effectively extract a subset of high-knowledge-density data from the otherwise sparse and noisy raw distribution.

\item\textbf{Rare Class Mining via JEPA Score.} Concept-based sampling is inherently biased towards common entities, leaving the significant long-tail of the data distribution underexplored. To rectify this, we leveraged a JEPA-based density metric to identify and retrieve rare instances located in the sparse regions of the latent space. Specifically, we utilized state-of-the-art JEPA (Joint-Embedding Predictive Architecture) series models to compute a "semantic density score" for each data point, akin to methods discussed in \citep{balestriero2025gaussianembeddingsjepassecretly}. By identifying samples in low-density regions of the latent space, we selectively retained rare but semantically unique examples based on adaptive thresholds, ensuring the model is exposed to a diverse visual manifold.

\item\textbf{Knowledge-Injected Recaptioning.} While large-scale open-source image-captioning datasets have fueled recent progress, they frequently suffer from brevity and noise. To elevate the model's fine-grained visual understanding, we developed a specialized "Knowledge-Injected Captioner" by fine-tuning a multimodal model. This captioner was deployed to rewrite the dataset, transforming sparse web texts into dense, detailed, and visually grounded descriptions. This process not only hallucinates less but also aligns the textual modality precisely with the visual content.
\end{itemize}

Finally, the curated dataset underwent a purification stage. We applied MD5 checksum deduplication to remove redundancy and utilized repetition detection heuristics to eliminate cyclical generation artifacts. The culmination of this pipeline is a clean, compact, and high-fidelity dataset containing approximately 1 trillion tokens, serving as the cornerstone of our visual-language alignment.

\subsubsection{Optical Character Recognition (OCR) Data}

To bolster proficiency in OCR and fine-grained chart understanding, we curated a high-quality dataset focusing on visual perception and reasoning. Initially, we integrated existing open-source human-annotated datasets to establish a baseline capability. However, we identified significant limitations in these legacy resources: (1) \textbf{Brevity}, where responses are often overly concise; (2) \textbf{Monotony}, characterized by a lack of diversity in prompt patterns; and (3) \textbf{Absence of Intermediate Steps}, as human annotations typically provide final answers without the intermediate reasoning process (\textit{i.e.}, short Chain-of-Thought). This absence leads to suboptimal training efficiency and limited generalization to complex, unseen queries. Consequently, we shifted our strategy to synthesize a large-scale, verbose dataset, which constitutes the core of our training corpus.

Our data synthesis pipeline employs a three-pronged approach to ensure high quality and diversity. First, to leverage vast amounts of unlabeled data, we aggregated a collection of raw PDFs, academic charts, and real-world document images. We utilized an LLM to generate diverse questions from various perspectives and difficulty levels, followed by a powerful VLM to produce detailed, descriptive answers, creating rich instruction-following pairs. Second, targeting the deficiencies in existing human-labeled data, we introduced a logic-consistency refinement pipeline. An LLM evaluates the logical coherence of existing QA pairs; for samples deemed opaque, a VLM regenerates the response with explicit step-by-step reasoning. This process is safeguarded by a closed-loop verification mechanism where an LLM validates that the new reasoning path remains consistent with the original ground truth. Finally, to further scale the training data, we constructed a massive synthetic OCR dataset from pure text corpora using a robust rendering engine. This engine applies randomized typographical augmentations and simulates real-world acquisition conditions—specifically by superimposing complex background textures (\textit{e.g.}, paper grain, shadows) and applying physical distortions (\textit{e.g.}, blur, noise, affine transformations, and rotation)—to generate millions of clean, accurate, and visually diverse training samples.

\subsubsection{Science, Technology, Engineering, and Mathematics (STEM) Data}

The STEM dataset is designed to enhance capabilities in image-based analysis and reasoning. The dataset spans a wide range of visually intensive STEM scenarios, including geometric diagrams, plots and charts, physics illustrations, chemical structures, engineering schematics, and exam-style problem figures. To construct a dataset capable of supporting complex reasoning, we implemented a sophisticated curation pipeline comprising three key stages:

\begin{figure}[hbtp]
  \centering
    \includegraphics[width=1.0\linewidth]{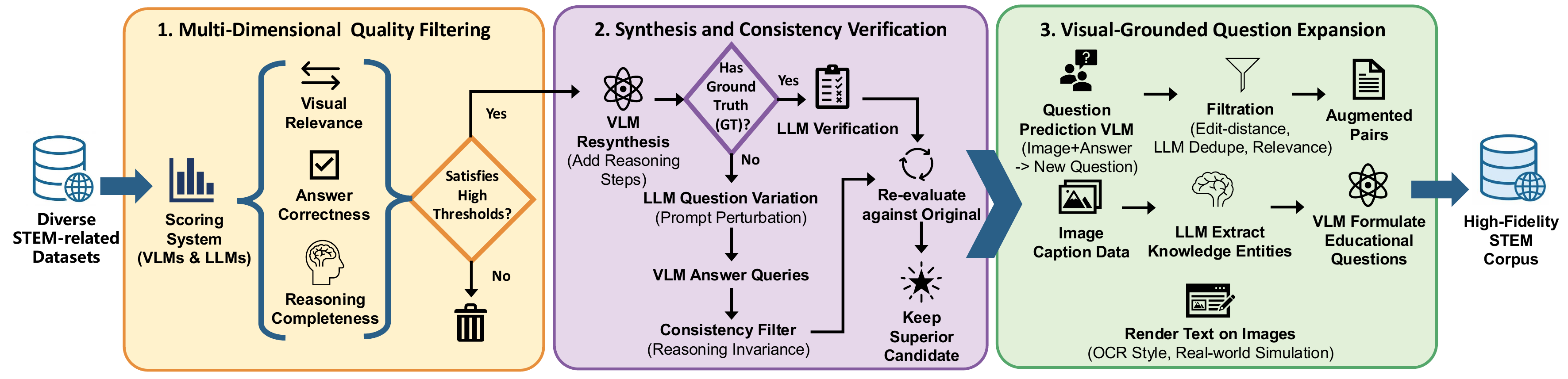}
  \caption{\textbf{Data construction pipeline for STEM}. The pipeline consists of three key phases: (1) Multi-Dimensional Quality Filtering to ensure high visual and logical standards; (2) Synthesis and Consistency Verification to enhance reasoning details and ensure fidelity; and (3) Visual-Grounded Question Expansion to augment the dataset with diverse queries and real-world simulations.
  }
  \label{fig:stem_pipeline}
\end{figure}

\begin{itemize}
    \item \textbf{Multi-Dimensional Quality Filtering.} We initiated the process by aggregating diverse STEM-related datasets. To purge low-quality samples, we deployed a scoring system utilizing both VLMs and LLMs. Each sample was rigorously evaluated across three critical dimensions: \textit{visual relevance}, \textit{answer correctness}, and \textit{reasoning completeness}. Only data satisfying high thresholds in all metrics were retained for the subsequent refinement phases.

    \item \textbf{Synthesis and Consistency Verification.} Recognizing that legacy data often lacks detailed reasoning steps, we utilized a VLM to resynthesize answers conditioned on the image and question. To ensure the fidelity of these synthetic responses, we adopted a bifurcated verification strategy. For samples with existing ground truth (GT), an LLM verified the semantic alignment between the synthesized answer and the original answer. For synthetic datasets lacking ground truth, we implemented a verification protocol. We employed an LLM to generate multiple semantically equivalent but syntactically distinct variations of the original question (i.e., prompt perturbation). The VLM was then tasked with answering these diverse queries. We enforced a strict consistency filter, retaining only those samples where the model demonstrated {reasoning invariance}—yielding consistent answers across all query variations. This ensures that the generated knowledge is stable. Finally, we re-evaluated the new, verbose answers against the original ones using our quality metrics, preserving the superior candidate to maximize information density.

    \item \textbf{Visual-Grounded Question Expansion.} A single question often fails to exhaust the knowledge embedded in a complex image. To rectify this, we fine-tuned a specialized "Question Prediction VLM" that takes an image and an answer as input to predict the corresponding question. This allowed us to generate diverse, meaningful queries for each image. These augmented pairs underwent filtration via edit-distance heuristics, LLM-based deduplication, and visual relevance scoring. Additionally, we repurposed our Image Caption data by using LLMs to extract knowledge entities from captions, prompting a VLM to formulate educational questions based on these points. Finally, to simulate real-world usage scenarios (e.g., photographing exam papers), we adopted the rendering strategy from our OCR pipeline. Textual questions were dynamically rendered onto the images, creating a robust dataset that mimics the visual noise and layout of physical documents.
\end{itemize}

Through the synergy of these pipelines, we ultimately constructed a high-fidelity STEM corpus characterized by rigorous reasoning chains, diverse query perspectives, and robust visual grounding, significantly enhancing the model's capabilities in STEM.

\subsubsection{Graphical User Interface (GUI) Data}

To empower Youtu-VL with agentic capabilities for autonomous GUI interaction, we conduct continual pre-training using a dual-stream data curation pipeline composed of \textbf{single-turn grounding data} and \textbf{long-horizon interaction trajectories.} 

\begin{itemize}
    \item \textbf{Granular Perception \& Grounding}: To establish robust atomic UI understanding and operational capability, we aggregate single-turn data from various open-source datasets. We implement a rigorous data production and filtering process, utilizing an ensemble LLM-as-a-judge framework \citep{lin2025cuarewardbench} to ensure label correctness. This high-fidelity data is crucial for enhancing the model's fine-grained capabilities, specifically in element description, dense captioning, and dense grounding, thereby enabling a robust understanding of diverse user interfaces.
    \item \textbf{Sequential Interaction Dynamics}: To advance the policy model's reasoning regarding environmental dynamics and multi-step interaction logic, we synthesize large-scale, cross-platform trajectories spanning desktop, mobile, and web environments \citep{xie2024osworld, shi2025youtu} using large-scale sandbox systems. This synthesis pipeline is designed to maximize task diversity and simulate realistic operation environments across different applications. Throughout the synthesis process, we implement strict privacy and access control protocols to ensure the produced trajectories in the sandbox are free from personally identifiable information and unauthorized content access.
    \item \textbf{Reward-Guided Hybrid Verification}: We enforce strict quality control across both data streams through a hierarchical filtering strategy. Candidates are ranked and filtered based on scores from our ensemble judging framework \citep{lin2025cuarewardbench}: the highest-scoring trajectories undergo rigorous human annotation to construct a premium dataset for post-training, while the remaining validated data is utilized for continual pre-training.
\end{itemize}

\subsubsection{Pure Text Data}
To preserve the model's fundamental language abilities, we incorporated the mid-training data used by Youtu-LLM, encompassing domains such as Web, Encyclopedia, STEM, and Coding. Comprehensive details regarding data acquisition and curation can be found in the Youtu-LLM technical report~\citep{lu2026youtullmunlockingnativeagentic}.

\begin{figure*}[htbp]
  \centering
  \includegraphics[width=0.7\linewidth]{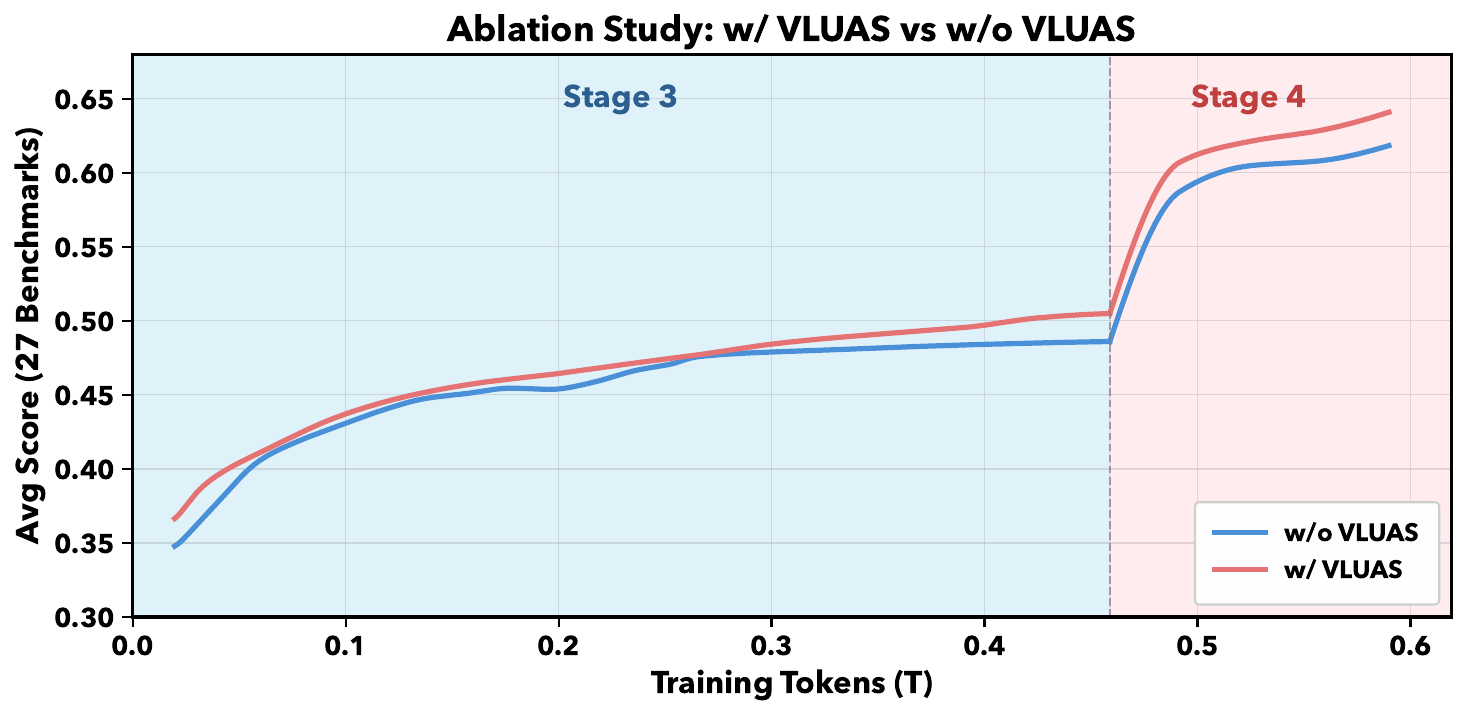}
  \caption{\textbf{Ablation Study on VLUAS Effectiveness.} 
  Comparative scaling curves of the model trained with (red) and without (blue) the proposed Unified Pre-training (VLUAS) strategy. The results indicate a critical divergence in scaling behavior: while the baseline model (wo/ VLUAS) exhibits clear signs of performance saturation during the later phases of Stage 3, the VLUAS-enhanced model maintains a superior performance trajectory with higher data efficiency. This consistent gap across both Stage 3 and Stage 4 empirically validates that incorporating visual supervision significantly alleviates data saturation and effectively raises the upper bound of multimodal capabilities.}
  \label{fig:ablation}
\end{figure*}

\subsection{Analysis of Pre-training Paradigm}

Standard VLM benchmarks predominantly employ zero-shot settings. However, pre-training checkpoints often lack instruction alignment, leading to format mismatches that obscure the model's intrinsic capabilities. To decouple representation quality from instruction-following skills, we adopt a few-shot evaluation protocol aligned with recent foundational LLM assessment methodologies~\citep{deepseekai2025deepseekv3technicalreport}. Specifically, we employ in-context learning to standardize output formats, thereby enabling a fair assessment of the base model's progress without relying on explicit instruction tuning.

Utilizing this rigorous protocol, we analyze the scaling dynamics of multimodal understanding across Stages 3 and 4. As illustrated in Figure~\ref{fig:ablation}, the average performance across 27 multimodal benchmarks highlights a distinct divergence between our VLUAS paradigm and the conventional text-dominant supervision baseline. The text-dominant model exhibits signs of performance saturation early in Stage 3, pointing to a fundamental bottleneck when supervision is confined to the linguistic modality. Conversely, VLUAS sustains a steep learning trajectory, effectively bypassing this premature plateau. This advantage extends into Stage 4; even with improved data quality, VLUAS demonstrates superior data efficiency and consistently widens the performance gap. These empirical results confirm that integrating generative visual supervision serves not merely as an auxiliary task, but fundamentally elevates the model's capability ceiling.

\subsection{Analysis of Data Scaling}

\begin{figure*}[htbp]
  \centering  
  \begin{subfigure}[b]{0.41\linewidth}
    \centering    
    \includegraphics[width=\linewidth]{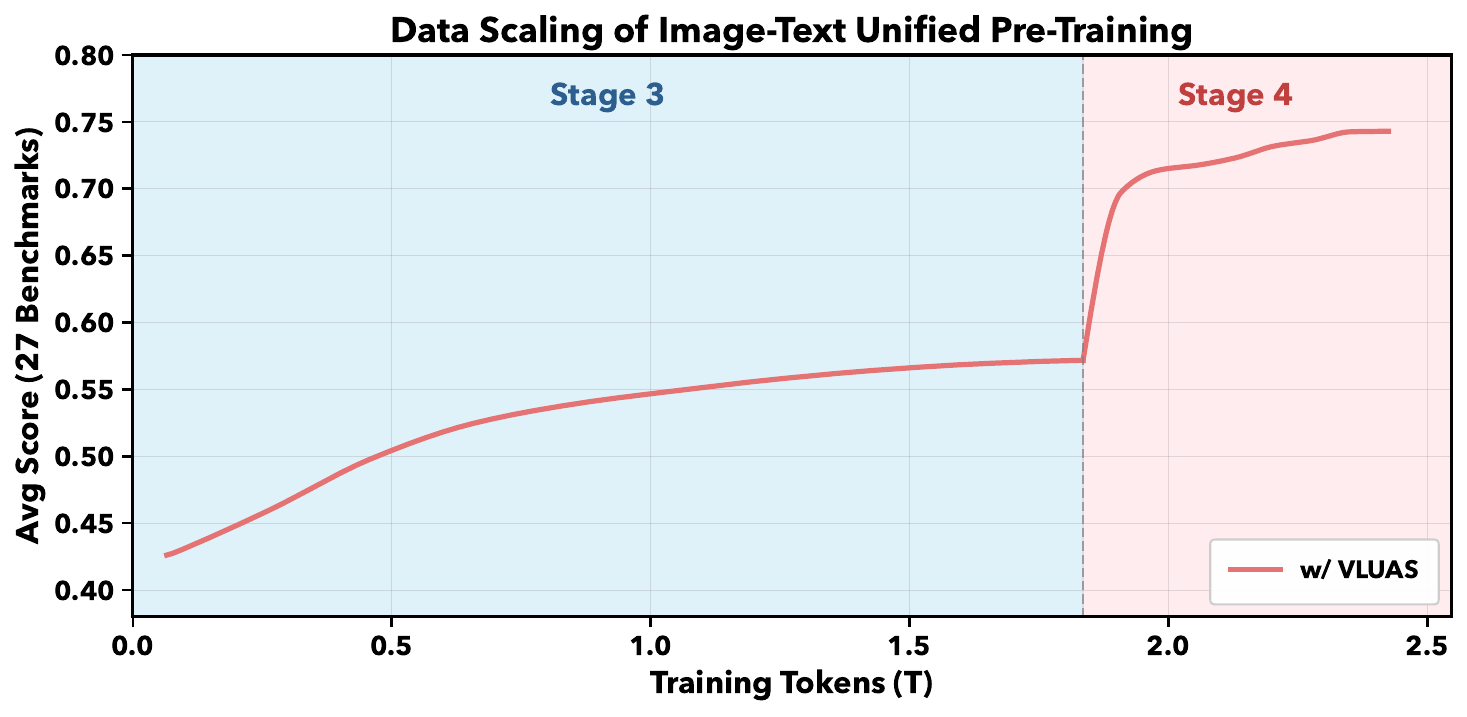}    
    \caption{\textbf{Scalability of Unified Pre-training.}}
    \label{fig:scaling}
  \end{subfigure}
  \hfill   
  \begin{subfigure}[b]{0.57\linewidth}
    \centering
    \includegraphics[width=\linewidth]{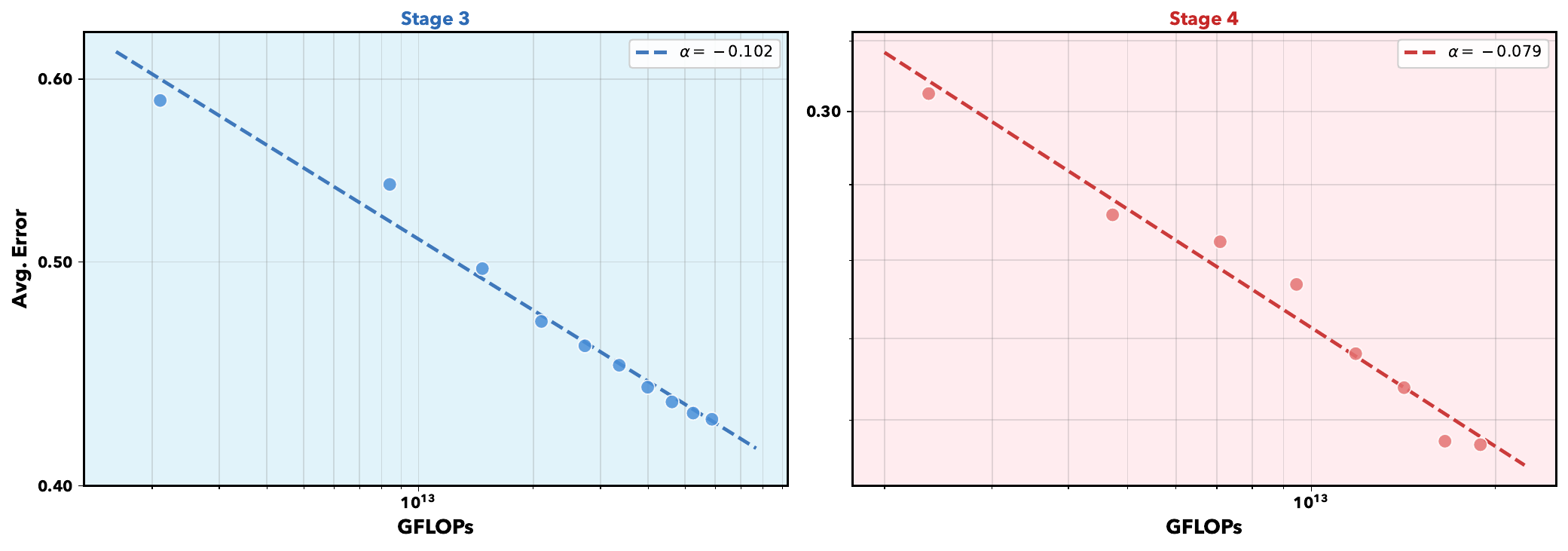}
    \caption{\textbf{Power-Law Scaling Dynamics.}}
    \label{fig:scaling_law}
  \end{subfigure}

  \caption{\textbf{Scaling analysis of the VLUAS paradigm.}
  (a) The scaling trajectory of the average performance across 27 multimodal benchmarks with respect to training tokens. The model demonstrates robust and continuous capability growth throughout the training process across distinct data regimes (Stage 3 and Stage 4), effectively leveraging massive-scale data without exhibiting empirical saturation.
  (b) The analysis of average evaluation error ($1 - \text{Score}$) versus training compute (GFLOPs) on a log-log scale. The training dynamics in both Stage 3 and Stage 4 strictly adhere to neural scaling laws, characterized by scaling exponents of $\alpha \approx 0.102$ and $\alpha \approx 0.079$, respectively. This precise linear relationship confirms the predictability and computational efficiency of the VLUAS paradigm.}  
  \label{fig:combined_scaling_analysis}
\end{figure*}

To systematically evaluate the scalability and stability of our pre-training, we conducted a comprehensive analysis spanning 2.4T tokens. As illustrated in Figure~\ref{fig:scaling}, the model exhibits a robust, monotonic positive correlation between the unified training volume and multimodal performance (averaged across 27 benchmarks). The training trajectory spans two distinct data regimes, Stage 3 (Multimodal Foundation Pre-training) and Stage 4 (Versatile Task Adaptation), where our optimization strategy consistently translates increased computational investment into tangible capability gains. Throughout the pre-training lifecycle, encompassing both the foundation learning on mixed text-visual corpora in Stage 3 and the high-quality instruction tuning in Stage 4, the average benchmark score demonstrates a sustained upward trend, rising from approximately 0.43 to a final convergence of over 0.74. This continuous growth confirms that our pre-training effectively leverages massive-scale data, maintaining high sample efficiency without encountering the early saturation or instability often observed in multi-task optimization.

By evaluating how evaluation error ($\epsilon$) scales with training compute (GFLOPs), we quantify the predictability of the model's performance trajectory. As visualized in the log-log plots of Figure~\ref{fig:scaling_law}, our unified pre-training process closely follows neural scaling laws, fitting the power-law equation $L(C) \propto C^{-\alpha}$. Specifically, the Stage 3 regime exhibits a steep scaling exponent of $\alpha \approx 0.102$, indicating that the integration of general text corpora with multimodal samples facilitates rapid information absorption and high data efficiency in the initial phase. Upon transitioning to Stage 4, the model enters a refined scaling regime with $\alpha \approx 0.079$. While the exponent naturally moderates as performance approaches the irreducible error floor, the sustained linearity on the logarithmic scale serves as strong empirical evidence that our training infrastructure maintains computational efficiency even during complex task adaptation and long-context CoT generation, effectively converting massive resources into predictable model improvements.

\subsection{Visual Representation Analysis}

\begin{figure}[htbp]
  \centering
  \includegraphics[width=1\linewidth]{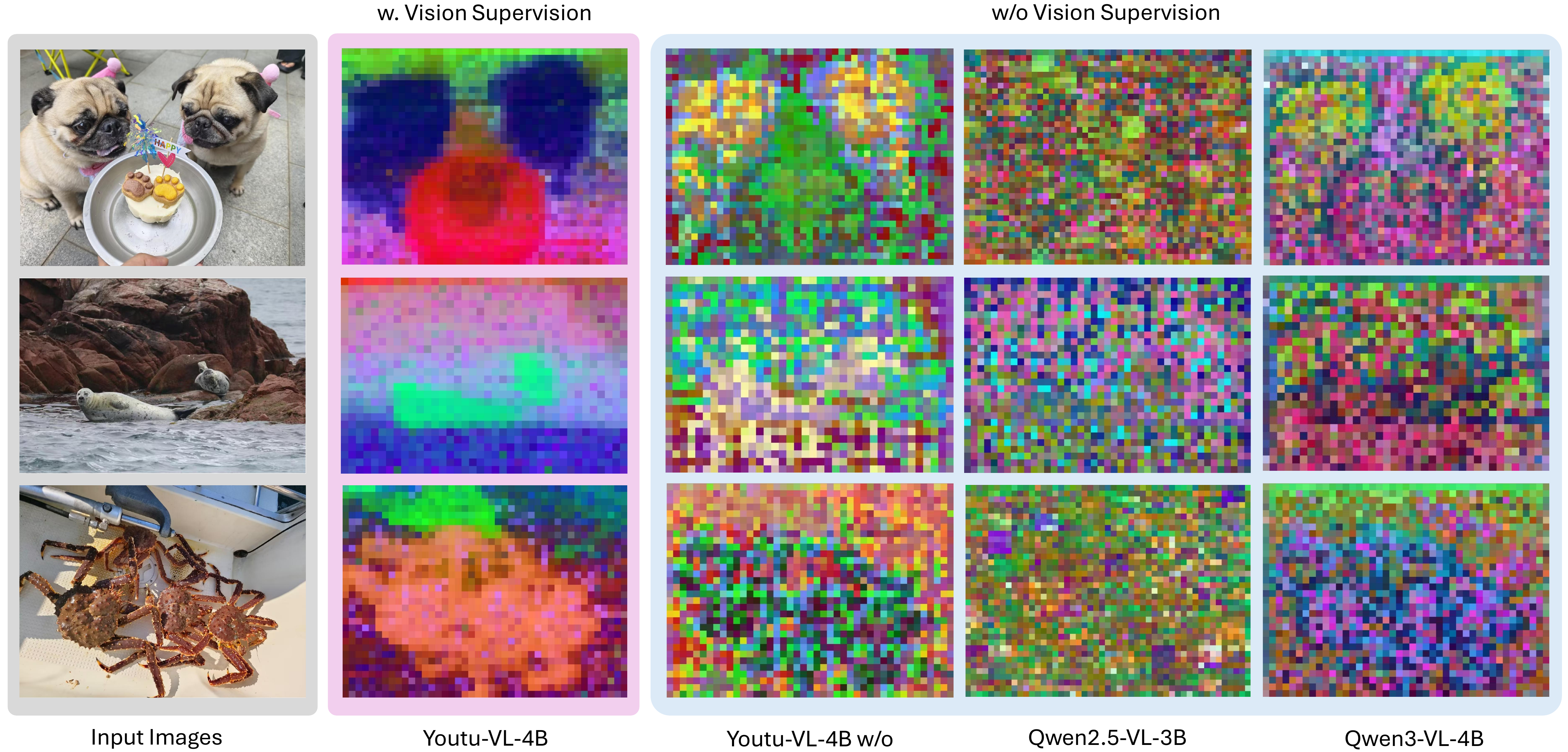}
  \caption{\textbf{Visualization of Vision Token Representations}. This figure compares the Principal Component Analysis (PCA) visualizations~\citep{oquab2023dinov2} of the last-layer hidden states of vision tokens. We contrast Youtu-VL-4B (with vision token supervision) against three models without such supervision: Youtu-VL-4B w/o, Qwen2.5-VL-3B-Instruct~\citep{qwen2.5}, and Qwen3-VL-4B-Instruct~\citep{Qwen3-VL}. Leveraging vision token supervision, our model exhibits outstanding feature separation and visualization quality compared to VLMs lacking this supervision.}
  \label{fig:rep_ana}
\end{figure}

To evaluate the quality of visual representations, we performed PCA on the last-layer hidden states of the vision tokens output by the LLM and projected them back onto the spatial dimensions of the original images, as shown in Figure \ref{fig:rep_ana}. For VLMs lacking vision token supervision, we observe a clear progression from Qwen2.5-VL to Qwen3-VL; the latter exhibits superior representation quality with greater distinctiveness between objects, which positively correlates with its improved model performance. Notably, our Youtu-VL-4B (without vision supervision) demonstrates representation capabilities comparable to Qwen3-VL. However, upon integrating vision token supervision, the visual representations of Youtu-VL-4B show significant improvement, characterized by clear semantic structures and sharp object separation. This qualitative result provides compelling evidence for the effectiveness and necessity of our proposed vision supervision, especially in the visual representation aspect.

%% file: sec/4_post-training.tex
\section{Post-training}

\subsection{Supervised Fine-Tuning}
\subsubsection{Training Recipe}
The supervised fine-tuning aims to refine the model's capability to understand complex instructions, enhance reasoning, and align with human preferences. During this phase, we extend the context window from the initial 16K tokens to 32K tokens. We employ the AdamW~\citep{loshchilovdecoupled} optimizer with a cosine learning rate scheduler. Specifically, we set a 5\% warmup ratio, with the learning rate decaying from a peak of $2 \times 10^{-5}$ to $2 \times 10^{-6}$.

\subsubsection{Data Curation}

We are dedicated to curating a comprehensive, high-quality SFT dataset that encompasses a diverse spectrum of multimodal tasks. To achieve this, we implemented a multi-source data acquisition strategy:
\begin{itemize}

    \item \textbf{High-Quality Mining from Pre-training Corpora.} To extract high-value instruction-following data from our massive pre-training corpus, we employed a stratified sampling strategy. We first leveraged a VLM to assess the quality and alignment of individual samples. To further ensure task diversity, we established a fine-grained keyword taxonomy. By balancing samples against this taxonomy and filtering based on VLM scores, we curated a premium subset from the raw pool, effectively maximizing data efficiency.

    \item \textbf{Refined Open-Source Data.} Existing open-source datasets are often limited by simple "Yes/No" or single-word answers, which are insufficient for training robust reasoning capabilities. We treated these datasets merely as a source of images and prompts, while synthesizing entirely new target responses. By employing a rigorous rewrite-and-expand pipeline, we prompted a powerful VLM to analyze the visual details and generate comprehensive, paragraph-level descriptions and reasoning steps. This approach effectively "up-scales" the information density of the original data by an order of magnitude.
\end{itemize}

\begin{figure*}[htbp]
  \centering
  \includegraphics[width=0.8\linewidth]{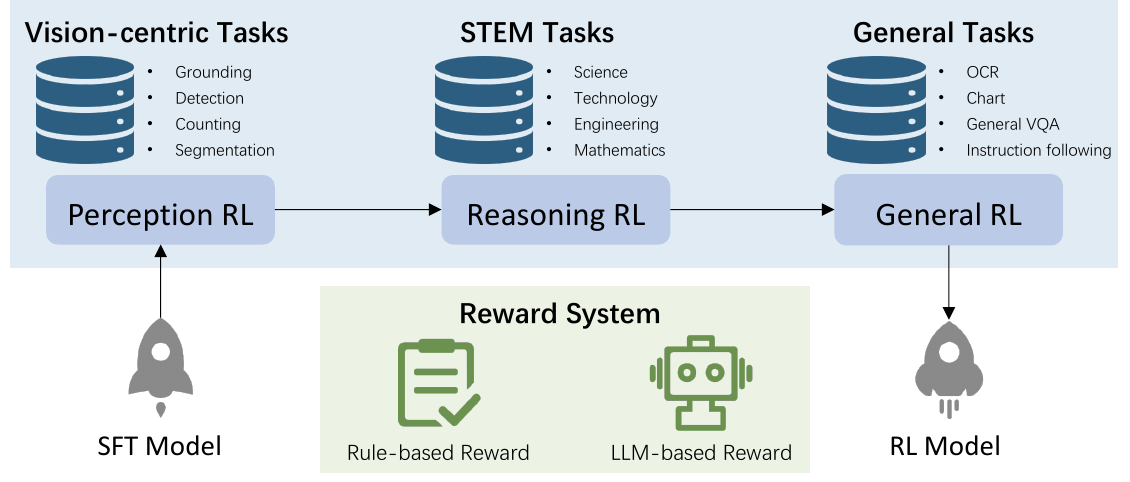}
  \caption{\textbf{The Architecture of our multi-stage reinforcement learning framework}.}
  \label{fig:rl_architecture}
\end{figure*}
\subsection{Reinforcement Learning}
To further unlock our model's potential beyond the supervised fine-tuning phase, we employ a multi-stage reinforcement learning framework.  As illustrated in Figure~\ref{fig:rl_architecture}, our proposed framework follows a
sophisticated three-stage training process:

\begin{itemize}
\item {\bf Perception RL.} The first stage is dedicated to enhancing the model's fine-grained visual perception. During this phase, we focus on vision-centric tasks, such as visual grounding, object detection, semantic segmentation, object counting, \textit{etc.} By utilizing RL objectives tailored for these tasks, our model is optimized to precisely localize and interpret structural elements within an image.
\item {\bf Reasoning RL.} The second stage targets augmenting the model's complex reasoning capabilities. We leverage a curated STEM (Science, Technology, Engineering, and Mathematics) dataset that integrates both text-only and multimodal data. By employing verifiable reward signals to guide the model, we effectively bridge the gap between fundamental visual recognition and sophisticated problem-solving.
\item {\bf General RL.} Building upon the reinforcement learning on verifiable STEM tasks, the third stage shifts the focus towards broader generalization across a wider range of tasks. By incorporating a diverse set of general VQA challenges, including OCR, chart understanding, instruction following, \textit{etc.}, we further improve the model's generalization capabilities.
\end{itemize}

\subsubsection{Data Curation}
To ensure the stability and efficiency of our multi-stage reinforcement learning framework, we implement a rigorous data curation pipeline. Our objective is to construct a diverse, high-quality and most importantly, \textit{verifiable} dataset across various domains, including vision-centric tasks, STEM, OCR, chart understanding, general VQA, and complex instruction following. The curation process consists of the following four steps:

\begin{itemize} 
	\item {\bf Task Categorization.} We begin by categorizing samples collected from a broad spectrum of open-source multimodal datasets into the following domains: vision-centric tasks, STEM, OCR, chart understanding, general VQA, and complex instruction-following. This categorization enables us to tailor domain-specific reward signals that align with the requirements of each task.

	\item {\bf Verifiability-driven Filtering.} To ensure the precision of reward signals during the RL process, we further filter out samples that cannot be easily verified. Specifically, we first classify collected samples from each task domain into objective and subjective types. Objective samples are prioritized for training, as they provide deterministic ground-truth labels that facilitate automated and high-fidelity verification. Furthermore, we filter out samples that are inherently difficult to verify or susceptible to random guessing, such as multiple-choice, true/false, and multi-question prompts.

	\item {\bf Quality Assurance.} To guarantee the correctness of the response used for training, we employ a consensus-based validation mechanism. We utilize a suite of existing models to perform cross-verification on the candidate data. Only samples where a clear consensus is reached are retained.

	\item {\bf Complexity Calibration.} Finally, we conduct offline difficulty grading to optimize the learning curriculum. For each candidate sample, we generate eight independent responses using our most advanced SFT model. Samples where all responses are correct are discarded.\end{itemize}

    \subsubsection{Reward System}
    As summarized in Table~\ref{tab:reward_system}, we design \textit{task-specific} reward signals to provide reliable optimization targets across different tasks. For vision-centric tasks (e.g., grounding, detection, segmentation, counting), rewards are computed by comparing structured predictions against deterministic annotations via standard metrics like IoU or mAP, yielding high-fidelity and fully verifiable feedback. In STEM domains, we enforce a strict output format to facilitate automated answer extraction and apply rule-based validation. For cases where rule-based validation fails, we additionally perform a consistency check with a strong LLM-based verifier. For OCR and chart understanding, we utilize exact string matching and edit distance for spotting and parsing tasks, respectively. To address more general VQA tasks, we use LLM-based consistency checking against reference answers and model responses. Finally, for complex instruction-following tasks, we employ an LLM-as-a-judge with detailed rubrics and reference answers to obtain a scalar reward.
    
    Beyond task-specific rewards, following Youtu-LLM~\citep{lu2026youtullmunlockingnativeagentic}, we incorporate two auxiliary rewards to prevent common failure modes. (i) \textbf{Language Consistency Reward} discourages code-switching and mixed-language outputs by penalizing responses whose dominant language deviates from the pre-specified target language. (ii) \textbf{Repetition Detection Reward} penalizes degenerate generations with excessive n-gram repetition, mitigating cases where the model loops and fails to terminate.

\begin{table*}[t]
\centering
\small
\setlength{\tabcolsep}{6pt}
\renewcommand{\arraystretch}{1.15}
\begin{tabular}{l m{3.2cm} p{9.5cm}}
\hline
Domain & Task & Reward Design \\ 
\hline

\multirow{8}{*}{Vision-Centric}

& \multirow{2}{*}{Grounding}
& IoU-based bounding box reward. \newline
Full score if IoU $\geq 0.5$, otherwise proportional to IoU. \\

& \multirow{2}{*}{Detection}
&  mAP over IoU thresholds $\{0.50{:}0.95\}$. \newline
Predictions of unseen classes are treated as false positives. \\

& \multirow{2}{*}{Counting}
& Exact match for small counts. \newline
Relative error-based soft reward for large counts. \\

& \multirow{2}{*}{Segmentation}
& Class-label IoU for semantic segmentation. \newline
Polygon IoU or AP@50 for referring and instance segmentation. \\

\hline
\multirow{2}{*}{STEM}
& \multirow{2}{*}{Math \& Science} 
& Rule-based answer validation. \newline
LLM-based consistency checking when needed. \\
\hline 
\multirow{3}{*}{OCR\& Chart}   & Spotting & Exact string matching between predicted and ground-truth text.
\\
                       & Parsing & Reward based on edit distance between structured outputs.\\
                       & VQA & Perform consistency checking with LLM.\\
\hline
General VQA            & VQA & Perform consistency checking with LLM.\\
\hline
\multirow{2}{*}{Complex Instructions}    & \multirow{2}{*}{VQA} & Perform LLM-based scoring with detailed rubrics and reference answers. \\ 
\hline 
\end{tabular}
\caption{\textbf{Details of the reward functions used for different tasks}.}
\label{tab:reward_system}
\end{table*}

\subsubsection{Training Recipe}
\textbf{Training Configuration.}
We follow the setup adopted in DAPO~\citep{yu2025dapo}. The maximum context length is set to 32,768 tokens. In each training stage, we sample 6,144 rollouts of the current policy. Policy optimization is performed using a mini-batch size of 1,536, with 4 gradient update steps per training iteration. The learning rate of the policy network is fixed at $1\times10^{-6}$.
To balance training stability and policy exploration, the clipping range is set to $[0.20, 0.24]$, which helps prevent overly aggressive policy updates while encouraging sufficient exploration during training.
In addition, to suppress excessively long generations, we introduce a soft overlong punishment mechanism. Specifically, we analyze the distribution of model output lengths and set the overlong threshold accordingly. 
Following previous work~\citep{qi2025defeating}, we remove the KL penalty term during training and adopt FP16 training. Empirically, these design choices lead to faster convergence and more stable training dynamics.

\textbf{Reward-Variance-Aware Sampling.} Directly applying standard sampling strategies in reinforcement learning often leads to training instability. When advantage values approach zero, the corresponding policy gradients diminish, reducing effective gradient magnitudes and increasing sensitivity to noise. This issue is further amplified by the introduction of soft overlong punishment, which assigns negative rewards to some samples and causes zero-reward samples without length penalties to dominate mini-batch updates. In addition, unlike prior settings with discrete rewards (e.g., binary 0/1 signals), our framework employs continuous reward values. This problem is especially pronounced in vision-centric tasks, where conventional reward-based sampling heuristics retain many samples with highly similar rewards, resulting in insufficient reward discrimination.
To address these issues, we propose a variance-based dynamic sampling strategy. Specifically, for each rollout group $O_q=\{o_1,\ldots,o_G\}$ generated for prompt $q$, we compute the variance of the corresponding rewards and discard groups whose reward variance falls below a predefined threshold. In addition, we require that at least one sample within a rollout group receives a positive reward, which helps avoid misleading optimization signals when the overall reward signal is non-positive and further improves training stability. This strategy effectively removes sample groups with insufficient reward discrimination, improving sample quality and stabilizing policy optimization.

\textbf{Consistent Sampling.}
The mismatch between the training backend and the inference engine has been identified in previous work as a critical factor affecting the effectiveness of reinforcement learning. To mitigate this issue, we follow Youtu-LLM~\citep{lu2026youtullmunlockingnativeagentic} and adopt a consistent sampling strategy to improve the stability of training. For a given prompt $q$, the KL metric $K(q)$ is used to measure the divergence between the training policy and the rollout policy.

\begin{equation}
K(q)=
\mathbb{E}_{a\sim\pi_{\text{rollout}}(\cdot|q)}
\left[
\frac{\pi_{\theta}(a|q)}{\pi_{\text{rollout}}(a|q)}
-1
-\log\frac{\pi_{\theta}(a|q)}{\pi_{\text{rollout}}(a|q)}
\right].
\end{equation}

For each prompt $q$, a rollout group $O_q=\{o_1,\ldots,o_G\}$ generated for the prompt is admitted for training only when the corresponding policy divergence satisfies $K(q)\le \tau_K$. Rollout groups that violate this condition are directly filtered out during the sampling stage, ensuring that the model is optimized exclusively on samples with constrained policy drift.

Based on the above sampling strategies, the final reinforcement learning objective is defined as:

\begin{equation}
\mathcal{J}(\theta)=
\mathbb{E}_{q\sim\mathcal{D},\,\{o_i\}_{i=1}^{G}\sim\pi_{\text{rollout}}(\cdot|q)}
\left[
\frac{1}{\sum_{i=1}^{G}|o_i|}
\sum_{i=1}^{G}\sum_{t=1}^{|o_i|}
\min\!\Big(
r_{i,t}(\theta)\,\hat{A}_{i,t},\;
\mathrm{clip}\!\big(r_{i,t}(\theta),1-\epsilon_{\text{low}},1+\epsilon_{\text{high}}\big)\,\hat{A}_{i,t}
\Big)
\right]
\end{equation}

\begin{equation}
\text{s.t.}\;\;
\max_{1\le i\le G} R(o_i) > 0,\quad
K(q)\le \tau_{K},\quad
\mathrm{Var}\!\left(\{R(o_i)\}_{i=1}^{G}\right)> \tau_{V}.
\end{equation}

%% file: sec/5_evaluation.tex
\section{Evaluation}
We conducted an extensive evaluation of Youtu-VL on a suite of 30 vision-centric and 45 general multimodal benchmarks. To our knowledge, this is the pioneering effort to unify the assessment of a VLM across dozens of distinct tasks, ranging from grounding, detection, classification, counting, segmentation, depth estimation, and pose estimation to visual query answering, OCR, and GUI operations. Addressing the limitations of existing frameworks~\citep{lmms_eval2024,duan2024vlmevalkit,zhang2024lmmsevalrealitycheckevaluation} in handling such diversity, we extended VLMEvalKit~\citep{duan2024vlmevalkit} by integrating additional vision-centric and plain text evaluation tasks, thereby establishing a holistic and unified evaluation pipeline. 
As illustrated in the two bar charts at the bottom of Figure ~\ref{fig:abstract}, we showcase the model's proficiency across various dimensions in comparison to other models. Specifically for the general multimodal benchmarks, we calculated the average score for each category using only those benchmarks where results were available for all three models. Detailed empirical findings for each evaluation benchmark are provided in the following subsections.

\subsection{Vision-Centric Tasks}

Existing benchmarks have inconsistent label set definitions across datasets. To resolve this, we include the benchmark name in the prompt to provide necessary context. Meanwhile, we support manually specifying label sets (e.g., for semantic segmentation). Given that full label sets are often too large for prompt context windows, the name of the benchmark is the main type. For practical inference tasks, users can supply a custom label set to identify target objects or some flexible prompts, supporting both present and absent categories. Evaluation details and prompts are given in Appendix \ref{sec:app_eval_cv}. The evaluation results are given in Table \ref{tab:eval_cv} with experiment analysis below.

\begin{table}[htbp]
\centering
\setcitestyle{numbers, square} 
\setlength{\tabcolsep}{4pt}
\resizebox{\linewidth}{!}{
\begin{tabular}{l|ccc|ccc|cc}
\toprule
 \multirow{3}{*}{\textbf{Benchmarks}} & \multicolumn{3}{c|}{\textbf{General-Purpose VLMs}} & \multicolumn{3}{c|}{\textbf{Vision-Centric VLMs}} & \multicolumn{2}{c}{\textbf{Classic Specialists}}  \\
  & \small{Youtu-VL} & \small{Qwen3-VL}& \small{InternVL-3.5} & \small{~~\dag UFO~~} & \small{~~GiT~~} & \scriptsize{*VisionLLM v2} & \small{*VLM} & \small{*Non-VLM} \\
 & \small{4B (instruct)} & \small{4B (instruct)}& \small{4B} & \small{8B} & \small{756M} & \small{7B} & \small{-} & \small{-} \\
\hline
\textit{\textbf{Visual Grounding}} &  &  &  &  &  &  &  &  \\
RefCOCO val & \textbf{93.6}  & 90.7 & 92.5 & \textbf{91.8} & - & 90.0 & \textbf{92.6 \scriptsize{~\citep{yan2023universal}}} & 90.5 \scriptsize{~\citep{liu2024grounding}} \\
RefCOCO testA & \textbf{95.2}  & 92.2 & 94.3 &  \textbf{94.3} & - & 93.1 & \textbf{94.3 \scriptsize{~\citep{yan2023universal}}} &93.1 \scriptsize{~\citep{liu2024grounding}} \\
RefCOCO testB & \textbf{90.8}  & 86.7 & 88.2 &  \textbf{87.5} & - & 87.1 & \textbf{91.4 \scriptsize{~\citep{yan2023universal}}} & 88.2 \scriptsize{~\citep{liu2024grounding}} \\
RefCOCO+ val & \textbf{90.1}  & 82.9 & 87.6 &  \textbf{86.9} & - & 81.1 & \textbf{88.7 \scriptsize{~\citep{wang2023one}}}& 82.7 \scriptsize{~\citep{liu2024grounding}} \\
RefCOCO+ testA & \textbf{93.9}  & 89.4 & 92.3 &  \textbf{91.3} & - & 87.3 & \textbf{92.2 \scriptsize{~\citep{wang2023one}}}& 88.9 \scriptsize{~\citep{liu2024grounding}} \\
RefCOCO+ testB & \textbf{85.4}  & 75.6 & 81.6 &  \textbf{80.6} & - & 74.5 & \textbf{83.2 \scriptsize{~\citep{wang2023one}}}& 75.9 \scriptsize{~\citep{liu2024grounding}} \\
RefCOCOg val & \textbf{92.2}  & 87.3 & 89.6 & \textbf{87.9} & - & 85.0 & \textbf{89.2 \scriptsize{~\citep{wang2023one}}}& 86.1 \scriptsize{~\citep{liu2024grounding}} \\
RefCOCOg test & \textbf{92.9}  & 87.7 & \textbf{89.3} &  \textbf{88.6} & - & 86.4 & \textbf{89.3 \scriptsize{~\citep{yan2023universal}}} & 87.0 \scriptsize{~\citep{liu2024grounding}} \\
\hline
\textit{\textbf{Object Detection}} &  &  &  &  &  &  &  &   \\
COCO val & \textbf{47.1} & - & - &  48.9 & 46.7 & \textbf{56.7} & \textbf{63.7}~\citep{beit3} & 63.1 \scriptsize{~\citep{zhangdino}} \\
\hline
\textit{\textbf{Semantic Segementation}} &  &  &  &  &  &  &  &  \\
ADE20k & \textbf{54.2} & × & × &  \textbf{54.5} & 47.8 & \textcolor{gray}{52.3} & 38.4 \scriptsize{~\citep{qiu2025refining}} & \textcolor{gray}{\textbf{56.4} \scriptsize{~\citep{cheng2022masked}}} \\
Cityscapes & \textbf{70.4} & × & × &  - & \textbf{61.8} & - & 42.0 \scriptsize{~\citep{lan2024proxyclip}} & \textcolor{gray}{\textbf{83.3} \scriptsize{~\citep{cheng2022masked}}} \\
Context59 & \textbf{60.4} & × & × &  - & \textbf{63.3} & - & \textbf{63.6} \scriptsize{~\citep{qiu2025refining}} & {60.8} \scriptsize{~\citep{carion2025sam}} \\
VOC20 & \textbf{92.5} & × & × &  - & - & - & \textbf{97.1} \scriptsize{~\citep{liu2024open}} & - \\
COCOStuff & \textbf{52.5} & × & × &  30.2 & \textbf{49.1} & - & \textcolor{gray}{39.6 \scriptsize{~\citep{zhou2022extract}}} & \textcolor{gray}{\textbf{45.7} \scriptsize{~\citep{xu2023side}}} \\
\hline
\textit{\textbf{Referring Segementation}} &  &  &  &  &  &  &  & \\
RefCOCO val & \textbf{80.7} & × & × &  \textbf{80.0} & × & 76.6 &  \textbf{80.5} \scriptsize{~\citep{liu2025unipixel}} & 79.3 \scriptsize{~\citep{ouyang2025region}}  \\
RefCOCO testA & \textbf{82.0} & × & × &  \textbf{81.6} & × & 79.3 &  \textbf{82.6} \scriptsize{~\citep{liu2025unipixel}} & 81.2 \scriptsize{~\citep{ouyang2025region}} \\
RefCOCO testB & \textbf{78.4} & × & × &  \textbf{78.1} & × & 74.3 & \textbf{76.9} \scriptsize{~\citep{liu2025unipixel}} & 77.8 \scriptsize{~\citep{ouyang2025region}} \\
RefCOCO+ val & \textbf{76.2} & × & × &  \textbf{76.7} & × & 64.5 & \textbf{74.3} \scriptsize{~\citep{liu2025unipixel}} & 69.5 \scriptsize{~\citep{ouyang2025region}} \\
RefCOCO+ testA & \textbf{79.6} & × & × &  \textbf{79.9} & × & 69.8 & \textbf{78.9} \scriptsize{~\citep{liu2025unipixel}} & 75.6 \scriptsize{~\citep{ouyang2025region}} \\
RefCOCO+ testB & \textbf{71.4} & × & × &  \textbf{72.3} & × & 61.5 & \textbf{68.4} \scriptsize{~\citep{liu2025unipixel}} & 63.0 \scriptsize{~\citep{ouyang2025region}} \\
RefCOCOg val & \textbf{76.5} & × & × &  \textbf{75.5} & × & 70.7 & \textbf{76.3} \scriptsize{~\citep{liu2025unipixel}} & 71.3 \scriptsize{~\citep{ouyang2025region}} \\
RefCOCOg test & \textbf{76.6} & × & × &  \textbf{76.3} & × & 71.2 & \textbf{77.0} \scriptsize{~\citep{liu2025unipixel}} & 72.0 \scriptsize{~\citep{ouyang2025region}} \\
\hline
\textit{\textbf{Depth Estimation}} &  &  &  &  &  &  &  &  \\
NYUv2 & \textbf{90.4} & × & × &  \textcolor{gray}{93.6} & × & × & 86.8 \scriptsize{~\citep{cai2025depthlm}} & \textcolor{gray}{\textbf{98.8} \scriptsize{~\citep{piccinelli2025unidepthv2}}} \\
Cityscapes & \textcolor{gray}{\textbf{92.7}} & × & × &  - & × & × & - & \textcolor{gray}{\textbf{92.1} \scriptsize{~\citep{taghavi2024swinmtl}}} \\
DDAD & \textbf{87.6} & × & × &  - & × & × & 74.7 \scriptsize{~\citep{cai2025depthlm}} & \textbf{88.2} \scriptsize{~\citep{piccinelli2025unidepthv2}} \\
\hline
\textit{\textbf{Human Pose}} &  &  &  &  &  &  &  &  \\
MPII & \textbf{89.1} & × & × &  × & × & - & 89.3\scriptsize{~\citep{wang2024locllm}} & \textbf{93.3}\scriptsize{~\citep{xu2022vitpose}} \\
\hline
\textit{\textbf{Image Classification}} &  &  &  &  &  &  &  &  \\
ImageNet-ReaL & \textbf{89.3} & - & - &  × & × & × &  91.1 \scriptsize{~\citep{fang2024eva}} & \textbf{\textcolor{gray}{91.2}} \scriptsize{~\citep{wortsman2022model}} \\
\hline
\textit{\textbf{Object Counting}} &  &  &  &  &  &  &  &   \\
TallyQA-Simple & \textbf{85.1} & 79.0 & 77.6 & × & × & × &  84.9\scriptsize{~\citep{beyer2024paligemma}} & \textbf{\textcolor{gray}{86.3}} \scriptsize{~\citep{wu2024omni}} \\
TallyQA-Complex & \textbf{74.4} & 64.0 & 66.4 & × & × & × & 72.3 \scriptsize{~\citep{beyer2024paligemma}}  & \textbf{\textcolor{gray}{77.1}} \scriptsize{~\citep{wu2024omni}}\\
CountBench & \textbf{88.6} & 78.4 & 79.4 & × & × & × &  {83.1} \scriptsize{~\citep{beyer2024paligemma}} & \textbf{93.8} \scriptsize{~\citep{carion2025sam}} \\
\bottomrule
\end{tabular}
}
\caption{\textbf{Comparison of Youtu-VL with general-purpose VLMs (standard architecture), vision-centric VLMs (task-specific design involved), and classic specialists (single or certain tasks)}. ``×'' indicates that this capacity is not supported by the model, ``-'' means results are not available or reported. Models with ``*'' use task-specific decoders (e.g., Deform-DETR in VisionLLM v2 ~\citep{wu2024visionllm} beyond the valina VLM architecture). ``\dag'' indicates extra task-specific tokens (e.g., multiple \texttt{<mask>} tokens in UFO ~\citep{tang2025ufo}). Results in \textcolor{gray}{gray} indicate task-specific fine-tuning on a single dataset, which usually brings higher results. Best results in each setting are marked in bold. ``Classic Specialists'' indicates models focusing on a few specific tasks using VLMs (e.g., CLIP~\citep{radford2021learning}) or Non-VLMs (e.g., Mask2Former~\citep{cheng2022masked}, DINO~\citep{zhangdino}, GroundingDINO~\citep{grounding_dino}, UniDepthV2 ~\citep{piccinelli2025unidepthv2}, SAM3~\citep{carion2025sam}). We report the results from the model whose parameters are closest to ours, and general-purpose VLMs indicate their Instruct version. We have compared with additional specialists in Appendix \ref{sec:ext_exp} and Appendix \ref{sec:ext_exp_loc} for reference with their differences.}
\label{tab:eval_cv}
\end{table}

\textbf{Visual Grounding.} We evaluate Youtu-VL on the standard RefCOCO/+/g benchmarks ~\citep{yu2016modeling} to assess its grounding capabilities. Youtu-VL demonstrates outstanding grounding performance, achieving an average score of 91.8\% across all RefCOCO splits. In comparison, the average result is 89.4\% for InternVL-3.5-4B and 91.6\% for the proprietary Seed1.5-VL~\citep{guo2025seed1}. These results demonstrate Youtu-VL's proficiency in visual grounding tasks.

\textbf{Object Detection.}
For object detection, Youtu-VL generates direct textual outputs without extra heads. This approach yields competitive performance on COCO~\citep{lin2014microsoft} using the 2017 {val} split. Specifically, Youtu-VL achieves 47.1\% mAP, while the result of GiT~\citep{wang2024git} is 46.7\%. It also maintains comparable performance to UFO~\citep{tang2025ufo} (48.9\% mAP), despite UFO being a larger, task-aware model that benefits from dense proposals and parallel decoding. Similarly, VisionLLM-v2~\citep{wu2024visionllm} gains benefits from its task-specific decoder with a 56.7\% mAP. In comparison to these methods, Youtu-VL requires no additional training or inference modifications while still achieving competitive performance.

\noindent\textbf{Semantic Segmentation.}
For semantic segmentation, Youtu-VL directly predicts dense predictions from the output logits of vision tokens. This approach demonstrates strong versatility, while other general-purpose VLMs like Qwen3-VL and InternVL-3.5 do not support these dense prediction tasks. In terms of performance, Youtu-VL achieves significant improvements over vision-centric models; for instance, on the ADE20k ~\citep{zhou2017scene} dataset, it attains $54.2$ mIoU, while the results of GiT is 47.8\%. Furthermore, Youtu-VL proves competitive results against the classic specialist method Mask2Former ~\citep{cheng2022masked} and CLIP-based VLMs ~\citep{lan2024proxyclip,qiu2025refining,liu2024open,zhou2022extract}. Youtu-VL also demonstrates strong performance on the COCOStuff ~\citep{caesar2018coco} datasets, without fine-tuning like SAN ~\citep{xu2023side} (52.5\% vs. 45.7\% in mIoU). In addition to quantitative analysis, segmentation also supports open scenarios, such as specifying certain categories and as many segmentation categories as possible. These capabilities are not supported by classic specialists and vision-centric VLMs. We conducted a qualitative analysis and visualization of this in Appendix \ref{sec:examples}.

\noindent\textbf{Referring Expression Segmentation.}
In this task, Youtu-VL employs a grounding-then-segmentation manner to generate precise segmentation masks based on specific textual descriptions. Specifically, we generate the grounding box and draw on the image, then crop to segment the foreground semantics. This operate do not rely on any extra decoder, like SAM or other task-specific heads, while significantly beyond the polygon-based referring expression segmentation. This method yields state-of-the-art performance among comparable models. On the RefCOCO ~\citep{yu2016modeling} val set, Youtu-VL achieves 80.7\% mIoU, while the results are 80.0\%~\citep{tang2025ufo}, 76.6\%~\citep{wu2024visionllm}, 80.5\%~\citep{liu2025unipixel}, and 79.3\%~\citep{ouyang2025region} for other settings. Our method does not rely on predicting extra mask embedding tokens or extra heads, which is simple yet effective. These results prove that a generalist model can achieve pixel-level precision without relying on the extra task-specific tokens or decoders with high-quality results.

\noindent\textbf{Depth Estimation.}
This task is similar to semantic segmentation, where the class names are replaced with the names of quantized bins. Youtu-VL predicts dense depth maps directly from monocular images. Unlike most general-purpose and vision-centric VLMs (e.g., Qwen3-VL ~\citep{Qwen3-VL}, GiT ~\citep{wang2024git}), which lack support for this spatial-related task, Youtu-VL integrates this capability seamlessly. On the NYUv2 ~\citep{silberman2012indoor} dataset, Youtu-VL achieves a $\delta_1$ of 90.4\%, which is slightly lower than the 8B-parameter UFO model ($\delta_1$ at 93.6\%). Because no task-specific fine-tuning is applied with less parameters. In comparison, the performance of another VLM-based method DepthLLM-3B~\citep{cai2025depthlm}, is 86.8\%, which requires inference for each point. In contrast, our model obtains depth information intrinsically without any additional inference. Youtu-VL demonstrates strong performance on the DDAD ~\citep{guizilini20203d} benchmark, a $\delta_1$ of 87.6\%, while DepthLLM-3B's result is 74.7\%. Our result is close to the specialist model UniDepth-v2~\citep{piccinelli2025unidepthv2} (88.2\%). This establishes Youtu-VL as a comprehensive model capable of understanding 3D information alongside 2D semantic tasks.

\noindent\textbf{Human Pose Estimation.} Youtu-VL realizes human pose estimation by directly regressing keypoint coordinates within a single generative framework, avoiding the task-specific heads of specialist methods. On MPII~\citep{andriluka20142d}, Youtu-VL achieves 89.1\% (PCKh@0.5) following the evaluation protocol of the specialist model ViTPose~\cite{xu2022vitpose} (93.3\%). This result is close to LocLLM~\citep{wang2024locllm} (89.3\% on the specific keypoints). Despite the comparable performance, Youtu-VL is a versatile Vision-Language Model (VLM) that operates without the need for explicit pose estimation priors. Detailed evaluation information can be found in Appendix \ref{sec:app_eval_cv}.

\textbf{Image Classification.}
In image classification tasks, Youtu-VL is prompted to generate a phrase identifying the primary object category within an image. On ImageNet-ReaL \citep{beyer2020we} (ImageNet-1k with corrected ground-truth), Youtu-VL achieves a Top-1 accuracy of 89.3\%. This performance approaches the level of leading specialist models, demonstrating the model's efficiency in object recognition.

\textbf{Object Counting.}
This task is evaluated by directly outputting numerical counts of objects without any options. In CountBench~\citep{paiss2023teaching}, Youtu-VL achieves an accuracy of 88.6\%, while the result of Qwen3-VL-4B is 78.4\% (answer prediction without option). In simple object counting scenarios (TallyQA-simple)~\citep{acharya2019tallyqa}, Youtu-VL reaches 85.1\% accuracy, performing on par with the generalist model Omni-SMoLA (86.3\%). For complex counting tasks (TallyQA-complex), Youtu-VL secures a leading score of 74.4\%, while the performance of the PaliGemma~\citep{beyer2024paligemma} is 72.3\%. This result is close to the specialist model Omni-SMoLA~\citep{wu2024omni} (77.1\%). Benefiting from extensive training on localization data, Youtu-VL exhibits superior object counting performance.

\subsection{General Multimodal Tasks}
In this section, we systematically evaluate Youtu-VL across several dimensions, including general visual question answering (VQA), multimodal reasoning, real-world and multi-image understanding, OCR and document understanding, hallucination suppression, and text-centric tasks. For benchmarks that are not officially reported for Qwen3-VL, we additionally run evaluations using its publicly released checkpoints to enable a more comprehensive comparison. In general, our model consistently outperforms or maintains parity with leading models of comparable size across a wide range of multimodal tasks.

\newcommand{\first}[1]{\textbf{#1}}
\newcommand{\second}[1]{{#1}}
\begin{table}[htbp]
\centering
\small
\renewcommand{\arraystretch}{0.95} 
\setlength{\tabcolsep}{12pt}
{
\begin{tabular}{l|c|ccc}
\toprule
 \multirow{2}{*}{\textbf{Benchmarks}} & 
Qwen3-VL&InternVL-3.5&Qwen3-VL&Youtu-VL\\
& 8B (instruct)&4B&4B (instruct)&4B (instruct)\\
\hline
\textit{\textbf{General VQA}}  &  &  &  \\
MMBench$_{CN}$ & {84.7} &--& \second{83.5} & \first{83.6} \\  
MMBench$_{EN}$ & {84.5}&80.3& \first{83.9} & \first{83.9} \\  
MMStar             &{70.9} & 65.0&\second{69.8} & \first{71.1} \\  
MME (/2800)&--      & 2272 & ~~\second{2309}$^*$ & \first{2384}\\  
CVBench$_{2d}$&--  & -- & ~~\second{79.1}$^*$ & \first{80.4} \\
CVBench$_{3d}$&-- & -- & ~~\second{92.4}$^*$ & \first{93.0} \\
ScienceQA$_{\text{val}}$&-- & -- & ~~\second{94.7}$^*$ & \first{97.0} \\
SEEDBench$_{IMG}$&-- &-- & ~~\first{77.0}$^*$ & \second{76.9} \\
SEEDBench2&--& --   & ~~\first{75.9}$^*$ & \second{74.5} \\
{MMVet}& --&--& ~~\first{68.3}$^*$ & \second{64.6} \\  
\hline
\textit{\textbf{Multimodal Reasoning \& Math}}  &  &  &  \\
VisuLogic        &22.5 & --&\second{19.0} & \first{25.7} \\
MMMU$_{val}$     & {69.6} &\second{66.6} & \first{67.4} & {61.1} \\  
MMMU-Pro      &{55.9}  &--& \first{53.2} & \second{43.0} \\  
CMMMU$_{val}$     &--   &-- & ~~\first{54.6}$^*$ & \second{52.6} \\
MathVista$_{mini}$&{77.2} & \first{77.1} & {73.7} & \second{76.5} \\ 
MathVerse$_{mini}$&{62.1}& 45.8& \second{46.8} & \first{56.5} \\  
LogicVista  &{55.3}  &41.8& \first{53.2} & \second{52.4} \\
VLMsAreBlind &{74.0}& --& \second{71.9} & \first{88.9} \\
\hline
\textit{\textbf{Hallucination}}  &  &  &  \\
{HallusionBench}&{61.1} &44.8 & \second{57.6} &\first{59.1} \\
CRPE$_{\text{exist}}$   &--     & --  & ~~\second{95.6}$^*$ & \first{96.9} \\
CRPE$_{\text{relation}}$  &--  &\first{75.0}& ~~{71.0}$^*$ & \second{72.2} \\
POPE    &--   &\second{88.9}   & ~~\first{89.3}$^*$ & 86.4 \\
\hline
\textit{\textbf{OCR-related Understanding}}  &  &  &  \\
AI2D$_{\text{test}}$&{85.7}& 82.6&\second{84.1} & \first{85.6} \\ 
InfoVQA$_{\text{val}}$          &{83.1}& 78.0&\first{80.3} & \second{79.1} \\ 
TextVQA$_{\text{val}}$          &-- & 77.9&~~\first{80.8}$^*$ & \second{79.6} \\  
DocVQA$_{\text{val}}$           &{96.1} & 92.4&\first{95.3} & \second{94.4} \\  
ChartQA$_{\text{test}}$ &{89.6} &\first{86.0}& {84.6} & \second{85.3} \\ 
OCRBench &{896}&\second{822}& \first{881} & 813 \\ 
SEEDBench2$_{\text{Plus}}$&-- &69.4& ~~\first{71.5}$^*$ & \second{71.3} \\ 
CharXiv$_{\text{DQ}}$&{83.0} &71.1& \second{76.2} & \first{79.4} \\  
CharXiv$_{\text{RQ}}$&{46.4}&39.6& \second{39.7} & \first{43.8} \\  
\hline

\textit{\textbf{Multi-image \& Real-world}}  &  &  &  \\BLINK&{69.1}  &58.1& \first{65.8} & \second{64.3} \\  
RealWorldQA &{71.5}  & 66.3&\second{70.9} & \first{74.6} \\  
MMERealWorld$_{EN}$ &--&--&~~\first{63.0}$^*$ & \second{61.5} \\ 
MMERealWorld$_{CN}$ &-- &59.8 & ~~\second{61.3}$^*$ & \first{63.5} \\
\hline
\textit{\textbf{GUI Agent}}  &  &  &  \\
ScreenSpot Pro &54.6& - & 59.5 & \first{59.6} \\
OSWorld         &33.9& -- & \second{26.2} & \first{38.8} \\ 
 \hline
\textit{\textbf{Text-Centric}}  &  &  &  \\
MMLU-Pro&71.6&--&\first{67.1}&\second{56.5}\\
MMLU-Redux&84.9&--&\first{81.5}&\second{76.8}\\
C-Eval&--&71.9&\first{76.5}&\second{69.1}\\
MuSR&--&--&\second{46.6}&\first{58.3}\\
IFEval&83.7&--&\first{82.3}&\second{76.9}\\
DROP(F1)&--&--&\first{85.0}&\second{79.3}\\
BBH&--&--&\first{84.8}&\second{71.9}\\
GPQA-Diamond&--&--&\first{42.9}&\second{39.8}\\
\bottomrule
\end{tabular}
}
\caption{\textbf{Comparison of Youtu-VL with the state-of-the-art VLMs on various multimodal benchmarks}. All numbers are reported or reproduced under the official evaluation protocols. ``--'' indicates that the corresponding number is not available or not reported. The superscript ``*'' indicates that the result was not reported and has been reproduced using our evaluation system.}
\label{tab:multimodal_tasks}
\end{table}

\textbf{General Visual Question Answering.} To evaluate the general-purpose VQA capabilities of Youtu-VL, we conduct extensive experiments on a comprehensive suite of benchmarks, including MMBench\_V11 (EN/CN)~\citep{liu2023mmbench}, MMStar~\citep{chen2024we}, MME~\citep{fu2023mme}, CVBench (2D/3D)~\citep{zhu2025cvbench}, ScienceQA~\citep{lu2022learn_scienceqa}, the SEED-Bench series~\citep{li2023seedbench,li2024seed,li2024seed2plus}, and MMVet~\citep{yu2024mm}, as detailed in Table~\ref{tab:multimodal_tasks}. 
Across these benchmarks, Youtu-VL demonstrates consistently strong performance on general visual understanding tasks that require joint perception and lightweight reasoning. On MMBench, the model achieves scores of 83.9 on the English split and 83.6 on the Chinese split, indicating stable multilingual visual–language alignment. On MMStar and MME, which emphasize broad visual reasoning and robustness across diverse question types, Youtu-VL obtains comparatively high scores, reflecting effective integration of visual perception and knowledge grounding.
Furthermore, the model delivers impressive performance on CVBench-2D and CVBench-3D, suggesting solid capability in handling structured visual representations and core vision-related reasoning. Performance on MMVet is comparatively lower, indicating room for improvement on more complex instruction-style VQA scenarios that demand stronger compositional reasoning and response calibration.

\textbf{Multimodal Reasoning \& Math} We evaluate the models across a broad spectrum of multimodal reasoning benchmarks, including VLMs Are Blind~\citep{rahmanzadehgervi2025visionlanguagemodelsblind}, VisuLogic~\citep{xu2025visulogicbenchmarkevaluatingvisual}, the MMMU family (MMMU~\citep{yue2024mmmu}, MMMU-Pro~\citep{mmmupro}, CMMMU~\citep{zhang2024cmmmu}), MathVista~\citep{mathvista}, MathVerse~\citep{zhang2024mathverse}, and LogicVista~\citep{xiao2024logicvista}. These benchmarks collectively assess image-grounded logic, abstract reasoning, mathematical understanding, and knowledge-intensive multimodal problem-solving.

As shown in Table~\ref{tab:multimodal_tasks}, Youtu-VL achieves strong results on several reasoning-oriented benchmarks. Notably, it achieves scores of 25.7\% on VisuLogic and 88.9\% on VLMs Are Blind, demonstrating robust visual reasoning capabilities. Together with its competitive performance on LogicVista, these results suggest that Youtu-VL is effective at handling logical relations, rule-based reasoning, and image-dependent inference. On math-related benchmarks, Youtu-VL achieves a score of 56.5\% on MathVerse and maintains stable performance on MathVista, demonstrating formidable multimodal mathematical reasoning capabilities.

On large-scale academic benchmarks such as MMMU, MMMU-Pro, and CMMMU, which resemble multidisciplinary university-level examinations and emphasize long-tail domain knowledge and long-context language reasoning, Youtu-VL attains mid-to-high score levels (e.g., around 60 on MMMU). While these scores are generally competitive, there remains a modest performance gap on the most knowledge-intensive and context-heavy tasks. This suggests that, beyond its solid multimodal perception and reasoning, future improvements could focus on expanding specialized domain coverage and further strengthening long-context reasoning stability. Overall, the current results show that Youtu-VL already provides advanced and reliable multimodal reasoning capabilities, particularly in scenarios demanding genuine image-grounded inference, abstract logical judgment, and structured visual–symbolic reasoning.

\textbf{Hallucination.} We evaluate hallucination suppression and instruction alignment using HallusionBench~\citep{hallusion_bench}, POPE~\citep{li2023evaluatingpope}, and CRPE~\citep{wang2023allseeing,wang2024allseeing_v2}, benchmarks specifically designed to probe adversarial image--question inconsistency as well as fine-grained object existence and relation judgment. 
As shown in Table~\ref{tab:multimodal_tasks}, Youtu-VL demonstrates a significantly reduced propensity for hallucination. On HallusionBench, which deliberately presents scenarios where questions contradict image content, Youtu-VL achieves 59.1\%. This improvement is particularly pronounced in cases where the question suggests the presence of objects clearly absent from the image, indicating that Youtu-VL prioritizes verification against visual evidence over reliance on linguistic priors. 
On CRPE$_{\text{exist}}$ and CRPE$_{\text{relation}}$, Youtu-VL delivers commendable performance, with residual errors primarily stemming from intrinsically ambiguous boundary cases, such as minute objects or severe occlusions. 
Overall, Youtu-VL maintains competitive instruction-following capabilities while exhibiting superior resistance to hallucination and stricter image-grounded alignment, favoring cautious responses when visual evidence is insufficient or contradictory. We attribute this robustness to the integration of effective visual supervision, which compels the model to ground its predictions strictly in visual information.

\textbf{OCR-related Understanding \& Document QA.} We evaluate fine-grained OCR capabilities and document-level question answering on AI2D~\citep{Kembhavi2016AI2D}, InfoVQA~\citep{Mathew2021InfographicVQA}, TextVQA~\citep{textvqa}, DocVQA~\citep{docvqa}, ChartQA~\citep{masry2022chartqa}, OCRBench~\citep{liu2024ocrbenchhiddenmysteryocr}, SEEDBench2$_{\text{Plus}}$~\citep{li2024seed2plus}, and the CharXiv description and reasoning subsets~\citep{wang2024charxiv}. These benchmarks cover a spectrum of tasks ranging from dense text recognition to structured document analysis and chart-centric vision--language integration. 

As shown in Table~\ref{tab:multimodal_tasks}, Youtu-VL delivers strong performance on OCR-centric benchmarks, especially in settings that require higher-level semantic understanding and reasoning over recognized text. In particular, it achieves a score of about 79\% on CharXiv$_{DQ}$ and around 44\% on CharXiv$_{RQ}$, providing clear gains of several points over comparable 4B-level systems on both descriptive and reasoning-style chart understanding. These tasks demand more than basic text extraction; they require structured interpretation of scientific charts and figures, with joint reasoning over axes, legends, curves, and surrounding textual explanations. Qualitative inspection further suggests that Youtu-VL tends to capture document layout and visual–textual relations in a more globally consistent way, often producing explanations that are both complete and logically organized.

On widely used OCR-heavy benchmarks such as TextVQA, DocVQA, ChartQA, OCRBench, SEEDBench2$_{Plus}$, AI2D, and InfoVQA, Youtu-VL maintains a generally competitive score profile, with performance that is stable across diverse data sources and task formats. While there remain a few datasets on which its scores are slightly behind the very best reported numbers, the overall results indicate that Youtu-VL is already effective at handling information-dense visual contexts and at bridging low-level text recognition with higher-level document and chart intelligence.

\textbf{Multi-image \& Real-world Understanding.}
We assess multi-image reasoning and real-world scene understanding on BLINK~\citep{fu2024blink}, RealWorldQA~\citep{realworldqa2024}, and MMERealWorld (EN and CN)~\citep{mme-realworld}. As summarized in Table~\ref{tab:multimodal_tasks}, Youtu-VL delivers competitive performance across these benchmarks. 
Specifically, Youtu-VL delivers strong performance on real-world and multi-image benchmarks. On RealWorldQA, it reaches a score of 74.6, indicating robust competence in handling high-resolution photographs with small objects, cluttered backgrounds, and complex object–scene interactions. On MMERealWorld, Youtu-VL attains solid scores in both English and Chinese settings, and its relatively higher score on the Chinese split is particularly notable given that the model is trained primarily on English data. This cross-lingual robustness suggests that the underlying visual perception is largely language-agnostic, benefiting from effective visual supervision that helps the model focus on fine-grained details while preserving global scene context.
On BLINK, which emphasizes fundamental multi-image matching and comparison, Youtu-VL maintains competitive score levels, showing that it can align and compare information across images in a consistent manner. Overall, these results indicate that Youtu-VL provides strong and reliable real-world visual understanding while preserving solid multi-image reasoning abilities, with remaining gaps mainly appearing in more context-heavy, conversational settings rather than in core visual perception.

\textbf{GUI Agent.}
We evaluated the GUI agentic performance of Youtu-VL using two primary benchmarks: ScreenSpot Pro ~\citep{li2025screenspot} and OSWorld ~\citep{xie2024osworld}. ScreenSpot Pro serves as a static, single-turn grounding benchmark designed to assess the model’s fundamental capacity for instruction parsing and precise coordinate prediction. Its inclusion of diverse operating systems and specialized cross-platform applications ensures a comprehensive measurement of the model’s perceptual generalization. In contrast, OSWorld provides a holistic evaluation of dynamic, closed-loop interaction. By utilizing a live Ubuntu sandbox with authentic software installations, OSWorld requires the agent to execute multi-turn reasoning and adapt to real-time state transitions. This environment accurately reflects "computer use" capabilities by testing the model's ability to maintain long-horizon goals within a functional operating system. 
Our empirical results demonstrate that Youtu-VL achieves SoTA performance across both benchmarks, most notably on OSWorld, where it achieved a success rate of 38.8. These metrics suggest that Youtu-VL possesses superior grounding-to-action mapping, effectively translating visual perceptions into valid system sequences rather than merely identifying UI elements. The substantial performance improvement highlights the model's resilience against ``error compounding''—a common failure mode in multi-turn tasks where a single misstep leads to an unrecoverable state. Furthermore, these results demonstrate that Youtu-VL can handle complex, real-world computer use scenarios with greater efficiency and robustness than baseline models.

\textbf{Text-Centric Tasks}
The final section of Table~\ref{tab:multimodal_tasks} presents the evaluation results of Youtu-VL on pure text benchmarks. Given that Youtu-VL is architected with a primary focus on multimodal capabilities—particularly vision-centric tasks—a discernible performance gap exists when compared to state-of-the-art models. Specifically, on IFEval ~\citep{ifeval}, which assesses instruction-following capabilities, Youtu-VL achieves a score of 76.9, demonstrating a competent ability to adhere to complex constraints despite not reaching the ceiling of top-tier language models. On MMLU-Redux ~\citep{mmluredux}, the model attains a score of 76.8, indicating a robust foundation in general knowledge and reasoning; however, its performance on GPQA-Diamond ~\citep{gpqa} (39.8) highlights remaining challenges in handling expert-level domain reasoning. In summary, while Youtu-VL maintains a functional baseline for textual interaction, there remains significant headroom for improvement in language capabilities, which serves as a key objective for future iterations.

%% file: sec/6_conclusion.tex
\section{Conclusion}

\textbf{Discussion.}
In this work, we present \textbf{Youtu-VL}, a framework that fundamentally reshapes the optimization landscape for Vision-Language Models. By introducing the \textit{Vision-Language Unified Autoregressive Supervision (VLUAS)} paradigm, we effectively mitigate the text-dominant optimization bias inherent in traditional architectures.  Transitioning from a static ``vision-as-input'' dependency to a generative ``vision-as-target'' objective, Youtu-VL incentivizes the model to simultaneously predict fine-grained visual details and high-level linguistic semantics, thereby bridging the gap between coarse understanding and dense perception. Crucially, our results demonstrate that this unified objective enables a standard VLM architecture to natively execute diverse vision-centric tasks without reliance on task-specific decoders or auxiliary heads. This illustrates that high-fidelity sensory perception can be modeled end-to-end within a single generalist transformer. Our findings mark a critical inflection point in the evolution of VLMs: moving beyond mere cross-modal alignment toward structural unification. We contend that true multimodal intelligence emerges not from stacking specialist modules, but from the synergistic integration of perception and reasoning. Youtu-VL thus serves not merely as a competitive baseline, but as a foundational blueprint for the next generation of Generalist Visual Agents.

\textbf{Limitation.}
While Youtu-VL achieves competitive performance across various benchmarks, it faces certain systemic challenges. First, the current visual representation granularity remains a bottleneck for high-precision tasks on low-resolution inputs. Second, the model’s performance in specialized geometry-aware tasks (e.g., depth and pose estimation) is still constrained by its sensitivity to sensor intrinsics and the diversity of training distributions, limiting its zero-shot robustness in out-of-distribution environments. Furthermore, while Youtu-VL excels in general perception, its high-level cognitive abilities, particularly in complex mathematical reasoning and dense knowledge retrieval, require further optimization. Addressing these foundational constraints remains a key objective for future development of the Youtu-VL framework.

%% file: sec/7_contributions.tex
\section*{Contributions and Acknowledgments}

We would like to express our sincere gratitude to all contributors, including those not listed in the paper, for their invaluable support and efforts. \textbf{The contributors within each group are listed in no particular order.}

\textbf{Core Contributors}\\
Zhixiang Wei\textsuperscript \quad
Yi Li\textsuperscript \quad
Zhehan Kan\textsuperscript \quad
Xinghua Jiang\textsuperscript \quad
Zuwei Long\textsuperscript \quad
Shifeng Liu\textsuperscript \quad
Hongze Shen\textsuperscript \quad
Wei Liu\textsuperscript \quad 
\\ Xiaoyu Tan\textsuperscript \quad
Haojia Lin\textsuperscript \quad
Yubo Zhu\textsuperscript \quad
Qianyu Li\textsuperscript \quad
Di Yin\textsuperscript \quad
Haoyu Cao\textsuperscript \quad
Weibo Gu\textsuperscript \quad
Xin Li$^\spadesuit$\textsuperscript \quad 
Yinsong Liu\textsuperscript \quad
\\Deqiang Jiang\textsuperscript \quad 
Xing Sun$^\dagger$\textsuperscript \quad 
Yunsheng Wu\textsuperscript \quad
\noindent\let\thefootnote\relax\footnotetext{$^\spadesuit$Project Lead}
\noindent\let\thefootnote\relax\footnotetext{$^\dagger$Corresponding author: winfredsun@tencent.com}

\textbf{Contributors}\\
Mingkong Tang\textsuperscript \quad
Shuangyin Liu\textsuperscript \quad
Lexiang Tang\textsuperscript \quad
Haodong Lin\textsuperscript \quad
Junru Lu\textsuperscript \quad
Jiarui Qin\textsuperscript \quad
Lingfeng Qiao\textsuperscript \quad
\\ Ruizhi Qiao\textsuperscript \quad
Bo Ke\textsuperscript \quad
Jianfeng He\textsuperscript \quad
Ke Li\textsuperscript \quad
Yangning Li\textsuperscript \quad
Yunhang Shen\textsuperscript \quad
Mengdan Zhang\textsuperscript \quad
Peixian Chen\textsuperscript \quad
\\ Kun Yin\textsuperscript \quad
Bing Liu\textsuperscript \quad
Yunfei Wu\textsuperscript \quad
Huang Chen\textsuperscript \quad
Zhongpeng Cai\textsuperscript \quad
Xiaotian Li\textsuperscript \quad
Youtu-VL Team\textsuperscript \quad

%% file: sec/b_evaluation_details.tex
\section{Evaluation Details}
\subsection{Vision-Centric Benchmarks and Prompts}
\label{sec:app_eval_cv}
In order to ensure better reproducibility, we have provided the evaluation details for the benchmarks of vision-centric tasks, along with specific prompts. All results were tested using a zero-shot approach, without utilizing few-shot examples, without the thinking mode (e.g., CoT).

\subsubsection{Visual Grounding}
\textbf{RefCOCO series~\citep{yu2016modeling}.} For visual grounding, we require the model to directly localize the queried object by predicting its bounding box. The specific prompt is as follows:

\noindent\fbox{%
    \begin{minipage}{\dimexpr\linewidth-2\fboxsep-2\fboxrule}
        \ttfamily 
        <image>\\
        Question: \\
        Please provide the bounding box coordinate of the region this sentence describes: the person bottom left\\
        Answer: <box><x\_155><y\_154><x\_221><y\_206></box>
    \end{minipage}%
}

\textbf{Visual Grounding for the Open World.}
Except for specific benchmarks, Youtu-VL also supports various open-scene prompts and objectives. We support some flexible prompts and Chinese prompts. For visual grounding, we provide some prompt examples for reference, and their Chinese versions are also suggested. \{keyword\} indicates the target needed to ground, which is usually a word or phrase.

\noindent\fbox{%
    \begin{minipage}{\dimexpr\linewidth-2\fboxsep-2\fboxrule}
        \ttfamily
Can you return the bounding box for "\{keyword\}" in this image? \\
Output the bounding box for "\{keyword\}" in this image. \\
Perform object grounding for "\{keyword\}" with bounding box. \\
Can you return the bounding box for \{keyword\} in this image? \\
Output the bounding box for \{keyword\} in this image. \\
Perform object grounding for \{keyword\} with bounding box. \\
Can you return the bounding box for \{keyword\} in this image? \\
Output the bounding box for \{keyword\} in this image. \\
Perform object grounding for \{keyword\} with bounding box. \\
Please provide the bounding box coordinate of the region this sentence describes: \{keyword\}. \\
Please get the bounding box coordinate for the target: \{keyword\}. \\
Get bounding box for \{keyword\}. \\
Identify the bounding box for "\{keyword\}" in this image. \\
Locate and return the bounding box coordinates of \{keyword\}. \\
Find the bounding box around the object described as "\{keyword\}". \\
Provide the coordinates of the bounding box for \{keyword\}. \\
Show me the bounding box for the object labeled "\{keyword\}". \\
Draw the bounding box for \{keyword\} and provide its coordinates. \\
Extract the bounding box for "\{keyword\}" from the image. \\
Where is the bounding box for \{keyword\} located in this image? \\
Mark the bounding box surrounding \{keyword\} and return its position. \\
Return the coordinates of the bounding box that encloses \{keyword\}. \\
Give me the bounding box details for the item described as "\{keyword\}". \\
Please highlight the bounding box of \{keyword\} and share the coordinates. \\
Get the spatial bounding box for the object "\{keyword\}". \\
Locate \{keyword\} in the image and provide its bounding box. \\
Identify the region bounding box corresponding to \{keyword\}. \\
Provide bounding box information for the entity named "\{keyword\}". \\
Find and output the bounding box for \{keyword\} in this picture. \\
Output the spatial coordinates of the bounding box for \{keyword\}. \\
Can you specify the bounding box around \{keyword\} in the image? \\
Please return the bounding box of the object referred to as "\{keyword\}". \\
Find the object "\{keyword\}" and give me its bounding box coordinates. \\

    \end{minipage}%
}

\subsubsection{Object Detection}

\textbf{COCO val~\citep{lin2014microsoft}.} For evaluation on COCO, we prompt the model to produce detections using the COCO label space, without specifying any particular category in advance. For detections generated via a naïve autoregressive output format, we parse each prediction into an object category and a bounding box. For each predicted box, when computing mAP, we set its confidence score to be the box area. Because our model is trained on higher-resolution images, our main results are obtained by aggregating predictions from multiple upscaled versions of each COCO image, specifically at 1×, 2×, 3×, and 4×. Bounding boxes predicted at multiple scales are merged using non-maximum suppression (NMS) with an IoU threshold of 0.7. The specific prompt for each image is as follows:

\noindent\fbox{%
    \begin{minipage}{\dimexpr\linewidth-2\fboxsep-2\fboxrule}
        \ttfamily
        <image>\\
        Question: \\
        Detect all objects in the provided image.\\
        Answer: <ref>spoon</ref><box><x\_87><y\_103><x\_929><y\_934></box><ref>bowl</ref>
        <box><x\_85><y\_1409><x\_887><y\_2094></box><box><x\_84><y\_97><x\_2073><y\_1466></box>
        \ldots
        <ref>dining table</ref><box><x\_89><y\_105><x\_2069><y\_2082></box>
    \end{minipage}%
}

\textbf{Object Detection for the Open World.}
The difference between object detection and visual grounding lies in the fact that grounding supports the localization of a single object, while detection supports multiple objects of a single category, as well as multiple objects of multiple categories. We perform detection for given categories, including combinations of single/multiple categories that may or may not be present in the image. \{keyword\} is a series of categories, such as ``dog'' or ``dog, cat, person, tree''. The specific prompts are as follows, and we support their Chinese versions:

\noindent\fbox{%
    \begin{minipage}{\dimexpr\linewidth-2\fboxsep-2\fboxrule}
Please detect \{keyword\} in the image. \\
Please detect “\{keyword\}” in the image. \\
Please detect: \{keyword\} \\
Detect and highlight \{keyword\} in the picture. \\
Can you detect \{keyword\} in this image? \\
Can you detect “\{keyword\}”? \\
Detect objects from: \{keyword\}. \\
Detect all \{keyword\} visible in the photo. \\
Detect all “\{keyword\}” in the photo. \\
Detect all objects in these classes: \{keyword\} \\
Perform detection for “\{keyword\}” \\
Run object detection for \{keyword\} in this picture. \\
Apply detection: \{keyword\}. \\
Use image detection to locate \{keyword\}. \\
Carry out detection for all \{keyword\}. \\
Carry out detection for “\{keyword\}”. \\
Perform detection for “\{keyword\}”. \\
Apply detection: \{keyword\}. \\
Run detection for any \{keyword\} visible. \\
Execute detection for classes: \{keyword\}. \\
Detect \{keyword\} in this image. \\
Please perform detection on \{keyword\} shown. \\
Run detection algorithms to find \{keyword\}. \\
Detect any visible \{keyword\} in this photo. \\
Detect the presence of \{keyword\} in the image. \\
Find and detect \{keyword\} within the picture. \\
Detect all objects labeled as \{keyword\}. \\
Perform a detailed detection for \{keyword\}. \\
Detect \{keyword\} and outline them in the image. \\
Locate and detect “\{keyword\}” precisely. \\
Run a detection pass for \{keyword\} here.
    \end{minipage}%
}

If no category keywords are entered, we also support detecting the main categories in the image. This method can be referred to as 'detect anything' to some extent. However, there may be issues like missing less prominent categories. We will continue to optimize this capability to enhance the generalization ability in open-world scenarios. Specific prompts include, but are not limited to:

\noindent\fbox{%
    \begin{minipage}{\dimexpr\linewidth-2\fboxsep-2\fboxrule}
Detect all object categories present in this picture. \\
Run object detection and output all found objects. \\
Detect objects across the entire image area. \\
Please detect every object and report the results. \\
Run a full object detection to capture all items. \\
Detect all objects in the frame and list them. \\
Apply object detection and return all object labels. \\
Detect and annotate all objects you can see. \\
Identify all objects present and provide detection data. \\
Run object detection to find every object in view. \\
Detect all objects in the picture and summarize them. \\
Perform object detection over the entire image. \\
Identify all items in this image using object detection. \\
Detect all visible objects and provide their labels. \\
Run object detection to detect all objects shown. \\
Detect all objects within this image and list categories. \\
Please perform object detection and list all findings. \\
Detect all objects in this image without specifying extra constraints. \\
Detect all. \\
Detect it. \\
Detect all objects. \\
Run object detection. \\
Return all object categories found. \\
Detect all objects with locations. \\
Detect objects. \\
Show all detected objects. \\
Run object detection. \\
Return all detections. \\
Enumerate all objects detected. \\
Detect all object categories present. \\
Output all detected objects. \\
Perform detection. \\
Detect all objects shown.
    \end{minipage}%
}

\subsubsection{Image Classification}

\textbf{ImageNet-ReaL~\citep{beyer2020we}.} During evaluation, we instruct the model to output a single word or short phrase describing the category of the dominant object in the image. A prediction is counted as correct if it matches the ground-truth label. The specific prompt is as follows:

\noindent\fbox{%
    \begin{minipage}{\dimexpr\linewidth-2\fboxsep-2\fboxrule}
        \ttfamily 
        <image>\\
        Question: \\
        What is the category of the primary object in this image? Answer the question with a single word or phrase.\\
        Answer: puck
    \end{minipage}%
}

\subsubsection{Object Counting}

\textbf{TallyQA~\citep{acharya2019tallyqa} \& CountBench~\citep{paiss2023teaching}.} For the counting evaluation, we require the model to output a single numeric value corresponding to the number of queried objects. A prediction is considered correct if and only if this number exactly matches the ground-truth count. For evaluation, the input image is upsampled by a factor of 2 or 2.5. The specific prompt is as follows:

\noindent\fbox{%
    \begin{minipage}{\dimexpr\linewidth-2\fboxsep-2\fboxrule}
        \ttfamily 
        <image>\\
        Question: \\
        How many aum symbols are there in the image? Output the final answer as a single number.\\
        Answer: 9
    \end{minipage}%
}

\subsubsection{Semantic Segmentation}

\textbf{ADE20K~\citep{zhou2017scene}.} In our testing of the ADE20K dataset (test), we employed an open vocabulary method. The prompts were constructed using the format "Segment: class1, class2, class3, …" where the class names are all lowercase English terms representing the various categories. To assist the model in querying each class individually without fitting to specific instructions, we randomly \emph{shuffled} the class names for each image. Given that the patch size undergoes significant downsampling after token merging (to 32), and considering the small resolution of images in this benchmark, we scaled each image up by four times its original resolution for testing. The results for the detected categories are formatted as <ref>class1</ref><ref>class2</ref><ref>class3</ref>, making them easier to parse. The pixel-level outputs are listed at the end as <mask>RLE string</mask>, with the numbering corresponding to the order of the class names as they appear. ``RLE string'' indicates the pixel labels compressed by Run-Length Encoding. The length of the 1D mask string is the same as the pixel number. Users need to reshape it to the 2D raw image size for further usage. For evaluation without background, a softmax with temperature 0.2 and DenseCRF~\citep{krahenbuhl2011efficient} is applied to the logits to refine the pixel-level predictions. This operation is applied to all tasks related to semantic segmentation during evaluation, and is optional in real practice. Here’s a specific example:

\noindent\fbox{%
    \begin{minipage}{\dimexpr\linewidth-2\fboxsep-2\fboxrule}
        \ttfamily
        <image>\\
        Question: \\
        Segment: tree, towel, tank, armchair, refrigerator, countertop, blanket, railing, hood, bathtub, radiator, chandelier, canopy, arcade machine, trade name, case, wardrobe, oven, sidewalk, toilet, chair, wall, door, land, fireplace, windowpane, plant, waterfall, swimming pool, computer, animal, apparel, bannister, poster, flag, step, chest of drawers, box, book, light, tower, ceiling, bulletin board, palm, bicycle, barrel, earth, buffet, person, bus, stairs, stove, stool, basket, washer, rug, monitor, stairway, cushion, pot, minibike, awning, desk, grass, clock, runway, sea, bookcase, sky, food, hovel, seat, van, road, fountain, streetlight, coffee table, painting, ball, car, bed , cradle, airplane, kitchen island, house, ottoman, column, rock, blind, sconce, fan, escalator, lake, boat, booth, stage, sofa, base, swivel chair, mirror, pole, shower, conveyer belt, table, bench, tray, glass, lamp, dishwasher, crt screen, plaything, pool table, water, screen door, truck, bridge, pillow, building, ship, signboard, traffic light, microwave, screen, path, vase, mountain, pier, field, bottle, sculpture, floor, hill, river, sand, flower, sink, grandstand, shelf, dirt track, ashcan, skyscraper, counter, cabinet, television receiver, fence, bar, plate, tent, curtain, bag, without the background class.\\
        Answer: \\
        The target categories include <ref>tree</ref><ref>sidewalk</ref><ref> ... </ref><ref>pot</ref><ref>signboard</ref>, numbered sequentially starting from 0, without the background class.<mask>RLE string</mask>
    \end{minipage}%
}
Note that we evaluated ADE20k using a randomly sampled prompt from ``Semantic Segmentation for the Open World'' to demonstrate prompt flexibility, whereas other benchmarks used a fixed prompt for simplicity.

\textbf{Cityscapes~\citep{cordts2016cityscapes}.} The evaluation of Cityscapes (val) is consistent with the output format of ADE20K; the only difference in the input prompt is the category name (shuffle the names, too). The distinction lies in the resolution, which is multiplied by 2 instead of 4. To reduce computation, we performed a 2x2 non-overlapping crop of the images for testing. The specific prompts are as follows:

\noindent\fbox{%
    \begin{minipage}{\dimexpr\linewidth-2\fboxsep-2\fboxrule}
        \ttfamily
        <image>\\
        Question: \\
        Segment: train, motorcycle, vegetation, person, wall, terrain, pole, sky, traffic light, fence, bicycle, road, traffic sign, rider, building, truck, sidewalk, bus, car, without the background class.\\
        Answer: \\
        The target categories include <ref>road</ref><ref>sidewalk</ref><ref>car</ref>, numbered sequentially starting from 0.<mask>RLE string</mask>
    \end{minipage}%
}

\textbf{Context59~\citep{mottaghi2014role}.} 
The Pascal context (val) benchmark follows the aforementioned prompt and output protocols, with the input resolution upscaled to $3\times$ its original size. Context59 comprises 59 foreground categories, excluding background classes. To accommodate scenarios requiring background identification, the model supports an optional configuration where the phrase "without the background class" is omitted from the prompt. In this mode, a background channel (represented by <OTHERS>) is integrated into the post-processing pipeline: logits are scaled by a factor of 0.25 before a sigmoid activation, followed by the addition of a constant background score (0.5) to facilitate the argmax operation. However, for standard evaluation on Context59, we adhere to the default softmax-based refinement with a temperature of 0.2 and DenseCRF, consistent with our protocol for other semantic segmentation tasks. A representative prompt is illustrated below:

\noindent\fbox{%
    \begin{minipage}{\dimexpr\linewidth-2\fboxsep-2\fboxrule}
        \ttfamily 
        <image>\\
        Question: \\
        Segment: window, ceiling, dog, person, ground, keyboard, cloth, bus, bag, boat, sheep, wall, bicycle, snow, platform, grass, flower, computer, floor, truck, bottle, light, car, curtain, sign, bird, pottedplant, tree, cat, table, door, bed, food, train, sidewalk, bench, bedclothes, sofa, mountain, rock, water, building, aeroplane, plate, track, cabinet, horse, chair, cup, fence, road, tvmonitor, motorbike, sky, book, mouse, cow, wood, shelves, without the background class.\\
        Answer: \\
        The target categories include <ref>ground</ref><ref>grass</ref> ... <ref>rock</ref><ref>water</ref><ref>sky</ref>, numbered sequentially starting from 0.<mask>RLE string</mask>
    \end{minipage}%
}

\textbf{VOC20.} VOC20, or PASCAL VOC, contains 20 foreground classes without background classes. The test resolution multiples, prompt format, and output format are consistent with Context59. A specific example is as follows:

\noindent\fbox{%
    \begin{minipage}{\dimexpr\linewidth-2\fboxsep-2\fboxrule}
        \ttfamily
        <image>\\
        Question: \\
        Segment: train, sofa, sheep, horse, bird, cat, car, cow, boat, aeroplane, diningtable, pottedplant, chair, dog, bottle, person, motorbike, bicycle, tvmonitor, bus, without the background class.\\
        Answer: \\
        The target categories include <ref>car</ref><ref>boat</ref><ref>person</ref> <ref>bus</ref>, numbered sequentially starting from 0.<mask>RLE string</mask>
    \end{minipage}%
}

\textbf{COCOStuff~\citep{caesar2018coco}.} CocoStuff uses the same testing protocol as ADE20K, except that the resize scale is 3. The rest of the prompt format (including label shuffling) and output format are consistent. A specific example is as follows:

\noindent\fbox{%
    \begin{minipage}{\dimexpr\linewidth-2\fboxsep-2\fboxrule}
        \ttfamily 
        <image>\\
        Question: \\
        Segment: wall-tile, ground-other, surfboard, moss, fire hydrant, towel, backpack, couch, floor-wood, mirror-stuff, rock, handbag, door-stuff, paper, salad, gravel, door, plant-other, hill, snowboard, eye glasses, snow, person, tennis racket, wall-concrete, playingfield, plastic, wall-wood, bird, banana, carrot, bed, sandwich, furniture-other, knife, rug, plate, vase, elephant, clouds, kite, tv, cell phone, fork, apple, straw, stone, frisbee, suitcase, fence, cake, clock, train, grass, wood, donut, curtain, cow, cloth, floor-other, toaster, tree, giraffe, sports ball, shelf, flower, building-other, potted plant, carpet, fruit, window-blind, spoon, railroad, pillow, traffic light, bush, desk-stuff, sheep, bear, railing, bottle, hat, blender, baseball glove, sea, sky-other, hair drier, microwave, parking meter, zebra, tent, mouse, skis, counter, desk, banner, tie, oven, branch, scissors, structural-other, dirt, keyboard, fog, cupboard, wall-panel, wall-brick, floor-stone, food-other, wall-other, dining table, napkin, stop sign, cat, net, mirror, leaves, bus, house, bowl, mud, stairs, truck, laptop, table, textile-other, clothes, sand, window, vegetable, pizza, floor-tile, wine glass, waterdrops, river, road, floor-marble, cardboard, refrigerator, window-other, wall-stone, platform, cup, mountain, street sign, shoe, umbrella, car, book, chair, skyscraper, cabinet, skateboard, ceiling-other, blanket, toothbrush, bench, orange, light, hot dog, bridge, pavement, bicycle, solid-other, ceiling-tile, teddy bear, water-other, motorcycle, broccoli, horse, cage, mat, baseball bat, dog, roof, boat, metal, sink, hair brush, remote, toilet, airplane, without the background class.\\
        Answer: \\
        The target categories include <ref>rock</ref><ref>plant-other</ref> <ref>bird</ref><ref>grass</ref><ref>bush</ref><ref>bear</ref><ref>dirt</ref> <ref>water-other</ref>, numbered sequentially starting from 0.<mask>RLE string</mask>
    \end{minipage}%
}

\textbf{Semantic Segmentation for the Open World.}
Open set segmentation and specific benchmarks differ in the output content and interpretation results. Open sets support single objects, multiple objects, and a collection of labels. If there is no improvement without the background, the model will output a label of <OTHERS>, which will activate the sigmoid + threshold interpretation method. Specifically, the logits are divided by 4 and then passed through sigmoid, with the background class threshold defaulting to 0.5. The corresponding accuracy for open sets is generally not high, and fixed thresholds can sometimes lead to inaccuracies in the background class. We can achieve more accurate open-set segmentation results using the detection-then-segmentation type. The specific approach is to first invoke detection, then draw each box on the image, apply a padding of 1.2, resize, and then segment the foreground and background. In contrast, directly outputting semantic segmentation results is faster but relatively lower in quality, though it can additionally support stuff classes. The specific prompts for direct semantic segmentation are as follows (the Chinese version is also supported):

\noindent\fbox{%
    \begin{minipage}{\dimexpr\linewidth-2\fboxsep-2\fboxrule}
        \ttfamily 
Please perform segmentation for the following classes: \{keyword\}.\\
Can you identify and segment these objects: \{keyword\}?\\
I'm looking for \{keyword\}. Please segment them in the image.\\
Segment the image based on these keywords: \{keyword\}.\\
Mark the regions corresponding to: \{keyword\} in the image.\\
Identify the boundaries of these objects: \{keyword\}.\\
Please label and segment the following categories: \{keyword\}.\\
Show me the segmentation mask for: \{keyword\}.\\
Create segmentation maps for these items: \{keyword\}.\\
Locate and segment the following objects: \{keyword\}.\\
Segment the image to highlight: \{keyword\}.\\
Distinguish and segment these classes: \{keyword\}.\\
Please provide segmentation for objects: \{keyword\}.\\
Identify and separate the following elements: \{keyword\}.\\
Generate segmentation for the specified classes: \{keyword\}.\\
Semantic segmentation for \{keyword\}.\\
Perform semantic segmentation for \{keyword\}.\\
Run seamntic segmentation: \{keyword\}.\\
Segment: \{keyword\}.\\
Semantic segmentation: \{keyword\}.\\
Find \{keyword\} and segment.\\
Segment these: \{keyword\}.\\
Segment \{keyword\}.\\
Segment \{keyword\}.\\
Segment and label \{keyword\}.\\
Please segment \{keyword\}.\\
Could you please segment objects classified as: \{keyword\}?\\
Provide a detailed segmentation of the following: \{keyword\}.\\
Perform pixel-level segmentation for: \{keyword\}.\\
Mark and segment the specified classes: \{keyword\}.\\
Generate a segmented output focusing on: \{keyword\}.\\
Show segmentation results highlighting: \{keyword\}.\\
Can you segment and annotate the following keywords: \{keyword\}?\\
Provide segmentation masks specifically for: \{keyword\}.\\
Please segment any visible \{keyword\} in the image.\\
Produce a segmentation mask isolating \{keyword\}.\\
Segment and highlight the regions occupied by \{keyword\}.\\
Perform semantic segmentation targeting: \{keyword\}.\\
Segment and classify the following entities: \{keyword\}.\\
Mark the presence and segment the \{keyword\} in the image.\\
Generate masks that segment the \{keyword\} clearly.\\
Create a detailed segmentation for the objects: \{keyword\}.\\
Generate segmentation annotations for: \{keyword\}.\\
Provide a detailed semantic segmentation of the following categories: \{keyword\}.\\
Perform pixel-level semantic segmentation for: \{keyword\}.\\
Create precise semantic segmentation boundaries around these classes: \{keyword\}.\\
Generate a semantic segmentation output focusing on: \{keyword\}.\\
Segment \{keyword\} in the image.\\
Show semantic segmentation for \{keyword\}.\\
Find and segment \{keyword\}.\\
Please segment \{keyword\}.\\
Segment \{keyword\} categories.\\
Show \{keyword\} segmentation.\\
Classify and segment \{keyword\}."
\}. \\
Generate segmentation annotations for: \{keyword\}.
    \end{minipage}%
}

\{keyword\} indicates some category names such as ``dog, cat, tree, soft''. We also support ``segment anything'' type without specific keyword assignment. It is important to note that this model does not necessarily segment all elements in detail, but rather segments some significant main objects. Since background classes will also be output, sigmoid processing will be applied, with a background threshold of 0.5. The combined version of detection and semantic segmentation can address issues with inaccurate backgrounds. Suggested prompts include:

\noindent\fbox{%
    \begin{minipage}{\dimexpr\linewidth-2\fboxsep-2\fboxrule}
        \ttfamily 
Please segment all objects present in the image. \\
I want to segment everything in the image. \\
Segment every object visible in the photo. \\
Segment all objects in this image. \\
Segment all visible elements in the photo. \\
Segment all objects in the image. \\
Identify and segment all items. \\
Segment everything visible. \\
Find and segment all objects. \\
Label all objects in the image. \\
Segment all elements. \\
Perform semantic segmentation for this image. \\
Do semantic segmentation for this image. \\
Segment all. \\
Conduct semantic segmentation for all. \\
Perform semantic segmentation for this image. \\
Do semantic segmentation for this image. \\
Segment all. \\
Segment this image. \\
Segment it. \\
Segment everything. \\
Segment all \\
Semantic segmentation for all. \\
Semantic segmentation. \\
Run semantic segmentation \\
Segment this image \\
Conduct semantic segmentation for all. \\
Apply semantic segmentation to the image. \\
Run semantic segmentation on this image. \\
Execute semantic segmentation for all objects. \\
Carry out semantic segmentation for everything. \\
Perform full semantic segmentation. \\
Complete semantic segmentation for all items.

    \end{minipage}%
}

\subsubsection{Referring Expression Segmentation}

\textbf{RefCOCO series~\citep{yu2016modeling} (polygon).} The basic version of referring expression segmentation is implemented using polygons. This mode is a simple version, but not evaluated.

\noindent\fbox{%
    \begin{minipage}{\dimexpr\linewidth-2\fboxsep-2\fboxrule}
        \ttfamily
        <image>\\
        Question: \\
        Can you segment ``bowl behind the others can only see par'' in this image? \\
        Answer: \\
        The segmentation result is <ins><poly><x\_468><y\_3><x\_471><y\_46><x\_473><y\_83><x\_521> <y\_107><x\_581><y\_117><x\_640><y\_106><x\_640><y\_105><x\_640><y\_1></poly></ins>.
    \end{minipage}%
}
This is the basic referring segmentation prompt. Note that <ins> represents an object, and an object may have multiple <poly>, indicating different parts.

\textbf{RefCOCO series~\citep{yu2016modeling} (grounding-then-segmentation).} Due to the precision limitations from the number of points, we recommend first using grounding and then semantic segmentation. Specifically, after grounding outputs bounding boxes on the original image, we draw boxes on the image with random colors and crop the image using a padding ratio of 1.2 (an extra 0.2). We then resize the shorter side to 1280 for semantic segmentation to distinguish between foreground and background classes (supporting lower resolutions for acceleration). Similar to semantic segmentation, DenseCRF is also utilized as a post-processing step (not used with polygons). ``RLE string'' in the output indicates the pixel labels compressed by Run-Length Encoding in the string format. The evaluation metric is cIoU. Examples based on Polygon and semantic segmentation are shown below:

In the testing, we actually used grounding prompts, then drew boxes on the image and performed a 1.2 padding crop. After setting the short side to 1280, we executed the following semantic segmentation commands, and then filled the segmentation results back into the original image for testing.

\noindent\fbox{%
    \begin{minipage}{\dimexpr\linewidth-2\fboxsep-2\fboxrule}
        \ttfamily
        <image>\\
        Question: \\
        Can you segment ``Segment the code '' in this image? \\
        Answer: \\
        The segmentation result is <ins><poly><x\_468><y\_3><x\_471><y\_46><x\_473><y\_83><x\_521> <y\_107><x\_581><y\_117><x\_640><y\_106><x\_640><y\_105><x\_640><y\_1></poly></ins>.
    \end{minipage}%
}
This is the basic referring segmentation prompt. Note that <ins> represents an object, and an object may have multiple <poly>, indicating different parts.

\noindent\fbox{%
    \begin{minipage}{\dimexpr\linewidth-2\fboxsep-2\fboxrule}
        \ttfamily 
        <image>\\
        Question: \\
        Please provide the bounding box coordinate of the region this sentence describes: the person bottom left\\
        Answer: <box><x\_155><y\_154><x\_221><y\_206></box>
    \end{minipage}%
}

\noindent\fbox{%
    \begin{minipage}{\dimexpr\linewidth-2\fboxsep-2\fboxrule}
        \ttfamily 
        <image>\\
        Question: \\
        Please provide the bounding box coordinate of the region this sentence describes: bowl behind the others can only see part 
        Answer: \\
        <box><x\_54><y\_0><x\_361><y\_141></box>\\
        <cropped image> \#Draw the box, padding, and resize. \\
        Question: \\
        Segment the core target. \\
        Answer: \\
        The results are 0 for <ref><BG></ref> and 1 for <ref><FG></ref>.<mask>RLE string</mask>
    \end{minipage}%
}
Note that the mask result is from the cropped image. We fill it back to the raw image for evaluation via cIoU.

\textbf{Referring Expression Segmentation for the Open World.}
Referring segmentation and points are the keywords of this task, which will activate the model to output polygon segmentation results for a single target. The approach based on grounding + semantic segmentation can refer to the flexible prompts of grounding.

\noindent\fbox{%
    \begin{minipage}{\dimexpr\linewidth-2\fboxsep-2\fboxrule}
        \ttfamily 
Can you segment "\{keyword\}" via points? \\
Can you segment "\{keyword\}" in the manner of referring segmentation? \\
Referring segment for "\{keyword\}". \\
Referring segment for \{keyword\}. \\
Referring expression segmentation for \{keyword\}.'
Can you segment \{keyword\} via points? \\
Can you segment \{keyword\} in the manner of referring segmentation? \\
Outline \{keyword\} via points. \\
Outline \{keyword\} via the polygon or points. \\
Segment \{keyword\} via points. \\
Outline "\{keyword\}" via points. \\
Use points to segment \{keyword\}. \\
Use points to segment "\{keyword\}". \\
Use point to segment \{keyword\}. \\
Use point to segment "\{keyword\}". \\
Please \{keyword\}. \\
Use points to segment "\{keyword\}". \\
Outline "\{keyword\}" via the polygon or points. \\
Segment "\{keyword\}" via points. \\
Perform referring segmentation for the keywords: \{keyword\}. \\
Perform referring segmentation for "\{keyword\}". \\
Please segment "\{keyword\}" in the image using polygon or points. \\
Could you outline "\{keyword\}" with  points? \\
Draw the boundary of "\{keyword\}" via points. \\
Generate a referring segmentation mask for "\{keyword\}". \\
Segment \{keyword\} based on referring expression. \\
Segment "\{keyword\}" using points outlining. \\
Please perform referring expression segmentation for \{keyword\}. \\
Use polygon or points to segment "\{keyword\}". \\
Draw a polygon or points to segment "\{keyword\}" in the image. \\
Referring segmentation for the object "\{keyword\}". \\
Segment \{keyword\} in this picture by outlining with polygon or points. \\
Please extract the polygon or pointsal region corresponding to "\{keyword\}
    \end{minipage}%
}

\subsubsection{Depth Estimation}
\textbf{NYUv2~\citep{silberman2012indoor}.} The depth estimation test has four key points: (1) The category IDs are preset from <custom\_1> to <custom\_1000>, eliminating the need to predict category names, which increases speed. (2) Instead of using DenseCRF, we first upsample by a factor of two, followed by argmax, and then resize to the original image size. Appropriate resizing before argmax improves results. (3) Real depth needs to be dequantized. For NYUv2, we use uniform linear quantization, mapping real depths from 0 to 10 meters to 1 to 1000. Invalid depths are set to 0 and ignored during training. During testing, we dequantize to the actual depths and exclude invalid depths. (4) The prompt must precede the image so that the model can learn the potential quantization methods. The output format is a string of depth names corresponding to the pixel sizes after argmax and resizing. For NYUv2, since the images are relatively small, we performed a threefold resize based on the original resolution. The example is given below:

\noindent\fbox{%
    \begin{minipage}{\dimexpr\linewidth-2\fboxsep-2\fboxrule}
        \ttfamily 
        Question: \\
        Please estimate the depth of this image from the NYUv2 dataset.\\
        <image>\\
        Answer: \\
        This is the <depth>.<mask>RLE string</mask>
    \end{minipage}%
}

\textbf{Cityscapes~\citep{cordts2016cityscapes}.} The testing criteria for Cityscapes are the same as for NYUv2. The difference lies in quantizing the effective depth from 0-80m to 1-1000, and since the resolution is sufficient, no resizing was done. Besides, we set invalid regions from the left and bottom pars following previous works. Specific examples are as follows:

\noindent\fbox{%
    \begin{minipage}{\dimexpr\linewidth-2\fboxsep-2\fboxrule}
        \ttfamily 
        Question: \\
        Please estimate the depth of this image from the Cityscapes dataset.\\
        <image>\\
        Answer: \\
        This is the <depth>.<mask>RLE string</mask>
    \end{minipage}%
}

\textbf{DDAD~\citep{guizilini20203d}.} The testing criteria for DDAD are the same as for NYUv2. The difference lies in quantizing the effective depth from 0.05-120m to 1-1000, and since the resolution is sufficient, no resizing was done. Specific examples are as follows:

\noindent\fbox{%
    \begin{minipage}{\dimexpr\linewidth-2\fboxsep-2\fboxrule}
        \ttfamily 
        Question: \\
        Please estimate the depth of this image from the DDAD dataset.\\
        <image>\\
        Answer: \\
        This is the <depth>.<mask>RLE string</mask>
    \end{minipage}%
}

\textbf{Depth Estimation for the Open World.}
Depth estimation in open world scenes defaults to Log-uniform Quantization, mapping real depths ranging from 0.5 to 100m to 1 to 1000. Values outside this range are set to 0 (IGNORE). During testing, the predicted values need to be de-quantized to obtain the real depth; otherwise, only relative depth is obtained. Additionally, we require inputs to be resized to a focal length of 2000 pixels to simulate a camera with a default focal length. Discrepancies in actual focal length may lead to biased predicted depths, resulting in relative depth outputs. Unlike specific datasets that require prompts to be placed in advance, our open-set supports prompts both before and after the image. It is important to note that this task relies on a large amount of training data, and the capabilities of depth anything still need improvement. However, we have demonstrated that the standard VLM model can accommodate several different quantization methods, inherently building a spatial depth perception without requiring additional structures. We recommend using the following prompts and their Chinese translation versions:

\noindent\fbox{%
    \begin{minipage}{\dimexpr\linewidth-2\fboxsep-2\fboxrule}
        \ttfamily 
  Estimate the depth. \\
  Estimate the depth in the default range. \\
  Predict the depth map. \\
  Estimate the depth map. \\
  What the depth map? \\
  Execute depth estimation. \\
  Run depth estimation. \\
  Generate a depth map. \\
  Calculate the depth. \\
  Provide depth estimation. \\
  Perform depth prediction. \\
  Create the depth map. \\
  Get the depth map for the image. \\
  Compute the depth information. \\
  Estimate distance using depth. \\
  Run a depth map generation. \\
  Can you predict the depth map? \\
  Show me the depth estimation. \\
  Depth map prediction, please. \\
  Extract depth from the image. \\
  Apply depth estimation.
    \end{minipage}%
}

\subsubsection{Human Pose Estimation}

\textbf{MPII~\citep{andriluka20142d}.}
We evaluate our model on the MPII validation set using the standard PCKh@0.5 metric. Following the evaluation setting of ViTPose~\citep{xu2022vitpose}, where results are reported on images cropped using ground-truth person bounding boxes, the input images are constructed from ground-truth bounding box crops. As our model predicts multiple human poses per image in a single forward pass, we adopt a post-processing strategy to enable one-to-one evaluation. Specifically, for each image, we compute the keypoint center of the ground-truth instance and select the predicted pose whose keypoint center is spatially closest to it, discarding all other predictions. The selected prediction is then evaluated against the corresponding ground-truth pose using the standard PCKh metric.

\noindent\fbox{%
    \begin{minipage}{\dimexpr\linewidth-2\fboxsep-2\fboxrule}
        \ttfamily
        <image>\\
        Question: Detect all persons and their poses from the image within the class set of MPII Human Pose Dataset. The output format should be a JSON-like string, containing person instances. Each person instance is enclosed in <person>...</person> tags. Within each person instance, provide the bounding box using <box>...</box> tags and their 16 keypoints using <kpt>...</kpt> tags. The bounding box is defined by <x\_x1><y\_y1><x\_x2><y\_y2> tags, and each keypoint is defined by <x\_...><y\_...><v\_...> tags, where x, y are coordinates and v is visibility. The joints must follow the standard MPII ordering.\\
        Answer: <person><box><x\_457><y\_187><x\_548><y\_315></box>
        <kpt><x\_531><y\_793><v\_1.0></kpt>
        <kpt><x\_624><y\_659><v\_1.0></kpt>
        \ldots
        <kpt><x\_602><y\_281><v\_1.0></kpt>
        </person>
    \end{minipage}%
}
\noindent\emph{Note.} In the structured output, each coordinate and visibility value (e.g., \texttt{<x\_*>}, \texttt{<y\_*>}, \texttt{<v\_*>}) is represented as a dedicated special token and is directly learned during training. For brevity, only a subset of keypoints is shown in the example above; in practice, the model predicts the complete set of 16 keypoints for each detected person following the MPII joint ordering.

\subsection{General Multimodal Benchmarks}
\label{sec:app_eval_mm}

We provide the detailed evaluation settings and prompts for the general multimodal benchmarks reported in the main paper. All evaluations were conducted using a refined VLM evaluation framework based on the VLMEvalKit~\citep{duan2024vlmevalkit}. Specifically, we enhanced the framework to address the limitations of the original exact-matching protocol used in certain Multiple-Choice Question (MCQ) benchmarks. To improve evaluation accuracy, we integrated an LLM-based judging mechanism that leverages a large language model to robustly extract and parse final answers from the model's generated responses.

\noindent\textbf{Implementation Details.} \quad 
To ensure optimal performance across diverse tasks, we tailored the visual resolution settings based on the benchmark requirements. 
For the majority of benchmarks that demand fine-grained visual details (e.g., DocVQA~\citep{docvqa}, MMMU~\citep{yue2024mmmu}, MMBench~\citep{MMBench}, RealWorldQA~\citep{realworldqa2024}), we employed a dynamic high-resolution strategy. Specifically, we set the minimum number of patches to 1,280 (default is 64), with a maximum of 65,536 patches. 
For benchmarks less sensitive to extreme resolution or where standard resolution suffices (e.g., MathVista~\citep{mathvista}, OCRBench~\citep{Liu_2024_OCRBench}, ScienceQA~\citep{lu2022learn_scienceqa}, HallusionBench~\citep{hallusion_bench}), we retained the default setting (\texttt{min\_num\_patches=64}).

Regarding the evaluation protocol, particularly for MCQ benchmarks, we adopted a hybrid judging strategy to ensure accuracy. We first attempt to extract the answer using exact matching. If the extraction fails or the model's output is ambiguous, we utilize an LLM-based judge to parse the response.

\noindent\textbf{Prompt Specifics.} \quad Regarding prompting strategies, we categorized benchmarks based on their task nature. For benchmarks requiring complex reasoning (e.g., MathVerse~\citep{zhang2024mathverse}, MMMU, VisuLogic~\citep{xu2025visulogicbenchmarkevaluatingvisual}) or real-world analysis (e.g., RealWorldQA), we selectively activated Chain-of-Thought (CoT) prompting (e.g., "Think step-by-step"). For hallucination and OCR tasks, we employed strict constraints to ensure concise and precise outputs. The detailed configurations of the prompts are provided below.

\textbf{VisuLogic}~\citep{xu2025visulogicbenchmarkevaluatingvisual}. \quad For this benchmark, which focuses on complex visual logical reasoning, we utilized a specialized prompt that explicitly instructs the model to reason step-by-step before providing the final answer in a boxed format.

\noindent\fbox{%
    \begin{minipage}{\dimexpr\linewidth-2\fboxsep-2\fboxrule}
        \ttfamily
        <image>\\
        \{Question\}\\
        Solve the complex visual logical reasoning problem through step-by-step reasoning.Think about the reasoning process first and answer the question following this format: Answer: \textbackslash boxed\{LETTER\}\\
        Think step-by-step.
    \end{minipage}%
}

\textbf{MathVerse \& Mathvista \& LogicVista}~\citep{zhang2024mathverse, mathvista, xiao2024logicvista}. \quad To tackle complex reasoning in these benchmarks, we incorporated the CoT strategy to elicit structured geometric and mathematical derivations.

\noindent\fbox{%
    \begin{minipage}{\dimexpr\linewidth-2\fboxsep-2\fboxrule}
        \ttfamily
        <image>\\
        \{Question\}\\
        Think step-by-step.
    \end{minipage}%
}

\textbf{MMMU (VAL \& Pro\_Standard)}~\citep{mmmupro,yue2023mmmu}. \quad For the massive multi-discipline understanding benchmarks, we appended a standard CoT instruction after the options to encourage detailed reasoning.

\noindent\fbox{%
    \begin{minipage}{\dimexpr\linewidth-2\fboxsep-2\fboxrule}
        \ttfamily
        <image>\\
        Hint: \{Hint\}\\
        Question: \{Question\}\\
        Options:\\
        A. \{OptionA\}\\
        B. \{OptionB\}\\
        ...\\
        Please select the correct answer from the options above.\\
        Think step-by-step.
    \end{minipage}%
}

\textbf{MMMU (Pro\_V)}~\citep{mmmupro}. \quad The vision-only protocol of MMMU-Pro. In this configuration, redundant textual prompts are removed and replaced by CoT instructions to elicit depth-first visual analysis and logical deduction.

\noindent\fbox{%
    \begin{minipage}{\dimexpr\linewidth-2\fboxsep-2\fboxrule}
        \ttfamily
        <image>\\
        Think step-by-step.
    \end{minipage}%
}

\textbf{RealWorldQA}~\citep{realworldqa2024}. \quad Although this is a VQA task, we found that spatial and physical understanding benefits significantly from reasoning. Therefore, we forced the "Think step-by-step" instruction.

\noindent\fbox{%
    \begin{minipage}{\dimexpr\linewidth-2\fboxsep-2\fboxrule}
        \ttfamily
        <image>\\
        \{Question\}\\
        Think step-by-step.
    \end{minipage}%
}

\textbf{MMBench \& CV-Bench}~\citep{liu2023mmbench,zhu2025cvbench}. \quad For general perception and reasoning, we utilized the standard CircularEval strategy provided by VLMEvalKit, augmented with a step-by-step thinking instruction.

\noindent\fbox{%
    \begin{minipage}{\dimexpr\linewidth-2\fboxsep-2\fboxrule}
        \ttfamily
        <image>\\
        Hint: \{Hint\}\\
        Question: \{Question\}\\
        Options:\\
        A. \{OptionA\}\\
        ...\\
        Please select the correct answer from the options above.\\
        Think step-by-step.
    \end{minipage}%
}

\textbf{MME-RealWorld}~\citep{mme-realworld}. \quad For this benchmark, we used a detailed system prompt to guide the model in selecting the best option, ending with a specific suffix to induce the answer generation.

\noindent\fbox{%
    \begin{minipage}{\dimexpr\linewidth-2\fboxsep-2\fboxrule}
        \ttfamily
        <image>\\
        \{Question\}\\
        \{Options\}\\
        Select the best answer to the above multiple-choice question based on the image. Respond with the letter (A, B, C, D, or E) of the correct option.\\
        The best answer is:
    \end{minipage}%
}

\textbf{HallusionBench \& POPE}~\citep{hallusion_bench,li2023evaluatingpope}. \quad To rigorously test for hallucinations, we disabled CoT and enforced a strict "Yes/No" constraint to prevent the model from generating evasive or verbose responses.

\noindent\fbox{%
    \begin{minipage}{\dimexpr\linewidth-2\fboxsep-2\fboxrule}
        \ttfamily
        <image>\\
        \{Question\} Please answer yes or no.
    \end{minipage}%
}

\textbf{DocVQA, InfoVQA, TextVQA, ChartQA \& AI2D}~\citep{docvqa,Mathew2021InfographicVQA,textvqa,masry2022chartqa}. \quad For document understanding and OCR-related tasks, we appended a constraint to ensure the output is a short phrase or single word, facilitating accurate metric calculation.

\noindent\fbox{%
    \begin{minipage}{\dimexpr\linewidth-2\fboxsep-2\fboxrule}
        \ttfamily
        <image>\\
        \{Question\} Answer the question with a single word or phrase.
    \end{minipage}%
}

\textbf{ScienceQA \& SEEDBench}~\citep{lu2022learn_scienceqa,li2023seedbench}. \quad For these general multiple-choice benchmarks, we used a direct answer prompt without CoT to evaluate the model's direct knowledge retrieval capabilities.

\noindent\fbox{%
    \begin{minipage}{\dimexpr\linewidth-2\fboxsep-2\fboxrule}
        \ttfamily
        <image>\\
        Hint: \{Hint\}\\
        Question: \{Question\}\\
        Choices:\\
        A. \{OptionA\}\\
        ...\\
        Answer the question with a single word or phrase.
    \end{minipage}%
}

\textbf{OSWorld}~\citep{xie2024osworld}. \quad For the OSWorld benchmark, we use the prompts below to evaluate the candidate models. 

\noindent\fbox{%
    \begin{minipage}{\dimexpr\linewidth-2\fboxsep-2\fboxrule}
        \ttfamily
        System:\\
        \# Tools
        
You may call one or more functions to assist with the user query.

You are provided with function signatures within <tools></tools> XML tags:
<tools>
\{"type": "function", "function": \{"name\_for\_human": "computer\_use", "name": "computer\_use", "description": "Use a mouse and keyboard to interact with a computer, and take screenshots.\\\* This is an interface to a desktop GUI. You do not have access to a terminal or applications menu. You must click on desktop icons to start applications.\\\* Some applications may take time to start or process actions, so you may need to wait and take successive screenshots to see the results of your actions. E.g. if you click on Firefox and a window doesn't open, try wait and taking another screenshot.\\\* The screen's resolution is \{Resolution\}.\\\* Whenever you intend to move the cursor to click on an element like an icon, you should consult a screenshot to determine the coordinates of the element before moving the cursor.\\\* If you tried clicking on a program or link but it failed to load even after waiting, try adjusting your cursor position so that the tip of the cursor visually falls on the element that you want to click.\\\* Make sure to click any buttons, links, icons, etc with the cursor tip in the center of the element. Don't click boxes on their edges unless asked.", "parameters": \{"properties": \{"action": \{"description": "\\\* `key`: Performs key down presses on the arguments passed in order, then performs key releases in reverse order.\\\* `type`: Type a string of text on the keyboard.\\\* `mouse\_move`: Move the cursor to a specified (x, y) pixel coordinate on the screen.\\\* `left\_click`: Click the left mouse button at a specified (x, y) pixel coordinate on the screen.\\\* `left\_click\_drag`: Click and drag the cursor to a specified (x, y) pixel coordinate on the screen.\\\* `right\_click`: Click the right mouse button at a specified (x, y) pixel coordinate on the screen.\\\* `middle\_click`: Click the middle mouse button at a specified (x, y) pixel coordinate on the screen.\\\* `double\_click`: Double-click the left mouse button at a specified (x, y) pixel coordinate on the screen.\\\* `triple\_click`: Triple-click the left mouse button at a specified (x, y) pixel coordinate on the screen (simulated as double-click since it's the closest action).\\\* `scroll`: Performs a scroll of the mouse scroll wheel.\\\* `hscroll`: Performs a horizontal scroll (mapped to regular scroll).\\\* `wait`: Wait specified seconds for the change to happen.\\\* `terminate`: Terminate the current task and report its completion status.\\\* `answer`: Answer a question.\\        ", "enum": ["key", "type", "mouse\_move", "left\_click", "left\_click\_drag", "right\_click", "middle\_click", "double\_click", "scroll", "wait", "terminate"], "type": "string"\}, "keys": \{"description": "Required only by `action=key`.", "type": "array"\}, "text": \{"description": "Required only by `action=type`.", "type": "string"\}, "coordinate": \{"description": "The x,y coordinates for mouse actions.", "type": "array"\}, "pixels": \{"description": "The amount of scrolling.", "type": "number"\}, "time": \{"description": "The seconds to wait.", "type": "number"\}, "status": \{"description": "The status of the task.", "type": "string", "enum": ["success", "failure"]\}\}, "required": ["action"], "type": "object"\}, "args\_format": "Format the arguments as a JSON object."\}\}
</tools>

\end{minipage}%
}

\noindent\fbox{%
    \begin{minipage}{\dimexpr\linewidth-2\fboxsep-2\fboxrule}
        \ttfamily
For each function call, return a JSON object with the function name and arguments within <tool\_call></tool\_call> XML tags:
<tool\_call>
\{"name": <function-name>, "arguments": <args-json-object>\}
</tool\_call>

\# Response format

Response format for every step:
1) Action: a short imperative describing what to do in the UI.
2) A single <tool\_call>...</tool\_call> block containing only the JSON: \{"name": <function-name>, "arguments": <args-json-object>\}.

Rules:
- Output exactly in the order: Action, <tool\_call>.
- Be brief: one sentence for Action.
- Do not output anything else outside those parts.
- If finishing, use action=terminate in the tool call.\\
User:\\
<image>\\
        Please generate the next move according to the UI screenshot, instruction and previous actions.\\
        Instruction: \{Question\}\\
        Previous actions:\\
        \{Prevision Action List\}
    \end{minipage}%
}

\textbf{ScreenSpot Pro}~\citep{li2025screenspot}. \quad For the ScreenSpot Pro benchmark, we used the following prompt to evaluate the candidate models.

\noindent\fbox{%
    \begin{minipage}{\dimexpr\linewidth-2\fboxsep-2\fboxrule}
        \ttfamily
        System:\\
\# Tools\\
You may call one or more functions to assist with the user query.\\
You are provided with function signatures within <tools></tools> XML tags:\\
<tools>\{ "name":"computer\_use", "description": "Use a mouse to interact with a computer. The screen’s resolution is \{Resolution\}." "notes": "Click with the cursor tip centered on targets; avoid edges unless asked. Do not use other tools (type, key, scroll, left\_click\_drag). Only left\_click and mouse\_move are allowed. If you can’t find the element, terminate and report failure.", "parameters":\{ "type":"object", "required":["action"], "properties":\{ "action":\{ "type":"string", "enum":["mouse\_move","left\_click"], "description":"The action to perform." \}, "coordinate":\{ "type":"array", "description":"(x, y): pixels from left/top. Required for action=mouse\_move and action=left\_click." \} \} \} \} \}\\
</tools>\\
For each function call, return a json object with function name and arguments within <tool\_call>. . .</tool\_call> XML tags:\\
<tool\_call>\\
\{"name": <function-name>, "arguments": <args-json-object>\}\\
</tool\_call>\\
Response format for every step:\\
1) Action: a short imperative describing what to do in the UI.\\
2) A single <tool\_call>...</tool\_call> block containing only the JSON: \{"name": <function-name>, "arguments": <args-json-object>\}.\\\\
Rules:\\
- Output exactly in the order: Action, <tool\_call>.\\
- Be brief: one sentence for Action.\\
- Do not output anything else outside those parts.\\
- If finishing, use action=terminate in the tool call.
    \end{minipage}%
}
\noindent\fbox{%
    \begin{minipage}{\dimexpr\linewidth-2\fboxsep-2\fboxrule}
        \ttfamily
        User:\\
        <image>\\
You are given a screenshot of a desktop GUI and an instruction describing a target UI element.\\
Your goal is to choose exactly one action using the computer\_use tool to click on that target.\\\\
Task instruction:\\
\{instruction\}\\\\
You must output exactly one step following the Response format, with a single computer\_use tool call whose `coordinate` corresponds to the center of the target element in a \{resolution\} relative coordinate system (x from left to right, y from top to bottom).
    \end{minipage}%
}

%% file: sec/c_extended_expriments.tex
\section{Extended Experiments and Comparison}

Vision-centric tasks encompass a wide variety of settings, making comprehensive comparison challenging. To enable thorough evaluation beside Table \ref{tab:eval_cv}, we provide two additional supplementary tables—Table \ref{tab:seg_details} and Table \ref{tab:loc_details}, focusing on segmentation and grounding, respectively. In these tables, we systematically compare against five representative settings: (1) vision specialist models, (2) CLIP-based vision–language models, (3) vision generalist architectures, (4) multimodal LLMs with task-specific additions, and (5) our standard multimodal LLM, which uses no extra modules, heads, or task embeddings. Our results show that, despite its architectural simplicity, our approach achieves competitive or state-of-the-art performance across diverse settings. This demonstrates that a standard MLLM, when equipped with proper supervision and training, can serve as a highly effective and universal predictor, combining strong performance, minimal design, and broad applicability.

\subsection{Comparison With Dense Prediction Methods}
\label{sec:ext_exp}

The results presented in Table \ref{tab:seg_details} demonstrate the superior performance of our proposed Youtu-VL across multiple dense prediction benchmarks, highlighting its capability as a robust vision-language generalist without requiring complex architectural additions.

In the task of semantic segmentation, Youtu-VL achieves remarkable results across all five datasets. Notably, it attains 54.2 mIoU on ADE20K, compared with 47.8 mIoU of the vision generalist models like GiT~\citep{wang2024git} and 32.1 mIoU of the CLIP-based methods such as SAN~\citep{xu2023side}. Crucially, distinct from standard Multimodal LLMs that are typically unable to perform these tasks (indicated by ``×''), Youtu-VL demonstrates a unique capability to handle fine-grained dense prediction directly.

In the depth estimation tasks, Youtu-VL shows competitive results. It achieves 90.4 on NYUv2, while the result of DepthLLM~\citep{cai2025depthlm} is 87.6. Our method just needs to process the image once, while DepthLLM needs to draw each point and inference multiple times. We also fine-tune the model like the vision specialist model SwinMTL~\citep{taghavi2024swinmtl} and achieve a 92.7 $\delta_1$ (vs. 92.1 of SwinMTL). While the specialist model UniDepth-v2~\citep{piccinelli2025unidepthv2} performs higher on NYUv2, the performance on DDAD is quite close (88.2 vs. 87.6), suggesting this general-purpose model sometimes meets the high performance of task-specific models.

\begin{table}[h]
\centering
\setlength{\tabcolsep}{1.5pt}
\resizebox{\textwidth}{!}{%
\begin{tabular}{l c | c c c c c  | c c c | c c c c c c c c}
\toprule
 &  &  \multicolumn{5}{c|}{\textbf{Semantic Segmentation}} & \multicolumn{3}{c|}{\textbf{Depth Estimation}} & \multicolumn{8}{c}{\textbf{Referring Segmentation} (RefCOCO)}\\
\textbf{Methods} & \textbf{Additions} & ADE & COCOstuff & Context59 & Cityscapes & VOC20 & NYUv2 & DDAD & Cityscapes & val & testA & testB & val+ & testA+ & testB+ & val-g & test-g \\
\midrule
\multicolumn{10}{l}{\textit{\textbf{Vision Specialist Models}}} \\
Segformer (MiT-B5)~\citep{xie2021segformer} & MLP Decoder & \textcolor{gray}{51.0} & \textcolor{gray}{\textbf{46.7}} & - & \textcolor{gray}{82.4} & - & × & × & × & × & × & × & × & × & × & × & ×\\
Mask2Former (Swin-L)~\citep{cheng2021mask2former} & Pixel Decoder & \textcolor{gray}{\textbf{56.0}} & - & - & \textcolor{gray}{\textbf{83.3}} & - & × & × & × & × & × & × & × & × & × & × & ×\\
UniDepth-v2 (ViT-L)~\citep{piccinelli2025unidepthv2} & Depth Decoder & - & - & - & - & - & \textbf{98.8} & \textbf{88.2} & - & × & × & × & × & × & × & × & ×\\
SwinMTL (Swin-B)~\citep{taghavi2024swinmtl} & MLP Decoder & - & - & - & - & \textcolor{gray}{\textbf{76.41}} & - & - & \textcolor{gray}{\textbf{92.1}} & × & × & × & × & × & × & × & ×\\
RVG (ViT-B)~\citep{ouyang2025region} & MLP Decoder & × & × & × & × & × & × & × & × & \textcolor{gray}{\textbf{79.4}} & \textcolor{gray}{\textbf{81.2}} & \textcolor{gray}{\textbf{77.8}} & \textcolor{gray}{\textbf{69.5}} & \textcolor{gray}{\textbf{75.7}} & \textcolor{gray}{\textbf{63.0}} & \textcolor{gray}{\textbf{71.3}} & \textcolor{gray}{\textbf{72.1}}\\
VPD (UNet)~\citep{zhao2023unleashing} & Denoising Decoder & \textcolor{gray}{53.7} & - & - & - & - & \textcolor{gray}{96.4} & - & - & \textcolor{gray}{73.3} & - & - & \textcolor{gray}{62.7} & - & - & \textcolor{gray}{62.0} & -\\
\midrule
\multicolumn{10}{l}{\textit{\textbf{CLIP-based Vision Language Models}}} \\
MaskCLIP~\citep{zhou2022extract} & Attn Adaptation & 12.3 & 16.9 & 26.2 & 25.6 & 62.9 & × & × & × & × & × & × & × & × & × & × & ×\\
CLIPSurgery~\citep{LI2025111409} & Consistent Attn & 16.1 & 21.9 & 29.3 & 31.4 & 77.5 & × & × & × & × & × & × & × & × & × & × & ×\\
CASS~\citep{kim2025distilling} & VFM Graph & 20.4 & 26.7 & 40.2 & \textbf{39.4} & 87.8 & × & × & × & × & × & × & × & × & × & × & ×\\
SAN (ViT-L)~\citep{xu2023side} & Decoupled Head & \textbf{32.1} & \textcolor{gray}{\textbf{45.8}} & \textbf{57.7} & - & \textbf{94.6} & × & × & × & × & × & × & × & × & × & × & ×\\
\midrule
\multicolumn{10}{l}{\textit{\textbf{Vision Generalist Models}}} \\
X-Decoder (DaViT-d5)~\citep{zou2023generalized} & X-Decoder & \textcolor{gray}{58.1} & - & 60.4 & \textcolor{gray}{\textbf{81.7}} & 97.7 & × & × & × & - & - & - & - & - & - & - & -\\
4M (ViT-L)~\citep{mizrahi20234m} & Task-specific Heads & \textcolor{gray}{53.4} & - & - & - & - & \textcolor{gray}{\textbf{94.4}} & - & - & × & × & × & × & × & × & × & ×\\
SEEM (DaViT-d5)~\citep{mizrahi20234m} & SEEM-Decoder & - & - & - & - & - & × & × & × & - & - & - & - & - & - & 65.6 & -\\
BEIT-3 (Multiway-T)~\citep{wang2023image} & None & \textcolor{gray}{\textbf{62.8}} & - & - & - & - & × & × & × & × & × & × & × & × & × & × & ×\\
GiT~\citep{wang2024git} & Parallel Decoding & 47.8 & \textbf{49.1} & \textbf{63.3} & 61.8 & - & × & × & × & × & × & × & × & × & × & × & ×\\
SAM3~\citep{carion2025sam} & DETR-like Decoder & 13.8 & - & 60.8 & 65.2 & - & - & - & - & \textbf{75.5} & \textbf{77.6} & \textbf{71.0} & \textbf{67.3} & \textbf{71.1} & \textbf{63.4} & \textbf{73.4} & \textbf{74.0} \\

\midrule
\multicolumn{10}{l}{\textit{\textbf{VLM with Additions}}} \\
GLaMM (Vicuna-7B)~\citep{rasheed2024glamm} & SAM/Pixel Decoder & × & × & × & × & × & × & × & × & 79.5 & \textbf{83.2} & 76.9 & 72.6 & 78.7 & 64.6 & 74.2 & 74.9\\
UniPixel (Qwne2.5-VL-3B)~\citep{rasheed2024glamm} & SAM Decoder & × & × & × & × & × & × & × & × & \textbf{80.5} & 82.6 & 76.9 & 74.3 & 78.9 & 68.4 & \textbf{76.3} & \textbf{77.0}\\
VisionLLM v2 (Swin-T)~\citep{wu2024visionllm} & Deform-DETR & \textcolor{gray}{52.3} & - & - & - & - & x & x & x & 76.6 & 79.3 & 74.3 & 64.5 & 69.8 & 61.5 & 70.7 & 71.2\\
UFO (InternVL2.5-8B)~\citep{tang2025ufo} & Mask Tokens & \textbf{54.5} & \textbf{30.2} & - & - & - & \textcolor{gray}{\textbf{93.6}} & - & - & 80.0 & 81.6 & \textbf{78.1} & \textbf{76.7} & \textbf{79.9} & \textbf{72.3} & 75.5 & 76.3\\
\midrule
\hline
\multicolumn{10}{l}{\textit{\textbf{Standard VLM}}} \\
\rowcolor{blue!4}InternVL-3.5 (4B)~\citep{wang2025internvl35advancingopensourcemultimodal} & None & × & × & × & × & × & × & × & × & × & × & × & × & × & × & × & ×\\
\rowcolor{blue!4}Qwen3-VL (4B)~\citep{Qwen3-VL} & None & × & × & × & × & × & × & × & × & × & × & × & × & × & × & × & ×\\
\rowcolor{blue!4}DepthLLM (Qwne2.5-VL-3B)~\citep{cai2025depthlm} & None & × & × & × & × & × & 86.8 & 74.7 & - & × & × & × & × & × & × & × & ×\\
\rowcolor{blue!4}VistaLLM (Vicuna-7B)~\citep{pramanick2024jack} & None & × & × & × & × & × & × & × & × & 74.5 & 76.0 & 72.7 & 69.1 & 73.7 & 64.0 & 69.0 & 70.9\\
\rowcolor{blue!8} \textbf{Youtu-VL} (\textbf{4B, Ours}) & None & \textbf{54.2} & \textbf{52.2} & \textbf{60.4} & \textbf{70.4} & \textbf{92.5} & \textbf{90.4} & \textbf{87.6} & \textcolor{gray}{\textbf{92.7}} & \textbf{80.7} & \textbf{82.0} & \textbf{78.4} & \textbf{76.2} & \textbf{79.6} & \textbf{71.4} & \textbf{76.5} & \textbf{76.6} \\
\bottomrule
\end{tabular}}
\caption{\label{tab:seg_details}\textbf{Quantitative comparison on dense prediction tasks}. We evaluate the performance on Semantic Segmentation (ADE20K, COCO-Stuff, Pascal Context, Cityscapes, VOC2012), Depth Estimation (NYUv2, DDAD, Cityscapes), and Referring Segmentation (RefCOCO, RefCOCO+, RefCOCOg). The table compares our proposed Youtu-VL against Vision Specialist Models, CLIP-based methods, Vision Generalist Models, and other Multimodal LLMs (both with architectural additions and standard setups). "Additions" refers to the usage of additional task-specific decoders, heads, or task tokens. ``×'' denotes that the method is not applicable and ``-'' indicates results are not available. Results in \textcolor{gray}{gray} indicate task-specific fine-tuning on a single dataset, which usually brings higher results. Best results in each setting are marked in bold.}
\end{table}

Youtu-VL demonstrates strong performance in the referring expression segmentation task. On the RefCOCO val, Youtu-VL achieves cIoU of 80.7, while the baseline results are 79.5 for GLaMM~\citep{rasheed2024glamm} and 80.5 for UniPixel~\citep{liu2025unipixel}, which requires an extra SAM decoder. Besides, the method based on the polygon prediction is merely 74.5. The UFO~\citep{tang2025ufo} achieves an 80.0 cIoU (80.7 of Youtu-VL), while this method needs extra mask token embeddings for a retrieval process. Compared with the above methods. Our method is straightforward yet highly effective, designed to be applicable to standard VLMs without requiring extra architectural modifications.

\subsection{Comparison on Localization Tasks}
\label{sec:ext_exp_loc}

The results presented in Table~\ref{tab:loc_details} demonstrate that Youtu-VL achieves superior performance in localization tasks without requiring complex architectural additions. This success is attributed to the synergistic interplay among detection, counting, and grounding capabilities, which establishes leading performance within a standard VLM framework.

In object detection, Youtu-VL achieves 47.1 mAP on the COCO validation split, underscoring the potential of standard VLMs in handling dense localization tasks. Furthermore, performing SFT exclusively on detection data starting from Stage 3 boosts performance to 48.0 mAP, effectively bridging the gap with specialized architectures.
Regarding visual grounding, Youtu-VL secures leading results across all splits of the RefCOCO series. Similarly, in counting tasks, the model exhibits remarkable performance on benchmarks such as CountBench, TallyQA-simple, and TallyQA-complex.

\begin{table}[h]
\centering
\setlength{\tabcolsep}{3pt}
\small
\resizebox{1.0\textwidth}{!}{%
\begin{tabular}{l | c | c c c | c c c c c c c c}
\toprule
\textbf{Methods} & \multicolumn{1}{c|}{\textbf{Detection}} & \multicolumn{3}{c|}{\textbf{Counting}} & \multicolumn{8}{c}{\textbf{Grounding}} \\
 & {COCO val} & {CountBench} & {TallyQA-Simple} & {TallyQA-Complex} & val & testA & testB & val+ & testA+ & testB+ & val-g & test-g \\
\hline

\multicolumn{13}{l}{\textit{\textbf{Vision Specialist Models}}} \\
Mr. DETR++ (Swin-L)~\citep{zhang2025mr} & \textbf{58.4} & x & x & x & x & x & x & x & x & x & x & x \\
Omni-SMoLA (PaLI-X$_{FT}$)~\citep{wu2024omni}  & x & - & \textbf{86.3} & \textbf{77.1} & x & x & x & x & x & x & x & x  \\
MDETR (ENB3)~\citep{kamath2021mdetr}  & x & x & x & x & \textbf{87.5} & \textbf{90.4} & \textbf{82.7} &\textbf{ 81.1} & \textbf{85.5} & \textbf{73.0} & \textbf{83.4} & \textbf{83.3} \\

\hline
\multicolumn{13}{l}{\textit{\textbf{Vision Generalist Models}}} \\
BEIT-3 (Multiway-T)~\citep{wang2023image} & \textbf{63.7} & x & x & x & x & x & x & x & x & x & x & x \\
Florence-2 (DaViT-B)~\citep{xiao2024florence}  & 43.4 & x & x & x & \textbf{93.4} & \textbf{95.3} & \textbf{92.0} & \textbf{88.3} & \textbf{92.9} & \textbf{83.6} & \textbf{91.2} & \textbf{91.7} \\
Grounding DINO (Swin-L)~\citep{liu2024grounding}  & 62.6 & x & x & x & 90.6 & 93.2 & 88.2 & 82.8 & 89.0 & 75.9 & 86.1 & 87.0 \\
SAM3 (Perception Encoder)~\citep{carion2025sam} & 56.4 & \textbf{93.8} &- & - & - & - & - & - & - & - & - & - \\

\hline
\multicolumn{13}{l}{\textit{\textbf{VLM with Additions}}} \\
GLaMM (Vicuna-7B)~\citep{rasheed2024glamm} & x & x & x & x & 79.5 & 83.2 & 76.9 & 72.6 & 78.7 & 64.6 & 74.2 & 74.9 \\
VisionLLM v2 (Swin-T)~\citep{wu2024visionllm} & \textbf{56.7} & - & - & - & 87.9 & 91.2 & 84.3 & 77.6 & 83.8 & 70.2 & 82.9 & 84.1 \\
UFO (InternVL2.5-8B)~\citep{tang2025ufo} & 48.9 & x & x & x & \textbf{93.1} & \textbf{94.8} & \textbf{89.2} & \textbf{87.7} & \textbf{92.1} & \textbf{82.3} & \textbf{88.2} & \textbf{89.2}\\

\midrule
\hline
\multicolumn{13}{l}{\textit{\textbf{Standard VLM}}} \\
\rowcolor{blue!4}Griffon (LLama2-13B)~\citep{zhan2024griffon} & 24.8 & x & x & x & 90.1 & 93.4 & 86.1 & 84.8 & 90.5 & 77.8 & 86.1 & 87.2 \\
\rowcolor{blue!4}InternVL-3.5 (4B)~\citep{wang2025internvl35advancingopensourcemultimodal} & - & 79.4&77.6&66.4 & 92.5&94.3&88.2&87.6&92.3&81.6&89.6&89.3\\
\rowcolor{blue!4}Qwen3-VL (4B)~\citep{Qwen3-VL} & - & 78.4&79.0&64.0 & 90.7&92.2&86.7&82.9&89.4&75.6&87.3&87.7 \\
\rowcolor{blue!8}\textbf{Youtu-VL} (\textbf{4B, Ours}) & \textbf{47.1} (48.0$^*$) & \textbf{88.6}&\textbf{85.1}&\textbf{74.4} & \textbf{93.6}&\textbf{95.2}&\textbf{90.8}&\textbf{90.1}&\textbf{93.9}&\textbf{85.4}&\textbf{92.2}&\textbf{92.9} \\

\bottomrule
\end{tabular}}
\caption{\label{tab:loc_details}\textbf{Quantitative comparison on localization tasks}. We evaluate the performance on Object Detection (COCO 2017 {val} split), Counting (CountBench, TallyQA-Simple, TallyQA-Complex), and Visual Grounding (RefCOCO, RefCOCO+, RefCOCOg). The table compares our proposed Youtu-VL against Vision Specialist Models, Vision Generalist Models, and other Multimodal LLMs (both with architectural additions and standard setups). ``×'' denotes that the method is not applicable and ``-'' indicates results are not available. Best results in each setting are marked in bold. ``$^*$'' indicates the performance of the model obtained after fine-tuning exclusively on detection data starting from Stage 3.}
\end{table}

%% file: sec/a_qualitative_examples.tex
\newtcolorbox{PromptBox}[1]{
    colback=black!5!white,
    colframe=black!75!black,
    title={#1},
    fonttitle=\bfseries\large,
    enhanced,
    attach boxed title to top left={yshift=-2mm, xshift=2mm},
    boxrule=0.5mm,
    coltitle=white,
    subtitle style={
        colback=gray!50!white,
        boxrule=0pt,
        top=2pt, bottom=2pt
    },
    before upper={\parindent0pt} 
}


\section{Case Study}
\label{sec:examples}

\vspace{-1cm}
\begin{figure}[htbp]
    \centering
    \begin{PromptBox}{Object Detection}        
        \includegraphics[width=0.95\linewidth]{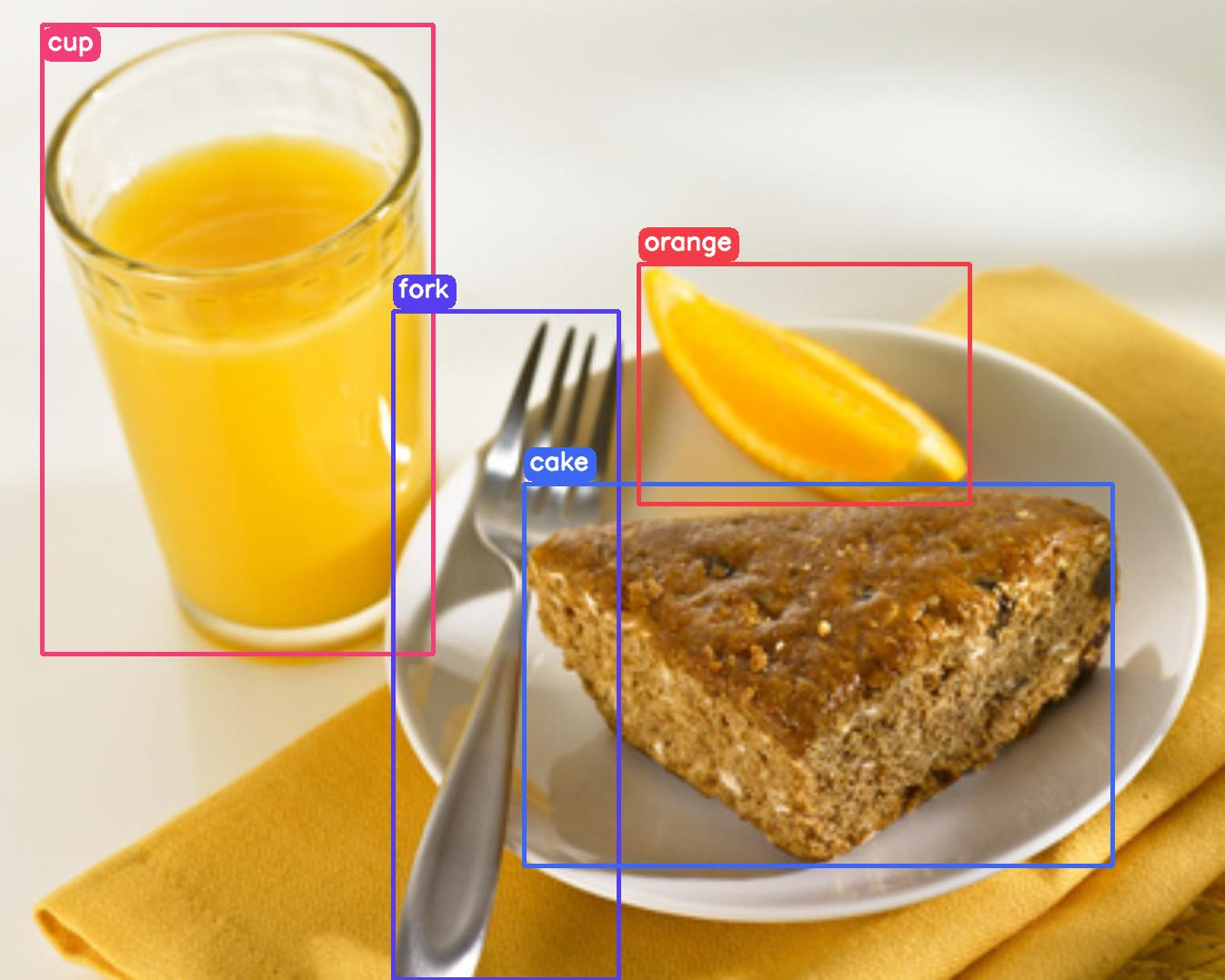}
        
        \textbf{User Instruction:} Detect all objects in the provided image.
        
        \tcbsubtitle{Model Response}
        <ref>cup</ref><box><x\_46><y\_27><x\_476><y\_719></box><ref>fork</ref><box><x\_432><y\_342>
        <x\_680><y\_1077></box><ref>cake</ref><box><x\_576><y\_532><x\_1223><y\_952></box>
        <ref>orange</ref><box><x\_702><y\_290><x\_1066><y\_554></box>
    \end{PromptBox}
    
    \caption{Qualitative example of object detection from the COCO dataset.}
    \label{fig:case_det}
\end{figure}

\begin{figure}[htbp]
    \centering
    \begin{PromptBox}{Object Detection}        
        \includegraphics[width=1\linewidth]{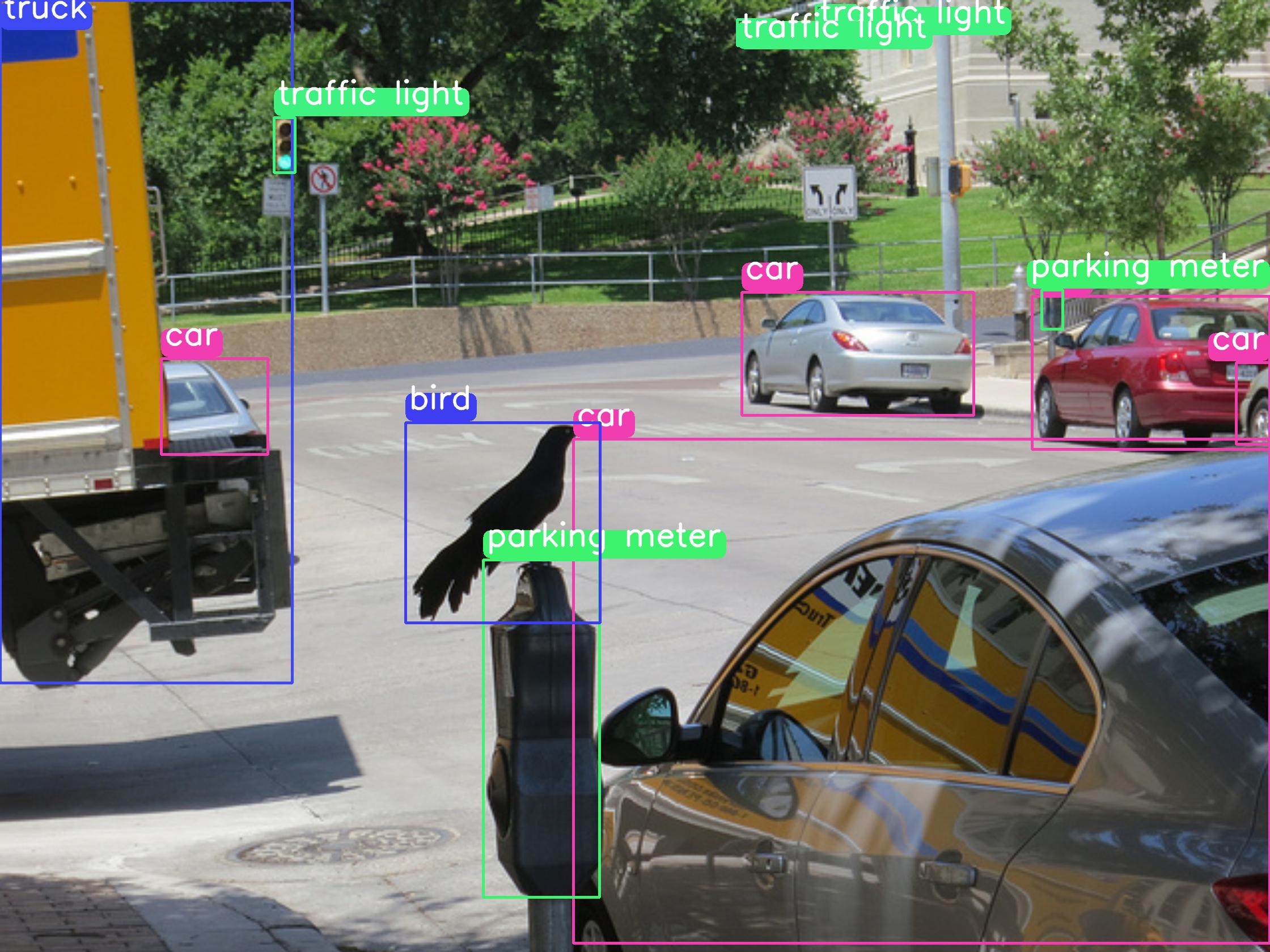}
        
        \textbf{User Instruction:} Detect all objects in the provided image.
        
        \tcbsubtitle{Model Response}
        <ref>truck</ref><box><x\_0><y\_0><x\_515><y\_1204></box><ref>car</ref><box><x\_1308> <y\_515><x\_1717><y\_732></box><box><x\_1820><y\_522><x\_2239><y\_792></box><box><x\_2180> <y\_639><x\_2239><y\_783></box><box><x\_284><y\_632><x\_472><y\_801></box><box> <x\_1011><y\_774><x\_2239><y\_1663></box><ref>traffic light</ref><box><x\_1439><y\_9> <x\_1509><y\_55></box><box><x\_1300><y\_34><x\_1368><y\_80></box><box><x\_483><y\_207> <x\_520><y\_304></box><ref>parking meter</ref><box><x\_1837><y\_511><x\_1874><y\_580> </box><box><x\_852><y\_987><x\_1057><y\_1582></box><ref>bird</ref><box><x\_715><y\_745> <x\_1058><y\_1098></box>
    \end{PromptBox}
    
    \caption{Qualitative example of object detection from the COCO dataset.}
    \label{fig:case_det}
\end{figure}

\begin{figure}[htbp]
    \centering
    \begin{PromptBox}{Open-World Detection}        
        \includegraphics[width=1\linewidth]{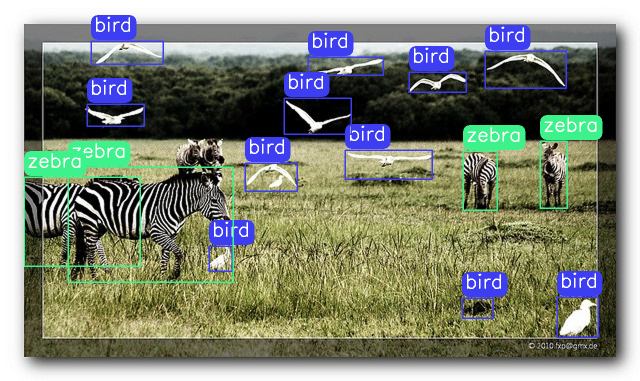}
        
        \textbf{User Instruction:} Detect all objects.
        
        \tcbsubtitle{Model Response}
        <ref>bird</ref><box><x\_209><y\_246><x\_232><y\_271></box><box><x\_557><y\_297><x\_598>
        <y\_337></box><box><x\_245><y\_163><x\_297><y\_191></box><box><x\_91><y\_41><x\_163>
        <y\_64></box><box><x\_87><y\_104><x\_144><y\_126></box><box><x\_345><y\_150><x\_432>
        <y\_179></box><box><x\_284><y\_98><x\_351><y\_134></box><box><x\_409><y\_72><x\_466>
        <y\_92></box><box><x\_485><y\_51><x\_566><y\_88></box><box><x\_462><y\_298><x\_493>
        <y\_318></box><box><x\_308><y\_57><x\_383><y\_75></box><ref>zebra</ref><box><x\_463>
        <y\_151><x\_497><y\_210></box><box><x\_540><y\_141><x\_567><y\_208></box><box><x\_68>
        <y\_167><x\_233><y\_282></box><box><x\_24><y\_177><x\_140><y\_266></box>
    \end{PromptBox}
    
    \caption{Qualitative example of open-world detection from the Objects365 dataset.}
    \label{fig:case_det_all}
\end{figure}

\begin{figure}[htbp]
    \centering
    \begin{PromptBox}{Detect-then-Count}        
        \centering
        \includegraphics[width=0.8\linewidth]{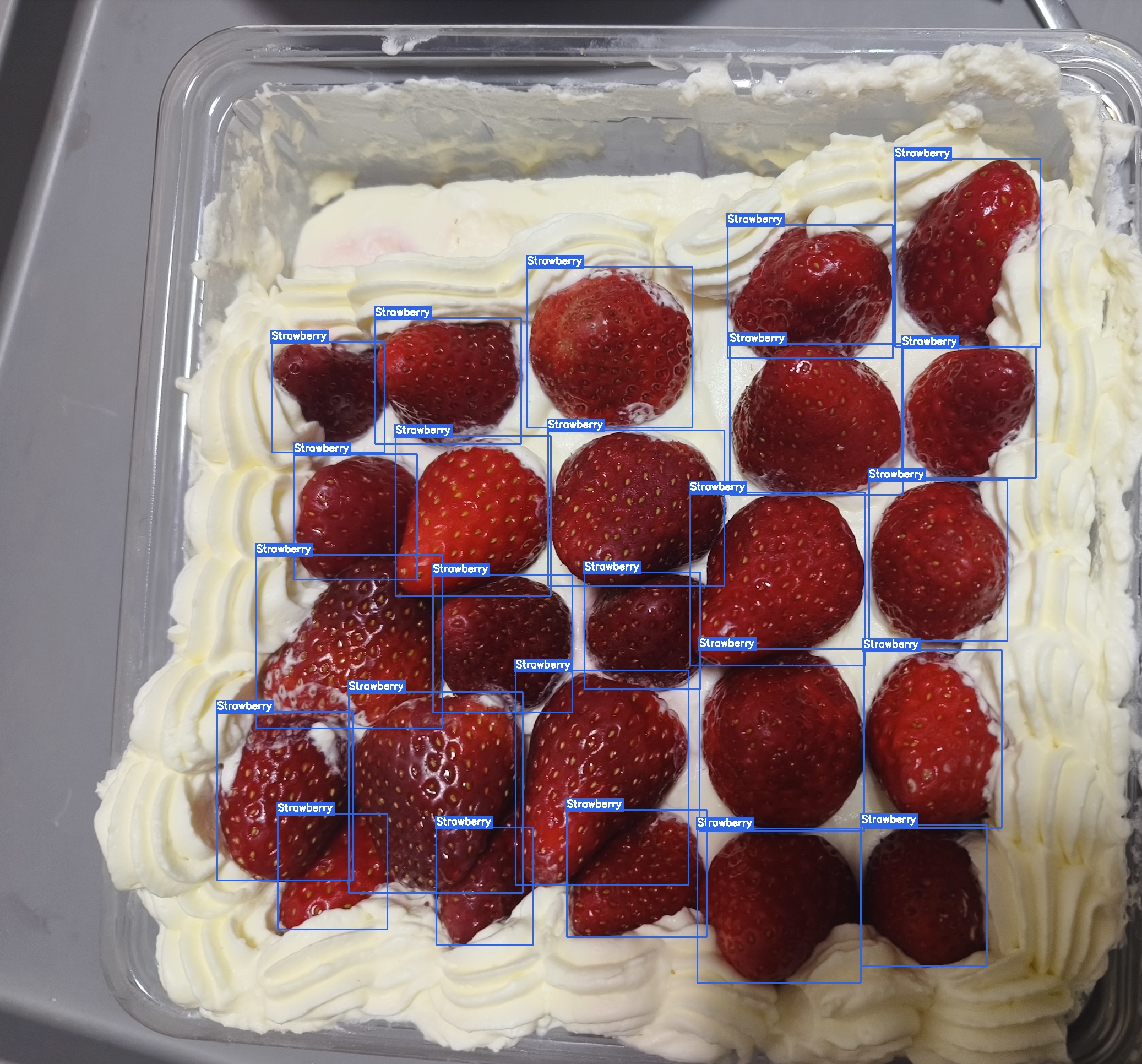}
        
        \textbf{User Instruction:} How many strawberry can you see in the image? First detect each one, then count the number.
        
        \tcbsubtitle{Model Response}
        <ref>Strawberry</ref><box><x\_1625><y\_288><x\_1889><y\_629></box><box><x\_1321><y\_408>
        <x\_1621><y\_650></box><box><x\_957><y\_484><x\_1257><y\_776></box><box><x\_681><y\_577>
        <x\_946><y\_806></box><box><x\_493><y\_620><x\_698><y\_821></box><box><x\_1325><y\_624>
        <x\_1640><y\_898></box><box><x\_1638><y\_630><x\_1881><y\_867></box><box><x\_995><y\_781>
        <x\_1315><y\_1063></box><box><x\_718><y\_791><x\_1000><y\_1082></box><box><x\_534><y\_824>
        <x\_757><y\_1052></box><box><x\_1578><y\_871><x\_1829><y\_1163></box><box><x\_1253><y\_894>
        <x\_1570><y\_1208></box><box><x\_465><y\_1007><x\_803><y\_1322></box><box><x\_1061><y\_1039>
        <x\_1271><y\_1251></box><box><x\_786><y\_1043><x\_1039><y\_1293></box><box><x\_1270><y\_1178>
        <x\_1569><y\_1509></box><box><x\_1568><y\_1180><x\_1819><y\_1503></box><box><x\_936><y\_1217>
        <x\_1250><y\_1606></box><box><x\_633><y\_1256><x\_949><y\_1621></box><box><x\_394><y\_1292>
        <x\_641><y\_1598></box><box><x\_1564><y\_1498><x\_1792><y\_1754></box><box><x\_1266><y\_1504>
        <x\_1563><y\_1784></box><box><x\_504><y\_1477><x\_703><y\_1687></box><box><x\_1029><y\_1470>
        <x\_1283><y\_1701></box><box><x\_792><y\_1502><x\_968><y\_1715></box>The answer is 25
    \end{PromptBox}
    
    \caption{Qualitative example of detect-then-count from a photo taken by an author from Youtu-VL Team.}
    \label{fig:case_dnc}
\end{figure}

\begin{figure}[htbp]
    \centering
    \begin{PromptBox}{Semantic Segmentation}        
        \includegraphics[width=0.95\linewidth]{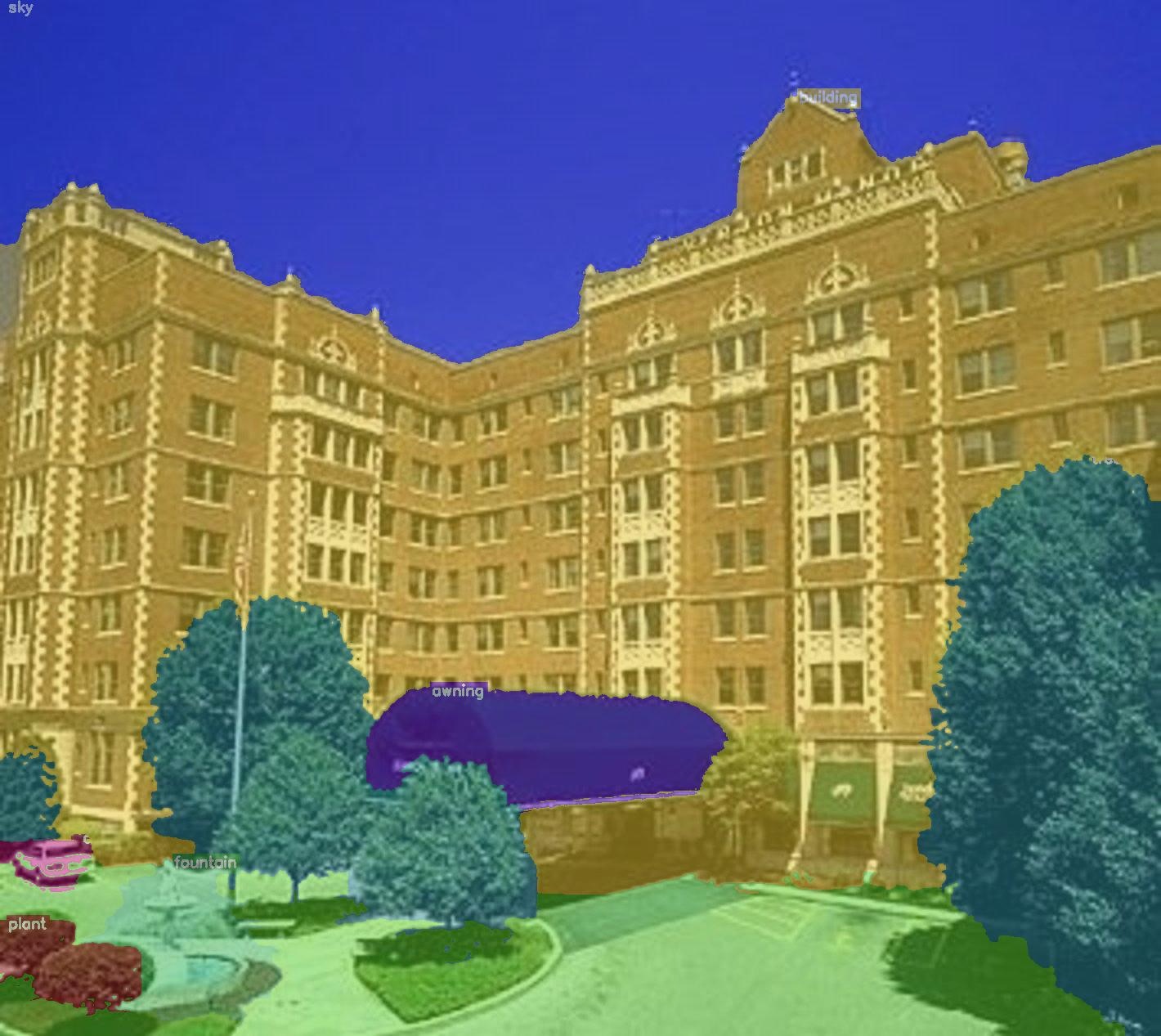}
        
        \textbf{User Instruction:} Label all river, tree, pot, tank, blind, bicycle, bannister, lake, screen, barrel, mirror, swimming pool, stairway, coffee table, bottle, bar, grass, plant, ball, fence, streetlight, sky, cushion, chair, microwave, stove, awning, rock, palm, sculpture, airplane, shower, conveyer belt, canopy, booth, radiator, flower, ashcan, fountain, house, bag, stage, minibike, screen door, escalator, oven, case, rug, plaything, poster, washer, pole, animal, swivel chair, ship, sand, bench, sidewalk, path, table, shelf, runway, counter, tower, bed , truck, basket, sofa, railing, crt screen, door, land, lamp, signboard, sconce, waterfall, ottoman, wardrobe, bus, countertop, kitchen island, column, hood, fan, glass, seat, base, stool, boat, hovel, bookcase, water, tent, traffic light, bathtub, tray, toilet, earth, sink, dishwasher, building, cabinet, pool table, apparel, painting, flag, trade name, armchair, sea, light, road, computer, monitor, desk, bridge, chandelier, fireplace, book, skyscraper, person, pillow, blanket, grandstand, windowpane, van, vase, towel, curtain, stairs, field, wall, hill, arcade machine, clock, pier, refrigerator, cradle, step, bulletin board, box, chest of drawers, mountain, food, television receiver, dirt track, plate, floor, buffet, ceiling, car in the image, without the background class.
        
        \tcbsubtitle{Model Response}
        The target categories include <ref>sky</ref><ref>road</ref><ref>car</ref><ref>tree</ref> <ref>building</ref><ref>awning</ref><ref>fountain</ref><ref>plant</ref>, numbered sequentially starting from 0, without the background class.<mask>RLE string</mark>
    \end{PromptBox}
    
    \caption{Qualitative example of semantic segmentation from the ADE20k dataset.}
    \label{fig:case_det}
\end{figure}

\begin{figure}[htbp]
    \centering
    \begin{PromptBox}{Semantic Segmentation}        
        \includegraphics[width=0.95\linewidth]{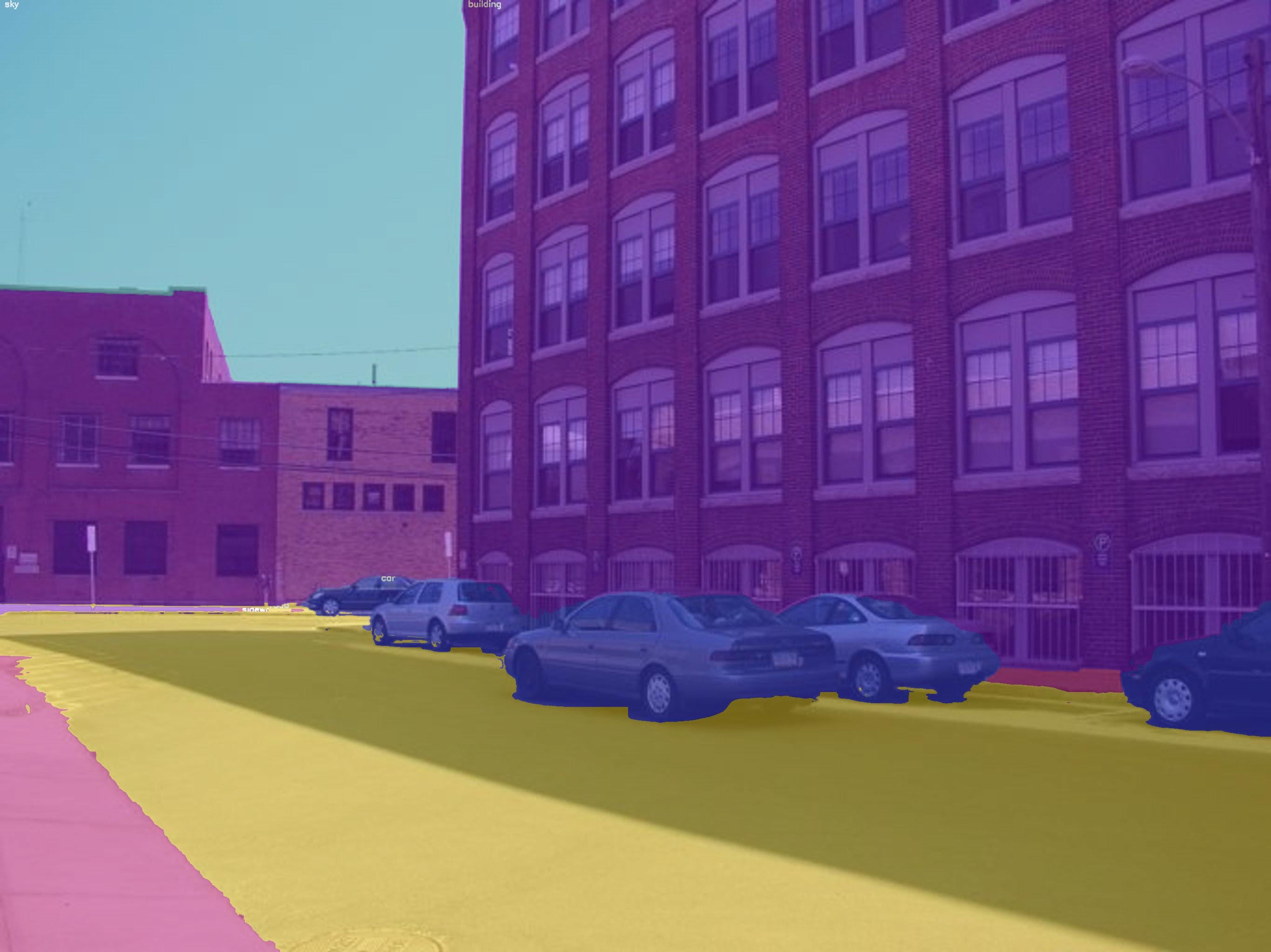}
        
        \textbf{User Instruction:} Create segmentation maps for these items: countertop, flower, hood, sky, shower, path, dishwasher, tank, skyscraper, railing, chest of drawers, arcade machine, painting, bar, building, sand, mountain, bannister, shelf, bus, bed , mirror, table, step, pier, refrigerator, floor, bag, minibike, conveyer belt, awning, canopy, runway, flag, book, basket, waterfall, water, car, apparel, bulletin board, pillow, signboard, washer, van, rug, plaything, chair, fountain, stairs, plate, animal, bottle, stove, door, blanket, stool, wall, buffet, trade name, fence, clock, box, dirt track, cushion, counter, radiator, seat, escalator, cradle, screen door, booth, ashcan, house, sofa, cabinet, coffee table, blind, fireplace, vase, swimming pool, bathtub, lamp, column, bicycle, toilet, hill, bridge, tower, ship, sea, pole, wardrobe, plant, tray, ceiling, windowpane, tree, television receiver, computer, light, oven, bench, grass, streetlight, monitor, river, bookcase, boat, rock, poster, person, food, curtain, sink, pool table, sidewalk, crt screen, earth, palm, lake, land, screen, stage, truck, desk, swivel chair, pot, traffic light, fan, chandelier, sculpture, ottoman, road, glass, hovel, base, towel, kitchen island, stairway, field, microwave, airplane, sconce, barrel, tent, case, grandstand, armchair, ball, without the background class.

        \tcbsubtitle{Model Response}
        The target categories include <ref>car</ref><ref>signboard</ref><ref>sidewalk</ref> <ref>sky</ref><ref>road</ref><ref>building</ref><ref>pole</ref><ref>streetlight</ref>, numbered sequentially starting from 0, without the background class.<mask>RLE string</mark>
    \end{PromptBox}
    
    \caption{Qualitative example of semantic segmentation from the ADE20k dataset.}
    \label{fig:case_det}
\end{figure}

\begin{figure}[htbp]
    \centering
    \begin{PromptBox}{Visual Grounding}        
        \includegraphics[width=1\linewidth]{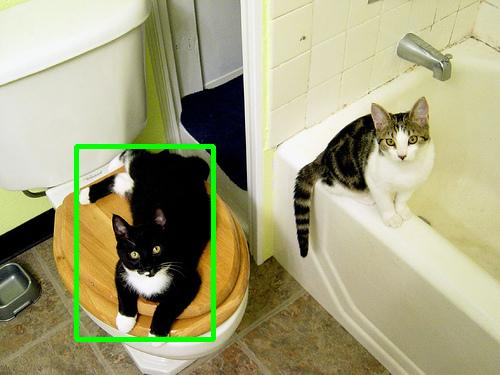}
        
        \textbf{User Instruction:} Please provide the bounding box coordinate of the region this sentence describes: the black cat sitting on a wooden toilet
        
        \tcbsubtitle{Model Response}
        <box><x\_76><y\_146><x\_213><y\_339></box>
    \end{PromptBox}
    
    \caption{Qualitative example of visual grounding from the RefCOCO dataset.}
    \label{fig:case_det}
\end{figure}

\begin{figure}[htbp]
    \centering
    \begin{PromptBox}{Visual Grounding}        
        \includegraphics[width=1\linewidth]{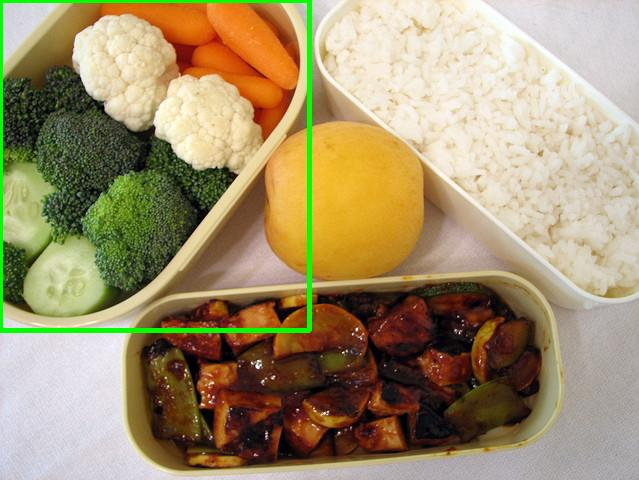}
        
        \textbf{User Instruction:} Please provide the bounding box coordinate of the region this sentence describes: the container holding fresh vegetables
        
        \tcbsubtitle{Model Response}
        <box><x\_0><y\_0><x\_309><y\_330></box>
    \end{PromptBox}
    
    \caption{Qualitative example of visual grounding from the RefCOCO dataset.}
    \label{fig:case_det}
\end{figure}

\begin{figure}[htbp]
    \centering
    \begin{PromptBox}{Referring Segmentation}                
        \includegraphics[width=1\linewidth]{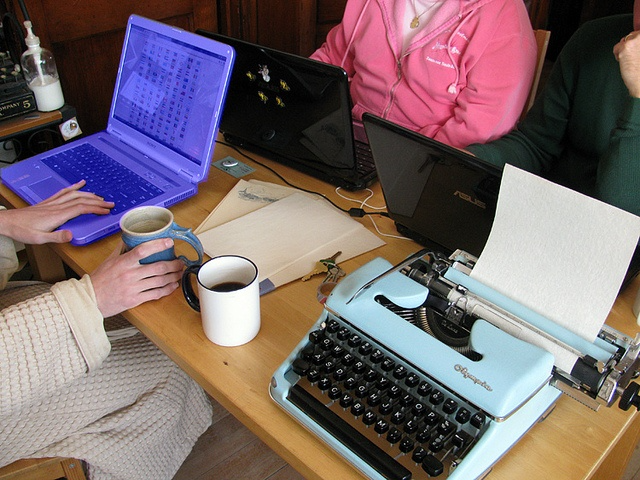}
        
        \textbf{User Instruction 1:} Please provide the bounding box coordinate of the region this sentence describes: the laptop left top

        \tcbsubtitle{Model Response 1}
        <box><x\_0><y\_13><x\_234><y\_245></box> \\
        (Operations: Draw box -> Crop image -> Resize) \\
        \ \\
        
        \textbf{User Instruction 2:} Segment the core target.
        
        \tcbsubtitle{Model Response 2}
        The results are 0 for <ref><BG></ref> and 1 for <ref><FG></ref>.<mask>RLE string</mask>
    \end{PromptBox}
    
    \caption{Qualitative example of referring expression segmentation from the RefCOCO dataset.}
    \label{fig:case_det}
\end{figure}

\begin{figure}[htbp]
    \centering
    \begin{PromptBox}{Depth Estimation}                
        \includegraphics[width=1\linewidth]{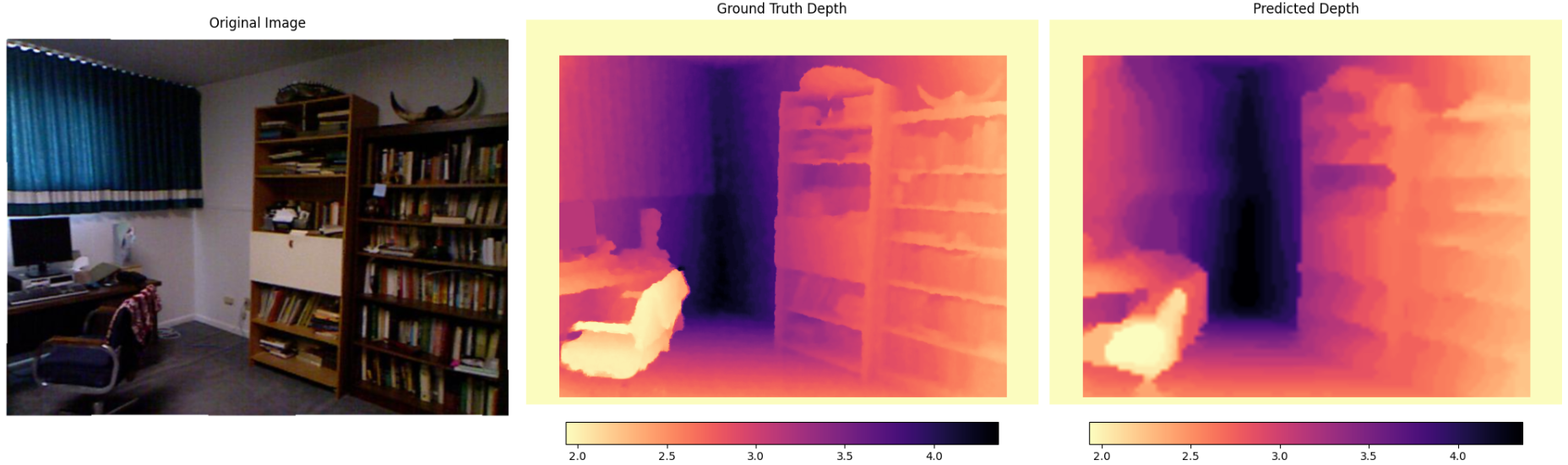}
        
        \textbf{User Instruction:} Please estimate the depth of this image from the NYUv2 dataset
        
        \tcbsubtitle{Model Response}
        This is the <depth>.<mask>RLE str</mask>        
    \end{PromptBox}
    
    \caption{Qualitative example of depth estimation from the NYUv2 dataset.}
    \label{fig:case_det}
\end{figure}

\begin{figure}[htbp]
    \centering
    \begin{PromptBox}{Depth Estimation}                
        \includegraphics[width=1\linewidth]{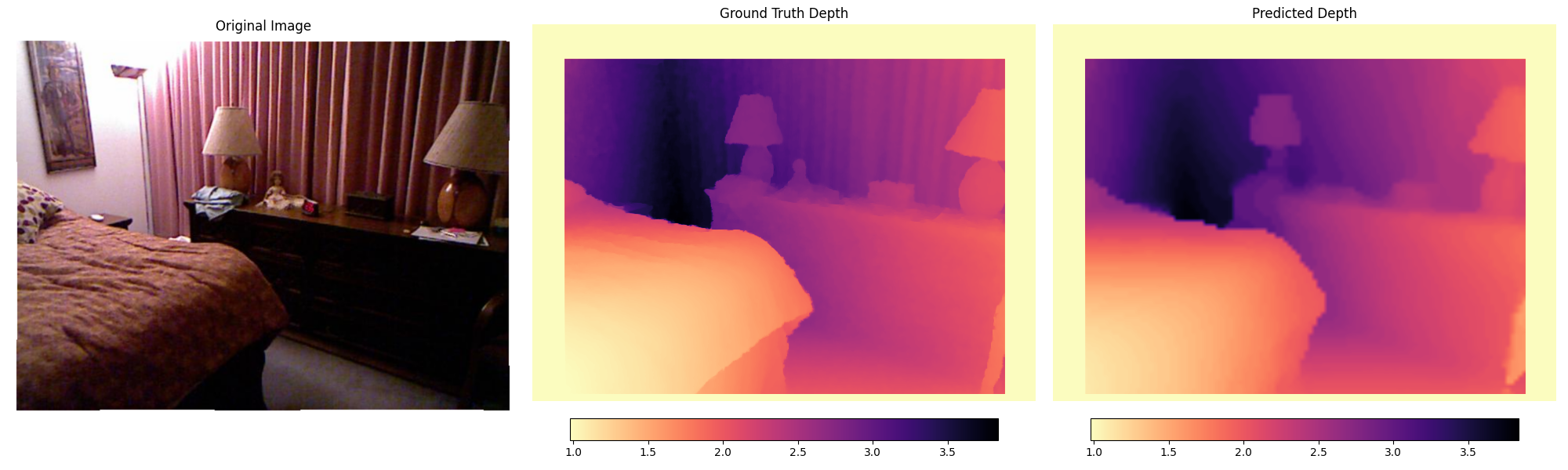}
        
        \textbf{User Instruction:} Please estimate the depth of this image from the NYUv2 dataset
        
        \tcbsubtitle{Model Response}
        This is the <depth>.<mask>RLE str</mask>        
    \end{PromptBox}
    
    \caption{Qualitative example of depth estimation from the NYUv2 dataset.}
    \label{fig:case_det}
\end{figure}

\begin{figure}[htbp]
    \begin{PromptBox}{Human Pose Estimation}                
        \includegraphics[width=1\linewidth]{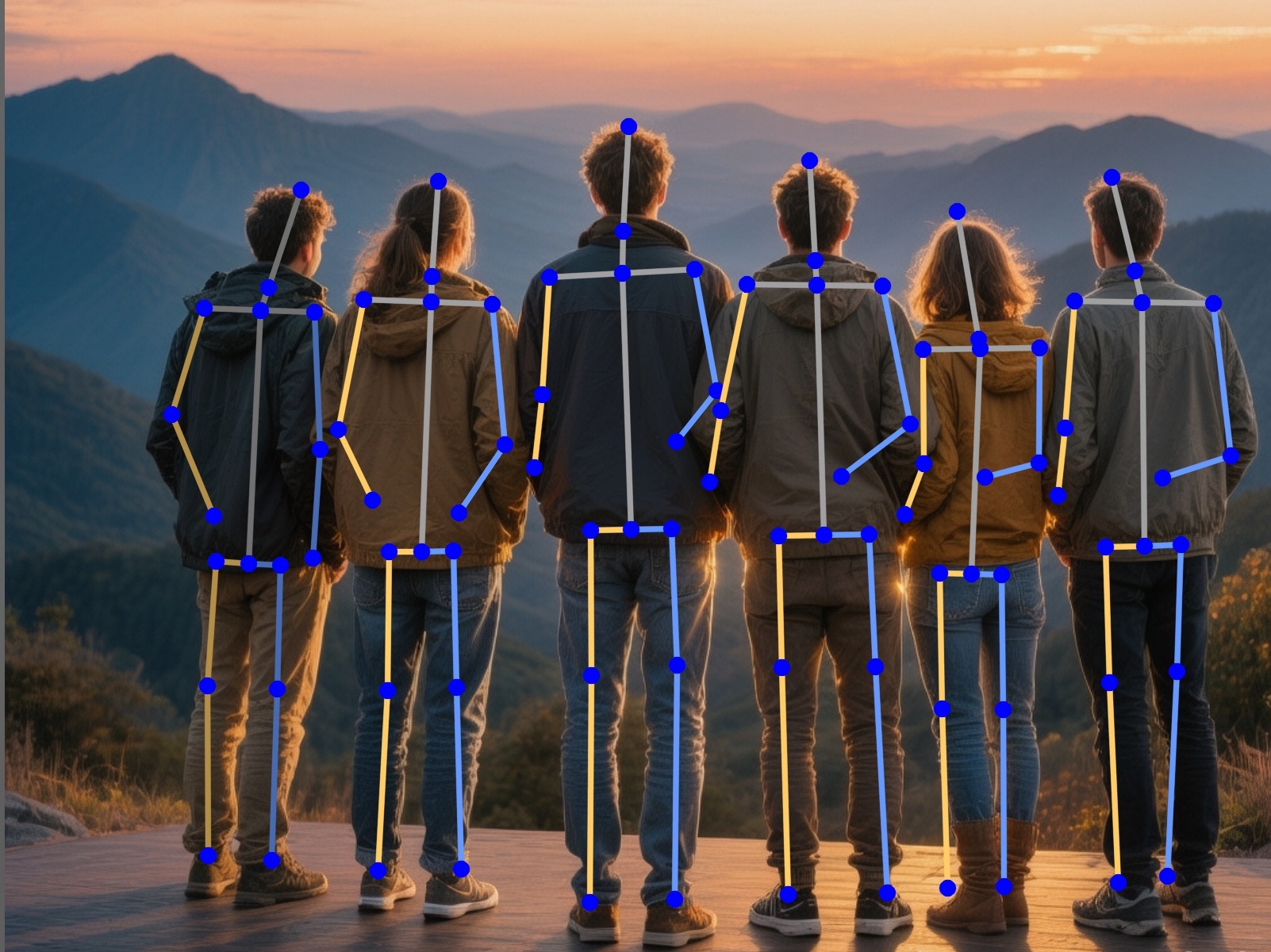}
        
        \textbf{User Instruction:} Detect all persons and their poses from the image within the class set of MPII Human Pose Dataset. The output format should be a JSON-like string, containing person instances. Each person instance is enclosed in <person>...</person> tags. Within each person instance, provide the bounding box using <box>...</box> tags and their 16 keypoints using <kpt>...</kpt> tags. The bounding box is defined by <x\_x1><y\_y1><x\_x2><y\_y2> tags, and each keypoint is defined by <x\_...><y\_...><v\_...> tags, where x, y are coordinates and v is visibility. The joints must be in this specific order: (0) right\_ankle, (1) right\_knee, (2) right\_hip, (3) left\_hip, (4) left\_knee, (5) left\_ankle, (6) pelvis, (7) thorax, (8) upper\_neck, (9) head\_top, (10) right\_wrist, (11) right\_elbow, (12) right\_shoulder, (13) left\_shoulder, (14) left\_elbow, (15) left\_wrist. Please output all detected persons with their bounding boxes and keypoints.
        
        \tcbsubtitle{Model Response}
<person><box><x\_306><y\_240><x\_410><y\_361></box><kpt><x\_342><y\_1084></kpt><kpt> <x\_349><y\_868></kpt><kpt><x\_353><y\_712></kpt><kpt><x\_272><y\_707></kpt><kpt><x\_261> <y\_864></kpt><kpt><x\_262><y\_1078></kpt><kpt><x\_313><y\_710></kpt><kpt> <x\_328><y\_391></kpt><kpt><x\_338><y\_362></kpt><kpt><x\_379><y\_239></kpt><kpt><x\_394> <y\_703></kpt><kpt><x\_403><y\_566></kpt><kpt><x\_396><y\_393></kpt><kpt> <x\_257><y\_388></kpt><kpt><x\_216><y\_522></kpt><kpt><x\_269>\\
... \\
<x\_1364><y\_224><x\_1468><y\_340></box><kpt><x\_1471><y\_1104></kpt><kpt><x\_1483> <y\_846></kpt><kpt><x\_1488><y\_686></kpt><kpt><x\_1393><y\_689></kpt><kpt><x\_1398> <y\_860></kpt><kpt><x\_1409><y\_1112></kpt><kpt><x\_1442><y\_688></kpt><kpt><x\_1439> <y\_381></kpt><kpt><x\_1430><y\_341></kpt><kpt><x\_1401><y\_223></kpt><kpt><x\_1465> <y\_602></kpt><kpt><x\_1551><y\_574></kpt><kpt><x\_1529><y\_382></kpt><kpt><x\_1354> <y\_379></kpt><kpt><x\_1342><y\_539></kpt><kpt><x\_1333><y\_624></kpt></person>
    \end{PromptBox}
    
    \caption{Qualitative example of human pose estimation from generated image (HY).}
    \label{fig:case_det}
\end{figure}


\begin{figure}[htbp]
    \centering
    
    \begin{PromptBox}{Creative Writing}                
        \includegraphics[width=0.5\linewidth]{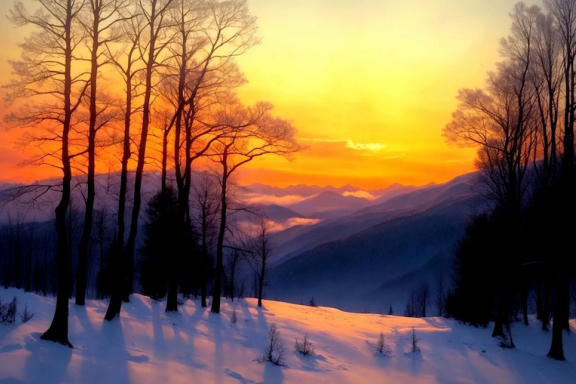}
        
        \vspace{0.5em}
        \textbf{User Instruction:} 
        \begin{CJK*}{UTF8}{gbsn}
        请根据这张图片，创作一首七言绝句    
        \end{CJK*}   
        
        \tcbsubtitle{Model Response}
        \begin{CJK*}{UTF8}{gbsn}
        雪岭晨曦染赤霞，\\
        枯枝剪影立寒坡。\\
        云海漫卷千山静，\\
        一抹金辉破夜韶。
        \end{CJK*}           
    \end{PromptBox}
    
    \caption{Qualitative example of visual-language creative generation. The model perceives the visual aesthetics of the snowy sunrise and composes a coherent Chinese poem matching the atmosphere.}
    \label{fig:case_poem}
\end{figure}

\begin{figure}[htbp]
    \begin{PromptBox}{Mathematical Reasoning}
        \includegraphics[width=0.65\linewidth]{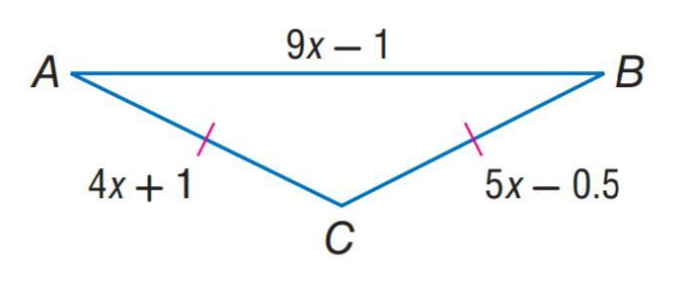}

        \textbf{User Instruction:} Find the length of AC in the isosceles triangle ABC.
        
        \tcbsubtitle{Model Response}        
        {\scriptsize
        
        \textbf{Phase 1: Comprehensive Information Extraction}
        \begin{itemize}
            \item \textbf{Textual Input}: The problem specifies triangle $ ABC $ is isosceles and asks for the length of $ AC $.
            \item \textbf{Visual Input}:
            \begin{itemize}
                \item Side $ AB = 9x - 1 $, side $ BC = 5x - 0.5 $, and side $ AC = 4x + 1 $.
                \item Pink tick marks on sides $ AC $ and $ BC $ indicate these two sides are equal (a standard geometric notation for congruent segments).
            \end{itemize}
        \end{itemize}

        \textbf{Phase 2: Strategic Problem Setup}
        \begin{itemize}
            \item \textbf{Goal}: Determine the value of $ x $ using the equality of $ AC $ and $ BC $, then compute $ AC $.
            \item \textbf{Assumption}: Since the tick marks confirm $ AC = BC $, this is the correct pair to equate.
            \item \textbf{Relevant Principle}: In an isosceles triangle, two sides are equal. Here, $ AC = BC $ implies $ 4x + 1 = 5x - 0.5 $.
        \end{itemize}

        \textbf{Phase 3: Rigorous Solution Execution}
        \begin{enumerate}
            \item \textbf{Set up the equation} based on the equality of $ AC $ and $ BC $:
            \[
            4x + 1 = 5x - 0.5
            \]
            \item \textbf{Solve for $ x $}: \\
            Subtract $ 4x $ from both sides:
            \[
            1 = x - 0.5
            \]
            Add $ 0.5 $ to both sides:
            \[
            x = 1.5
            \]
            \item \textbf{Substitute $ x = 1.5 $ into the expression for $ AC $}:
            \[
            AC = 4x + 1 = 4\times1.5 + 1 = 6 + 1 = 7
            \]
            \item \textbf{Verification}:
            \begin{itemize}
                \item Compute $ BC $ with $ x = 1.5 $:
                \[
                BC = 5x - 0.5 = 5\times1.5 - 0.5 = 7.5 - 0.5 = 7
                \]
                \item Both $ AC $ and $ BC $ equal 7, confirming the solution is consistent with the diagram's tick marks.
            \end{itemize}
        \end{enumerate}

        \textbf{Phase 4: Solution Validation}
        \begin{itemize}
            \item The computed value $ x = 1.5 $ satisfies the condition $ AC = BC $, as required by the visual notation.
            \item Substituting back confirms all side lengths align with the given expressions.
            \item No contradictions exist between the diagram and calculations.
        \end{itemize}

        Therefore, the final answer is \textbf{7}.
        
        }
    \end{PromptBox}
    
    \caption{Qualitative example of mathematical reasoning. The model successfully extracts geometric constraints (isosceles triangle properties) from the image, formulates the algebraic equation, and solves for the specific side length step-by-step.}
    \label{fig:case_math}
\end{figure}
\begin{figure}[htbp]
    \begin{PromptBox}{Chart \& Table Understanding}
        \includegraphics[width=0.5\linewidth]{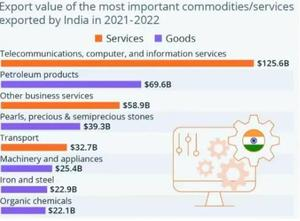}

        \vspace{0.5em}
        \textbf{User Instruction:} What is the average amount of goods where the second letter in the name is 'e'?
        
        \tcbsubtitle{Model Response}        
        {\scriptsize
        We are asked to find the \textbf{average amount of goods} where the \textbf{second letter in the name is `e'}.

        First, let's identify which items in the chart are \textbf{goods} (purple bars) and whose \textbf{name contains the letter `e' as the second letter}.
        
        Looking at the list:
        \begin{enumerate}
            \item \textbf{Petroleum products} --- purple bar $\rightarrow$ 69.6B \\
            Name: ``Petroleum'' $\rightarrow$ letters: P-E-T-O-L-I-M-E-N-T $\rightarrow$ second letter is \textbf{E}
            
            \item \textbf{Other business services} --- orange bar $\rightarrow$ not goods
            
            \item \textbf{Pearls, precious and semiprecious stones} --- purple bar $\rightarrow$ 39.3B \\
            Name: ``Pearls'' $\rightarrow$ P-E-A-R-L-S $\rightarrow$ second letter is \textbf{E}
            
            \item \textbf{Transport} --- orange bar $\rightarrow$ not goods
            
            \item \textbf{Machinery and appliances} --- purple bar $\rightarrow$ 25.4B \\
            Name: ``Machinery'' $\rightarrow$ M-A-C-H-I-R-E-Y $\rightarrow$ second letter is \textbf{A} \\
            ``appliances'' $\rightarrow$ A-P-P-L-I-C-A-T-I-O-N-S $\rightarrow$ second letter is \textbf{P} \\
            $\rightarrow$ So overall, second letter is \textbf{A} or \textbf{P}, not E
            
            \item \textbf{Iron and steel} --- purple bar $\rightarrow$ 22.9B \\
            Name: ``Iron'' $\rightarrow$ I-R-O-N $\rightarrow$ second letter is \textbf{R} \\
            ``steel'' $\rightarrow$ S-T-E-E-L $\rightarrow$ second letter is \textbf{T} \\
            $\rightarrow$ Not E
            
            \item \textbf{Organic chemicals} --- purple bar $\rightarrow$ 22.1B \\
            Name: ``Organic'' $\rightarrow$ O-R-G-A-N-I-C $\rightarrow$ second letter is \textbf{R} \\
            ``chemicals'' $\rightarrow$ C-H-E-M-I-C-A-L-S $\rightarrow$ second letter is \textbf{H} \\
            $\rightarrow$ Not E
        \end{enumerate}

        So only two goods meet the condition:
        \begin{itemize}
            \item \textbf{Petroleum products}: 69.6B
            \item \textbf{Pearls, precious and semiprecious stones}: 39.3B
        \end{itemize}

        Now, compute the average:
        \[
        \text{Average} = (69.6 + 39.3) / 2 = 108.9 / 2 = \textbf{54.45}
        \]

        \par\noindent\rule{\linewidth}{0.4pt} 
        
        \textbf{Final Answer: 54.45B}
        }
    \end{PromptBox}
    
    \caption{Qualitative example of chart understanding and multi-step reasoning. The model accurately interprets visual cues (color legends) to distinguish data categories, applies character-level text analysis to filter labels, and performs arithmetic operations on the extracted values to derive the answer.}
    \label{fig:case_chart}
\end{figure}

\begin{figure}[htbp]
    \begin{PromptBox}{Visual Perception}
        \includegraphics[width=0.5\linewidth]{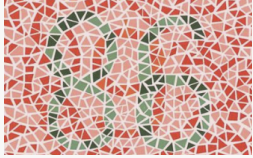}
        \vspace{0.5em}
        
        \textbf{User Instruction:} 
        This is a color vision test chart. Please carefully observe the numbers in the picture and answer in the following format: {Numbers: what you see}.
        \tcbsubtitle{Model Response}        
        {\scriptsize
        {Numbers: 86}
        }
    \end{PromptBox}
    
    \caption{Qualitative example of fine-grained visual perception. The model accurately recognizes the number hidden in the Ishihara color test plate, demonstrating the capability to distinguish subtle chromatic differences and patterns to identify the target information.}
    \label{fig:case_math}
\end{figure}
\begin{figure}[htbp]
    \begin{PromptBox}{Mathematical OCR \& LaTeX Conversion}
        \includegraphics[width=0.5\linewidth]{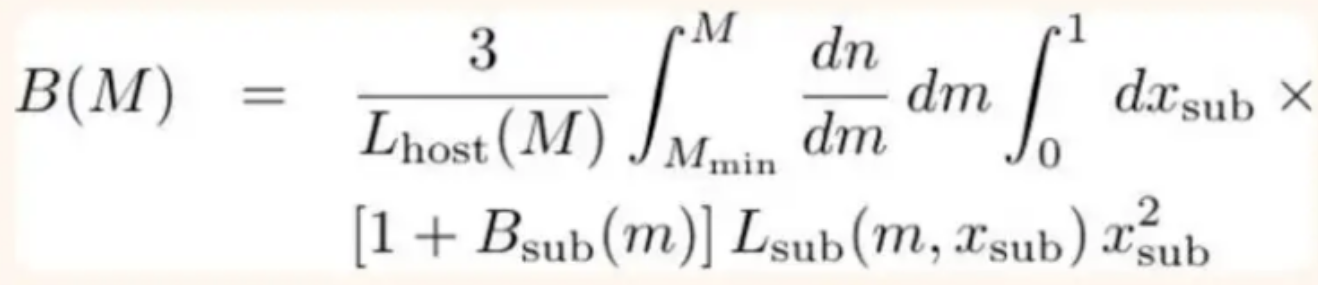}
        \vspace{0.5em}
        \begin{CJK*}{UTF8}{gbsn}

        \textbf{User Instruction:} 
        
        能把图中公式整理为latex格式吗
        \tcbsubtitle{Model Response}        
        {\scriptsize
        以下是图中公式整理后的 LaTeX 格式：
        $$
        B(M) = \frac{3}{L_{\text{host}}(M)} \int_{M_{\min}}^{M} \frac{dn}{dm} \, dm \int_{0}^{1} dx_{\text{sub}} \times 
        \left[ 1 + B_{\text{sub}}(m) \right] L_{\text{sub}}(m, x_{\text{sub}}) x_{\text{sub}}^2
        $$
        }
        \end{CJK*}
    \end{PromptBox}
    
    \caption{Qualitative example of mathematical formula recognition. The model accurately parses the complex spatial structure (integrals, fractions, subscripts) and specific symbols from the image, translating them into syntactically correct LaTeX code.}
    \label{fig:case_math}
\end{figure}

\begin{figure}[htbp]
    \centering
    \begin{PromptBox}{GUI Agent}                
        \includegraphics[width=0.95\linewidth]{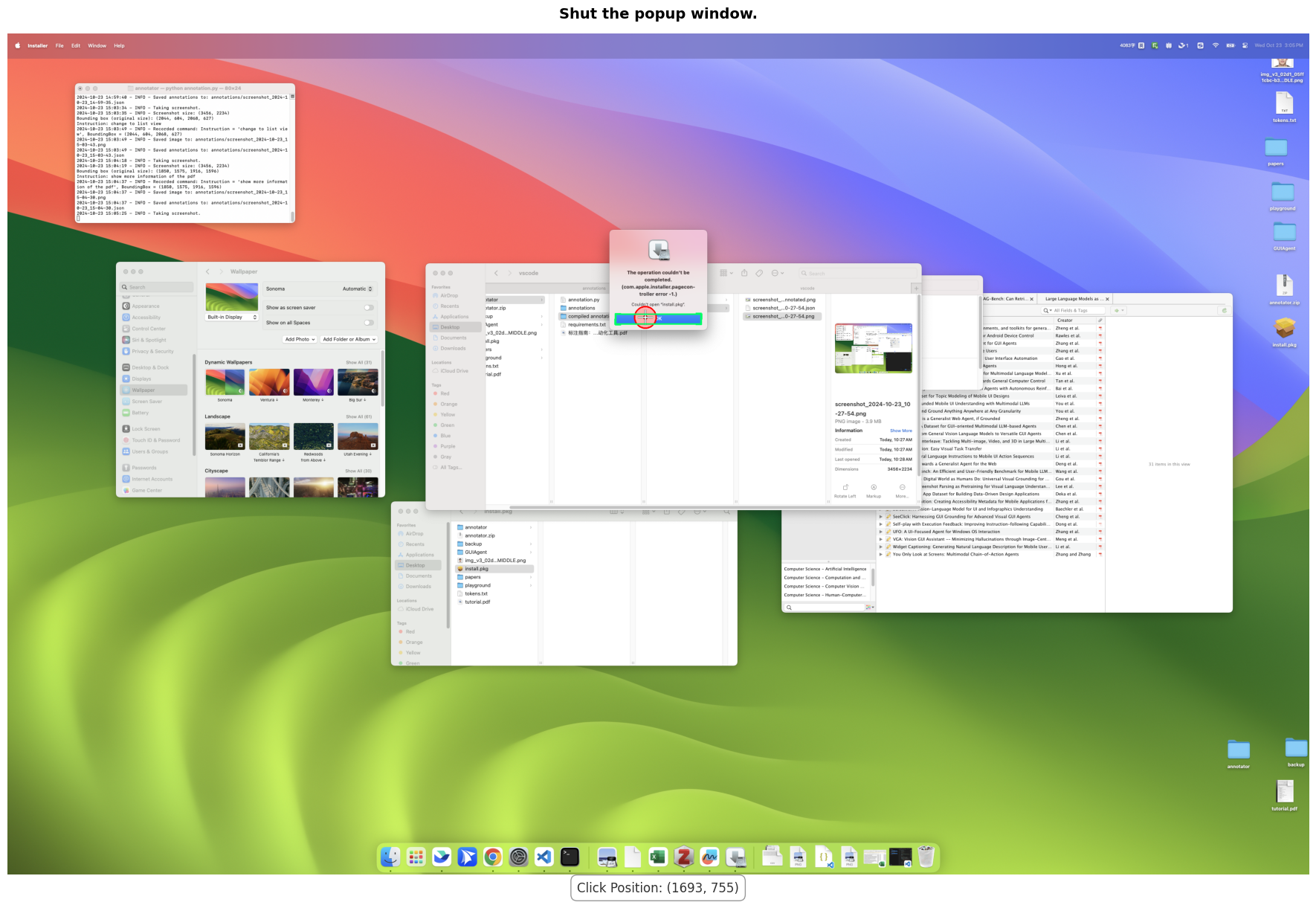}
        
        \vspace{0.5em}
        \textbf{User Instruction:} Shut the popup window.
        
        \tcbsubtitle{Model Response}
        Action: "Click the OK button on the error popup dialog to close it."\\
<tool\_call>\\
\{"name": "computer\_use", "arguments": \{"action": "left\_click", "coordinate": [1693, 755]\}\}\\
</tool\_call>
    \end{PromptBox}
    
    \caption{Qualitative example of GUI agent performance. The ground-truth target is highlighted in green, and the model’s click location is shown in red. The model demonstrates robust, precise interaction with interface elements across multiple windows. The screenshot is selected from the ScreenSpot Pro \cite{li2025screenspot} evaluation set.}
    \label{fig:case_poem}
\end{figure}

\begin{figure}[htbp]
    \centering
    \begin{PromptBox}{GUI Agent}                
        \includegraphics[width=0.95\linewidth]{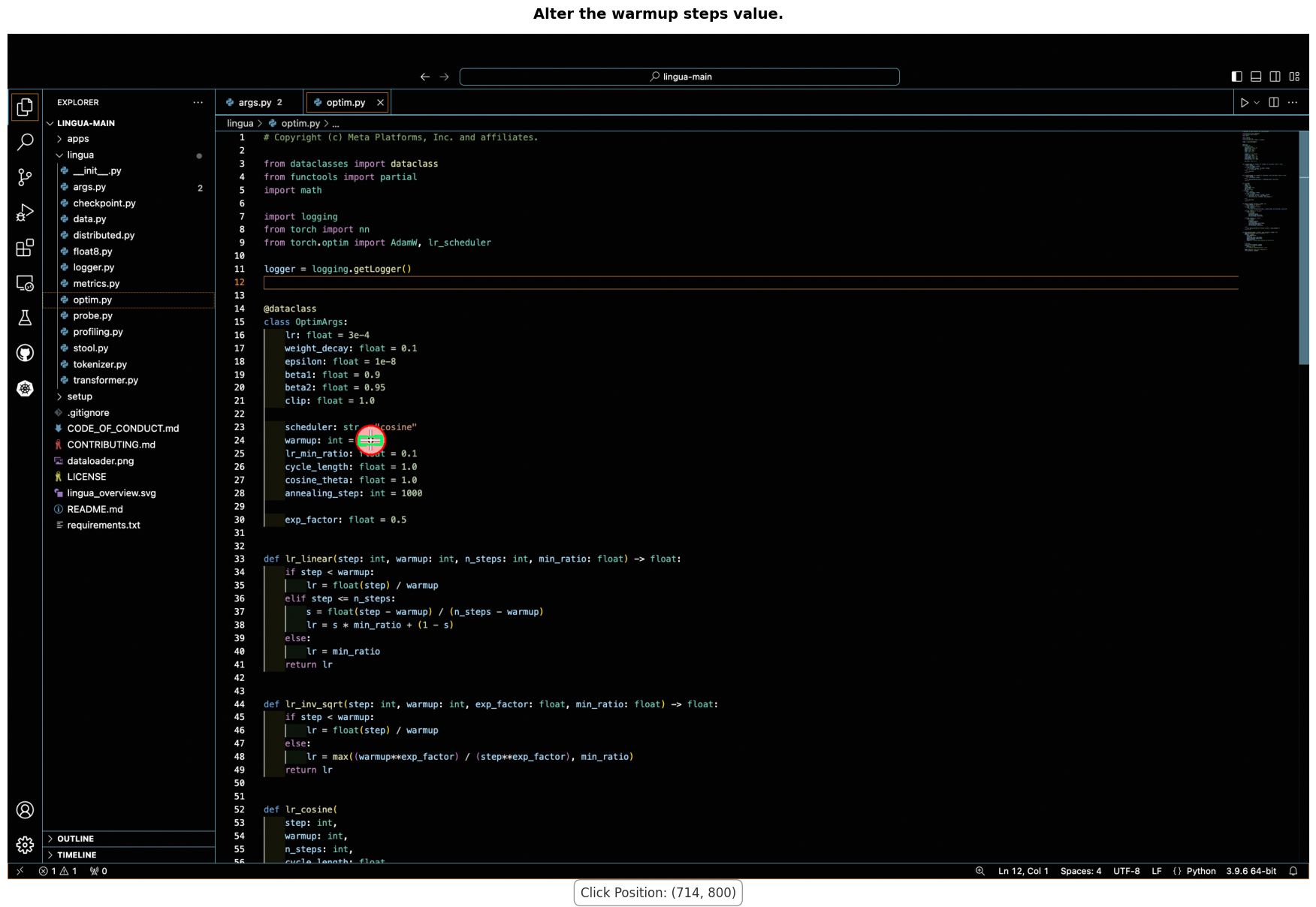}
        
        \vspace{0.5em}
        \textbf{User Instruction:} Alter the warmup steps value.
        
        \tcbsubtitle{Model Response}
        Action: "Click on the value '2000' for the warmup parameter in the OptimArgs dataclass to prepare for editing it."\\
<tool\_call>\\
\{"name": "computer\_use", "arguments": \{"action": "left\_click", "coordinate": [714, 800]\}\}\\
</tool\_call>
    \end{PromptBox}
    
    \caption{Qualitative example of GUI agent performance. This case demonstrates the model's ability to interpret instructions within a specific context and execute precise edits accordingly in the editor environment. The screenshot is selected from the ScreenSpot Pro \cite{li2025screenspot} evaluation set.}
    \label{fig:case_poem}
\end{figure}

%% file: main.bbl
\begin{thebibliography}{123}
\providecommand{\natexlab}[1]{#1}
\providecommand{\url}[1]{\texttt{#1}}
\expandafter\ifx\csname urlstyle\endcsname\relax
  \providecommand{\doi}[1]{doi: #1}\else
  \providecommand{\doi}{doi: \begingroup \urlstyle{rm}\Url}\fi

\bibitem[Bai et~al.(2025)Bai, Cai, Chen, Chen, Chen, Cheng, Deng, Ding, Gao, Ge, Ge, Guo, Huang, Huang, Huang, Hui, Jiang, Li, Li, Li, Li, Lin, Lin, Liu, Liu, Liu, Liu, Liu, Liu, Lu, Luo, Lv, Men, Meng, Ren, Ren, Song, Sun, Tang, Tu, Wan, Wang, Wang, Wang, Wang, Xie, Xu, Xu, Xu, Yang, Yang, Yang, Yang, Yu, Zhang, Zhang, Zhang, Zheng, Zhong, Zhou, Zhou, Zhou, Zhu, and Zhu]{Qwen3-VL}
Shuai Bai, Yuxuan Cai, Ruizhe Chen, Keqin Chen, Xionghui Chen, Zesen Cheng, Lianghao Deng, Wei Ding, Chang Gao, Chunjiang Ge, Wenbin Ge, Zhifang Guo, Qidong Huang, Jie Huang, Fei Huang, Binyuan Hui, Shutong Jiang, Zhaohai Li, Mingsheng Li, Mei Li, Kaixin Li, Zicheng Lin, Junyang Lin, Xuejing Liu, Jiawei Liu, Chenglong Liu, Yang Liu, Dayiheng Liu, Shixuan Liu, Dunjie Lu, Ruilin Luo, Chenxu Lv, Rui Men, Lingchen Meng, Xuancheng Ren, Xingzhang Ren, Sibo Song, Yuchong Sun, Jun Tang, Jianhong Tu, Jianqiang Wan, Peng Wang, Pengfei Wang, Qiuyue Wang, Yuxuan Wang, Tianbao Xie, Yiheng Xu, Haiyang Xu, Jin Xu, Zhibo Yang, Mingkun Yang, Jianxin Yang, An~Yang, Bowen Yu, Fei Zhang, Hang Zhang, Xi~Zhang, Bo~Zheng, Humen Zhong, Jingren Zhou, Fan Zhou, Jing Zhou, Yuanzhi Zhu, and Ke~Zhu.
\newblock Qwen3-vl technical report.
\newblock \emph{arXiv preprint arXiv:2511.21631}, 2025.

\bibitem[Wang et~al.(2025)Wang, Gao, Gu, Pu, Cui, Wei, Liu, Jing, Ye, Shao, Wang, Chen, Zhang, Yang, Wang, Wei, Yin, Li, Cui, Chen, Ding, Tian, Wu, Xie, Li, Yang, Duan, Wang, Hou, Hao, Zhang, Li, Zhao, Duan, Deng, Fu, He, Wang, He, Shi, He, Xiong, Lv, Wu, Shao, Zhang, Deng, Qi, Ge, Guo, Zhang, Zhang, Cao, Lin, Tang, Gao, Huang, Gu, Lyu, Tang, Wang, Lv, Ouyang, Wang, Dou, Zhu, Lu, Lin, Dai, Su, Zhou, Chen, Qiao, Wang, and Luo]{wang2025internvl35advancingopensourcemultimodal}
Weiyun Wang, Zhangwei Gao, Lixin Gu, Hengjun Pu, Long Cui, Xingguang Wei, Zhaoyang Liu, Linglin Jing, Shenglong Ye, Jie Shao, Zhaokai Wang, Zhe Chen, Hongjie Zhang, Ganlin Yang, Haomin Wang, Qi~Wei, Jinhui Yin, Wenhao Li, Erfei Cui, Guanzhou Chen, Zichen Ding, Changyao Tian, Zhenyu Wu, Jingjing Xie, Zehao Li, Bowen Yang, Yuchen Duan, Xuehui Wang, Zhi Hou, Haoran Hao, Tianyi Zhang, Songze Li, Xiangyu Zhao, Haodong Duan, Nianchen Deng, Bin Fu, Yinan He, Yi~Wang, Conghui He, Botian Shi, Junjun He, Yingtong Xiong, Han Lv, Lijun Wu, Wenqi Shao, Kaipeng Zhang, Huipeng Deng, Biqing Qi, Jiaye Ge, Qipeng Guo, Wenwei Zhang, Songyang Zhang, Maosong Cao, Junyao Lin, Kexian Tang, Jianfei Gao, Haian Huang, Yuzhe Gu, Chengqi Lyu, Huanze Tang, Rui Wang, Haijun Lv, Wanli Ouyang, Limin Wang, Min Dou, Xizhou Zhu, Tong Lu, Dahua Lin, Jifeng Dai, Weijie Su, Bowen Zhou, Kai Chen, Yu~Qiao, Wenhai Wang, and Gen Luo.
\newblock Internvl3.5: Advancing open-source multimodal models in versatility, reasoning, and efficiency, 2025.
\newblock URL \url{https://arxiv.org/abs/2508.18265}.

\bibitem[Liu et~al.(2023{\natexlab{a}})Liu, Li, Wu, and Lee]{llava}
Haotian Liu, Chunyuan Li, Qingyang Wu, and Yong~Jae Lee.
\newblock Visual instruction tuning.
\newblock \emph{arXiv:2304.08485}, 2023{\natexlab{a}}.

\bibitem[Assran et~al.(2023)Assran, Duval, Misra, Bojanowski, Vincent, Rabbat, LeCun, and Ballas]{assran2023selfsupervisedlearningimagesjointembedding}
Mahmoud Assran, Quentin Duval, Ishan Misra, Piotr Bojanowski, Pascal Vincent, Michael Rabbat, Yann LeCun, and Nicolas Ballas.
\newblock Self-supervised learning from images with a joint-embedding predictive architecture, 2023.
\newblock URL \url{https://arxiv.org/abs/2301.08243}.

\bibitem[Lu et~al.(2026)Lu, Qin, Qiao, Li, Dai, Ke, He, Qiao, Yin, Sun, Wu, Liu, Liu, Tang, Lin, Kuang, Meng, Tang, Xi, Huang, Yang, Shen, Li, Zhang, Yu, An, Dong, Wang, Wang, Chen, Wen, Guo, Shen, Yu, Li, Li, Li, and Tan]{lu2026youtullmunlockingnativeagentic}
Junru Lu, Jiarui Qin, Lingfeng Qiao, Yinghui Li, Xinyi Dai, Bo~Ke, Jianfeng He, Ruizhi Qiao, Di~Yin, Xing Sun, Yunsheng Wu, Yinsong Liu, Shuangyin Liu, Mingkong Tang, Haodong Lin, Jiayi Kuang, Fanxu Meng, Xiaojuan Tang, Yunjia Xi, Junjie Huang, Haotong Yang, Zhenyi Shen, Yangning Li, Qianwen Zhang, Yifei Yu, Siyu An, Junnan Dong, Qiufeng Wang, Jie Wang, Keyu Chen, Wei Wen, Taian Guo, Zhifeng Shen, Daohai Yu, Jiahao Li, Ke~Li, Zongyi Li, and Xiaoyu Tan.
\newblock Youtu-llm: Unlocking the native agentic potential for lightweight large language models, 2026.
\newblock URL \url{https://arxiv.org/abs/2512.24618}.

\bibitem[Tschannen et~al.(2025)Tschannen, Gritsenko, Wang, Naeem, Alabdulmohsin, Parthasarathy, Evans, Beyer, Xia, Mustafa, Hénaff, Harmsen, Steiner, and Zhai]{tschannen2025siglip2multilingualvisionlanguage}
Michael Tschannen, Alexey Gritsenko, Xiao Wang, Muhammad~Ferjad Naeem, Ibrahim Alabdulmohsin, Nikhil Parthasarathy, Talfan Evans, Lucas Beyer, Ye~Xia, Basil Mustafa, Olivier Hénaff, Jeremiah Harmsen, Andreas Steiner, and Xiaohua Zhai.
\newblock Siglip 2: Multilingual vision-language encoders with improved semantic understanding, localization, and dense features, 2025.
\newblock URL \url{https://arxiv.org/abs/2502.14786}.

\bibitem[Su et~al.(2023)Su, Lu, Pan, Murtadha, Wen, and Liu]{su2023roformerenhancedtransformerrotary}
Jianlin Su, Yu~Lu, Shengfeng Pan, Ahmed Murtadha, Bo~Wen, and Yunfeng Liu.
\newblock Roformer: Enhanced transformer with rotary position embedding, 2023.
\newblock URL \url{https://arxiv.org/abs/2104.09864}.

\bibitem[Dao et~al.(2022)Dao, Fu, Ermon, Rudra, and Ré]{dao2022flashattentionfastmemoryefficientexact}
Tri Dao, Daniel~Y. Fu, Stefano Ermon, Atri Rudra, and Christopher Ré.
\newblock Flashattention: Fast and memory-efficient exact attention with io-awareness, 2022.
\newblock URL \url{https://arxiv.org/abs/2205.14135}.

\bibitem[Bai et~al.(2023)Bai, Bai, Yang, Wang, Tan, Wang, Lin, Zhou, and Zhou]{bai2023qwenvlversatilevisionlanguagemodel}
Jinze Bai, Shuai Bai, Shusheng Yang, Shijie Wang, Sinan Tan, Peng Wang, Junyang Lin, Chang Zhou, and Jingren Zhou.
\newblock Qwen-vl: A versatile vision-language model for understanding, localization, text reading, and beyond, 2023.
\newblock URL \url{https://arxiv.org/abs/2308.12966}.

\bibitem[Siméoni et~al.(2025)Siméoni, Vo, Seitzer, Baldassarre, Oquab, Jose, Khalidov, Szafraniec, Yi, Ramamonjisoa, Massa, Haziza, Wehrstedt, Wang, Darcet, Moutakanni, Sentana, Roberts, Vedaldi, Tolan, Brandt, Couprie, Mairal, Jégou, Labatut, and Bojanowski]{siméoni2025dinov3}
Oriane Siméoni, Huy~V. Vo, Maximilian Seitzer, Federico Baldassarre, Maxime Oquab, Cijo Jose, Vasil Khalidov, Marc Szafraniec, Seungeun Yi, Michaël Ramamonjisoa, Francisco Massa, Daniel Haziza, Luca Wehrstedt, Jianyuan Wang, Timothée Darcet, Théo Moutakanni, Leonel Sentana, Claire Roberts, Andrea Vedaldi, Jamie Tolan, John Brandt, Camille Couprie, Julien Mairal, Hervé Jégou, Patrick Labatut, and Piotr Bojanowski.
\newblock Dinov3, 2025.
\newblock URL \url{https://arxiv.org/abs/2508.10104}.

\bibitem[Shi et~al.(2025{\natexlab{a}})Shi, Luo, Ge, Yang, Shan, and Wang]{shi2025scalableimagetokenizationindex}
Fengyuan Shi, Zhuoyan Luo, Yixiao Ge, Yujiu Yang, Ying Shan, and Limin Wang.
\newblock Scalable image tokenization with index backpropagation quantization, 2025{\natexlab{a}}.
\newblock URL \url{https://arxiv.org/abs/2412.02692}.

\bibitem[Rasheed et~al.(2024)Rasheed, Maaz, Shaji, Shaker, Khan, Cholakkal, Anwer, Xing, Yang, and Khan]{rasheed2024glamm}
Hanoona Rasheed, Muhammad Maaz, Sahal Shaji, Abdelrahman Shaker, Salman Khan, Hisham Cholakkal, Rao~M Anwer, Eric Xing, Ming-Hsuan Yang, and Fahad~S Khan.
\newblock Glamm: Pixel grounding large multimodal model.
\newblock In \emph{Proceedings of the IEEE/CVF Conference on Computer Vision and Pattern Recognition}, pages 13009--13018, 2024.

\bibitem[Wu et~al.(2024{\natexlab{a}})Wu, Zhong, Xing, Lai, Liu, Chen, Wang, Zhu, Lu, Lu, et~al.]{wu2024visionllm}
Jiannan Wu, Muyan Zhong, Sen Xing, Zeqiang Lai, Zhaoyang Liu, Zhe Chen, Wenhai Wang, Xizhou Zhu, Lewei Lu, Tong Lu, et~al.
\newblock Visionllm v2: An end-to-end generalist multimodal large language model for hundreds of vision-language tasks.
\newblock \emph{Advances in Neural Information Processing Systems}, 37:\penalty0 69925--69975, 2024{\natexlab{a}}.

\bibitem[Liu et~al.(2025)Liu, Ma, Pu, Qi, Wu, Shan, and Chen]{liu2025unipixel}
Ye~Liu, Zongyang Ma, Junfu Pu, Zhongang Qi, Yang Wu, Ying Shan, and Chang~Wen Chen.
\newblock Unipixel: Unified object referring and segmentation for pixel-level visual reasoning.
\newblock \emph{arXiv preprint arXiv:2509.18094}, 2025.

\bibitem[Tang et~al.(2025)Tang, Xie, Wang, Bao, Weng, Li, Zheng, and Wang]{tang2025ufo}
Hao Tang, Chenwei Xie, Haiyang Wang, Xiaoyi Bao, Tingyu Weng, Pandeng Li, Yun Zheng, and Liwei Wang.
\newblock Ufo: A unified approach to fine-grained visual perception via open-ended language interface.
\newblock \emph{arXiv preprint arXiv:2503.01342}, 2025.

\bibitem[Li et~al.(2025{\natexlab{a}})Li, Wang, Ding, Wang, and Li]{Li_2025_ICCV}
Yi~Li, Hualiang Wang, Xinpeng Ding, Haonan Wang, and Xiaomeng Li.
\newblock Token activation map to visually explain multimodal llms.
\newblock In \emph{Proceedings of the IEEE/CVF International Conference on Computer Vision (ICCV)}, pages 48--58, October 2025{\natexlab{a}}.

\bibitem[Kr{\"a}henb{\"u}hl and Koltun(2011)]{krahenbuhl2011efficient}
Philipp Kr{\"a}henb{\"u}hl and Vladlen Koltun.
\newblock Efficient inference in fully connected crfs with gaussian edge potentials.
\newblock \emph{Advances in neural information processing systems}, 24, 2011.

\bibitem[Shrivastava et~al.(2016)Shrivastava, Gupta, and Girshick]{shrivastava2016training}
Abhinav Shrivastava, Abhinav Gupta, and Ross Girshick.
\newblock Training region-based object detectors with online hard example mining.
\newblock In \emph{Proceedings of the IEEE conference on computer vision and pattern recognition}, pages 761--769, 2016.

\bibitem[Radford et~al.(2021)Radford, Kim, Hallacy, Ramesh, Goh, Agarwal, Sastry, Askell, Mishkin, Clark, et~al.]{radford2021learning}
Alec Radford, Jong~Wook Kim, Chris Hallacy, Aditya Ramesh, Gabriel Goh, Sandhini Agarwal, Girish Sastry, Amanda Askell, Pamela Mishkin, Jack Clark, et~al.
\newblock Learning transferable visual models from natural language supervision.
\newblock In \emph{International conference on machine learning}, pages 8748--8763. PmLR, 2021.

\bibitem[Wei et~al.(2025)Wei, Wang, Ma, Mei, Chen, Jin, and Rao]{wei2025hq}
Zhixiang Wei, Guangting Wang, Xiaoxiao Ma, Ke~Mei, Huaian Chen, Yi~Jin, and Fengyun Rao.
\newblock Hq-clip: Leveraging large vision-language models to create high-quality image-text datasets and clip models.
\newblock In \emph{Proceedings of the IEEE/CVF International Conference on Computer Vision}, pages 22447--22456, 2025.

\bibitem[Fang et~al.(2023)Fang, Jose, Jain, Schmidt, Toshev, and Shankar]{fang2023data}
Alex Fang, Albin~Madappally Jose, Amit Jain, Ludwig Schmidt, Alexander Toshev, and Vaishaal Shankar.
\newblock Data filtering networks.
\newblock \emph{arXiv preprint arXiv:2309.17425}, 2023.

\bibitem[Chuang et~al.(2025)Chuang, Li, Wang, Yeh, Lyu, Raghavendra, Glass, Huang, Weston, Zettlemoyer, et~al.]{chuang2025meta}
Yung-Sung Chuang, Yang Li, Dong Wang, Ching-Feng Yeh, Kehan Lyu, Ramya Raghavendra, James Glass, Lifei Huang, Jason Weston, Luke Zettlemoyer, et~al.
\newblock Meta clip 2: A worldwide scaling recipe.
\newblock \emph{arXiv preprint arXiv:2507.22062}, 2025.

\bibitem[Balestriero et~al.(2025)Balestriero, Ballas, Rabbat, and LeCun]{balestriero2025gaussianembeddingsjepassecretly}
Randall Balestriero, Nicolas Ballas, Mike Rabbat, and Yann LeCun.
\newblock Gaussian embeddings: How jepas secretly learn your data density, 2025.
\newblock URL \url{https://arxiv.org/abs/2510.05949}.

\bibitem[Lin et~al.(2025)Lin, Tan, Qin, Xu, Shi, Li, Li, Cai, Cai, Fu, et~al.]{lin2025cuarewardbench}
Haojia Lin, Xiaoyu Tan, Yulei Qin, Zihan Xu, Yuchen Shi, Zongyi Li, Gang Li, Shaofei Cai, Siqi Cai, Chaoyou Fu, et~al.
\newblock Cuarewardbench: A benchmark for evaluating reward models on computer-using agent.
\newblock \emph{arXiv preprint arXiv:2510.18596}, 2025.

\bibitem[Xie et~al.(2024)Xie, Zhang, Chen, Li, Zhao, Cao, Hua, Cheng, Shin, Lei, et~al.]{xie2024osworld}
Tianbao Xie, Danyang Zhang, Jixuan Chen, Xiaochuan Li, Siheng Zhao, Ruisheng Cao, Toh~J Hua, Zhoujun Cheng, Dongchan Shin, Fangyu Lei, et~al.
\newblock Osworld: Benchmarking multimodal agents for open-ended tasks in real computer environments.
\newblock \emph{Advances in Neural Information Processing Systems}, 37:\penalty0 52040--52094, 2024.

\bibitem[Shi et~al.(2025{\natexlab{b}})Shi, Cai, Cai, Xu, Chen, Qin, Zhou, Fei, Qiu, Tan, et~al.]{shi2025youtu}
Yuchen Shi, Yuzheng Cai, Siqi Cai, Zihan Xu, Lichao Chen, Yulei Qin, Zhijian Zhou, Xiang Fei, Chaofan Qiu, Xiaoyu Tan, et~al.
\newblock Youtu-agent: Scaling agent productivity with automated generation and hybrid policy optimization.
\newblock \emph{arXiv preprint arXiv:2512.24615}, 2025{\natexlab{b}}.

\bibitem[DeepSeek-AI et~al.(2025)DeepSeek-AI, Liu, Feng, Xue, Wang, Wu, Lu, Zhao, Deng, Zhang, Ruan, Dai, Guo, Yang, Chen, Ji, Li, Lin, Dai, Luo, Hao, Chen, Li, Zhang, Bao, Xu, Wang, Zhang, Ding, Xin, Gao, Li, Qu, Cai, Liang, Guo, Ni, Li, Wang, Chen, Chen, Yuan, Qiu, Li, Song, Dong, Hu, Gao, Guan, Huang, Yu, Wang, Zhang, Xu, Xia, Zhao, Wang, Zhang, Li, Wang, Zhang, Zhang, Tang, Li, Tian, Huang, Wang, Zhang, Wang, Zhu, Chen, Du, Chen, Jin, Ge, Zhang, Pan, Wang, Xu, Zhang, Chen, Li, Lu, Zhou, Chen, Wu, Ye, Ye, Ma, Wang, Zhou, Yu, Zhou, Pan, Wang, Yun, Pei, Sun, Xiao, Zeng, Zhao, An, Liu, Liang, Gao, Yu, Zhang, Li, Jin, Wang, Bi, Liu, Wang, Shen, Chen, Zhang, Chen, Nie, Sun, Wang, Cheng, Liu, Xie, Liu, Yu, Song, Shan, Zhou, Yang, Li, Su, Lin, Li, Wang, Wei, Zhu, Zhang, Xu, Xu, Huang, Li, Zhao, Sun, Li, Wang, Yu, Zheng, Zhang, Shi, Xiong, He, Tang, Piao, Wang, Tan, Ma, Liu, Guo, Wu, Ou, Zhu, Wang, Gong, Zou, He, Zha, Xiong, Ma, Yan, Luo, You, Liu, Zhou, Wu, Ren, Ren, Sha, Fu, Xu, Huang, Zhang, Xie, Zhang, Hao,
  Gou, Ma, Yan, Shao, Xu, Wu, Zhang, Li, Gu, Zhu, Liu, Li, Xie, Song, Gao, and Pan]{deepseekai2025deepseekv3technicalreport}
DeepSeek-AI, Aixin Liu, Bei Feng, Bing Xue, Bingxuan Wang, Bochao Wu, Chengda Lu, Chenggang Zhao, Chengqi Deng, Chenyu Zhang, Chong Ruan, Damai Dai, Daya Guo, Dejian Yang, Deli Chen, Dongjie Ji, Erhang Li, Fangyun Lin, Fucong Dai, Fuli Luo, Guangbo Hao, Guanting Chen, Guowei Li, H.~Zhang, Han Bao, Hanwei Xu, Haocheng Wang, Haowei Zhang, Honghui Ding, Huajian Xin, Huazuo Gao, Hui Li, Hui Qu, J.~L. Cai, Jian Liang, Jianzhong Guo, Jiaqi Ni, Jiashi Li, Jiawei Wang, Jin Chen, Jingchang Chen, Jingyang Yuan, Junjie Qiu, Junlong Li, Junxiao Song, Kai Dong, Kai Hu, Kaige Gao, Kang Guan, Kexin Huang, Kuai Yu, Lean Wang, Lecong Zhang, Lei Xu, Leyi Xia, Liang Zhao, Litong Wang, Liyue Zhang, Meng Li, Miaojun Wang, Mingchuan Zhang, Minghua Zhang, Minghui Tang, Mingming Li, Ning Tian, Panpan Huang, Peiyi Wang, Peng Zhang, Qiancheng Wang, Qihao Zhu, Qinyu Chen, Qiushi Du, R.~J. Chen, R.~L. Jin, Ruiqi Ge, Ruisong Zhang, Ruizhe Pan, Runji Wang, Runxin Xu, Ruoyu Zhang, Ruyi Chen, S.~S. Li, Shanghao Lu, Shangyan Zhou, Shanhuang
  Chen, Shaoqing Wu, Shengfeng Ye, Shengfeng Ye, Shirong Ma, Shiyu Wang, Shuang Zhou, Shuiping Yu, Shunfeng Zhou, Shuting Pan, T.~Wang, Tao Yun, Tian Pei, Tianyu Sun, W.~L. Xiao, Wangding Zeng, Wanjia Zhao, Wei An, Wen Liu, Wenfeng Liang, Wenjun Gao, Wenqin Yu, Wentao Zhang, X.~Q. Li, Xiangyue Jin, Xianzu Wang, Xiao Bi, Xiaodong Liu, Xiaohan Wang, Xiaojin Shen, Xiaokang Chen, Xiaokang Zhang, Xiaosha Chen, Xiaotao Nie, Xiaowen Sun, Xiaoxiang Wang, Xin Cheng, Xin Liu, Xin Xie, Xingchao Liu, Xingkai Yu, Xinnan Song, Xinxia Shan, Xinyi Zhou, Xinyu Yang, Xinyuan Li, Xuecheng Su, Xuheng Lin, Y.~K. Li, Y.~Q. Wang, Y.~X. Wei, Y.~X. Zhu, Yang Zhang, Yanhong Xu, Yanhong Xu, Yanping Huang, Yao Li, Yao Zhao, Yaofeng Sun, Yaohui Li, Yaohui Wang, Yi~Yu, Yi~Zheng, Yichao Zhang, Yifan Shi, Yiliang Xiong, Ying He, Ying Tang, Yishi Piao, Yisong Wang, Yixuan Tan, Yiyang Ma, Yiyuan Liu, Yongqiang Guo, Yu~Wu, Yuan Ou, Yuchen Zhu, Yuduan Wang, Yue Gong, Yuheng Zou, Yujia He, Yukun Zha, Yunfan Xiong, Yunxian Ma, Yuting Yan, Yuxiang
  Luo, Yuxiang You, Yuxuan Liu, Yuyang Zhou, Z.~F. Wu, Z.~Z. Ren, Zehui Ren, Zhangli Sha, Zhe Fu, Zhean Xu, Zhen Huang, Zhen Zhang, Zhenda Xie, Zhengyan Zhang, Zhewen Hao, Zhibin Gou, Zhicheng Ma, Zhigang Yan, Zhihong Shao, Zhipeng Xu, Zhiyu Wu, Zhongyu Zhang, Zhuoshu Li, Zihui Gu, Zijia Zhu, Zijun Liu, Zilin Li, Ziwei Xie, Ziyang Song, Ziyi Gao, and Zizheng Pan.
\newblock Deepseek-v3 technical report, 2025.
\newblock URL \url{https://arxiv.org/abs/2412.19437}.

\bibitem[Oquab et~al.(2023)Oquab, Darcet, Moutakanni, Vo, Szafraniec, Khalidov, Fernandez, Haziza, Massa, El-Nouby, et~al.]{oquab2023dinov2}
Maxime Oquab, Timoth{\'e}e Darcet, Th{\'e}o Moutakanni, Huy Vo, Marc Szafraniec, Vasil Khalidov, Pierre Fernandez, Daniel Haziza, Francisco Massa, Alaaeldin El-Nouby, et~al.
\newblock Dinov2: Learning robust visual features without supervision.
\newblock \emph{arXiv preprint arXiv:2304.07193}, 2023.

\bibitem[Yang et~al.(2024)Yang, Yang, Zhang, Hui, Zheng, Yu, Li, Liu, Huang, et~al.]{qwen2.5}
An~Yang, Baosong Yang, Beichen Zhang, Binyuan Hui, Bo~Zheng, Bowen Yu, Chengyuan Li, Dayiheng Liu, Fei Huang, et~al.
\newblock Qwen2.5 technical report.
\newblock \emph{arXiv:2412.15115}, 2024.

\bibitem[Loshchilov and Hutter()]{loshchilovdecoupled}
Ilya Loshchilov and Frank Hutter.
\newblock Decoupled weight decay regularization.
\newblock In \emph{International Conference on Learning Representations}.

\bibitem[Yu et~al.(2025)Yu, Zhang, Zhu, Yuan, Zuo, Yue, Dai, Fan, Liu, Liu, et~al.]{yu2025dapo}
Qiying Yu, Zheng Zhang, Ruofei Zhu, Yufeng Yuan, Xiaochen Zuo, Yu~Yue, Weinan Dai, Tiantian Fan, Gaohong Liu, Lingjun Liu, et~al.
\newblock Dapo: An open-source llm reinforcement learning system at scale.
\newblock \emph{arXiv preprint arXiv:2503.14476}, 2025.

\bibitem[Qi et~al.(2025)Qi, Liu, Zhou, Pang, Du, Lee, and Lin]{qi2025defeating}
Penghui Qi, Zichen Liu, Xiangxin Zhou, Tianyu Pang, Chao Du, Wee~Sun Lee, and Min Lin.
\newblock Defeating the training-inference mismatch via fp16.
\newblock \emph{arXiv preprint arXiv:2510.26788}, 2025.

\bibitem[Li et~al.(2024{\natexlab{a}})Li, Zhang, Zhang, Pu, Du, Dong, Liu, Zhang, Zhang, Li, and Liu]{lmms_eval2024}
Bo~Li, Peiyuan Zhang, Kaichen Zhang, Fanyi Pu, Xinrun Du, Yuhao Dong, Haotian Liu, Yuanhan Zhang, Ge~Zhang, Chunyuan Li, and Ziwei Liu.
\newblock Lmms-eval: Accelerating the development of large multimoal models, March 2024{\natexlab{a}}.
\newblock URL \url{https://github.com/EvolvingLMMs-Lab/lmms-eval}.

\bibitem[Duan et~al.(2024)Duan, Yang, Qiao, Fang, Chen, Liu, Dong, Zang, Zhang, Wang, et~al.]{duan2024vlmevalkit}
Haodong Duan, Junming Yang, Yuxuan Qiao, Xinyu Fang, Lin Chen, Yuan Liu, Xiaoyi Dong, Yuhang Zang, Pan Zhang, Jiaqi Wang, et~al.
\newblock Vlmevalkit: An open-source toolkit for evaluating large multi-modality models.
\newblock In \emph{Proceedings of the 32nd ACM international conference on multimedia}, pages 11198--11201, 2024.

\bibitem[Zhang et~al.(2024{\natexlab{a}})Zhang, Li, Zhang, Pu, Cahyono, Hu, Liu, Zhang, Yang, Li, and Liu]{zhang2024lmmsevalrealitycheckevaluation}
Kaichen Zhang, Bo~Li, Peiyuan Zhang, Fanyi Pu, Joshua~Adrian Cahyono, Kairui Hu, Shuai Liu, Yuanhan Zhang, Jingkang Yang, Chunyuan Li, and Ziwei Liu.
\newblock Lmms-eval: Reality check on the evaluation of large multimodal models, 2024{\natexlab{a}}.
\newblock URL \url{https://arxiv.org/abs/2407.12772}.

\bibitem[Yan et~al.(2023)Yan, Jiang, Wu, Wang, Luo, Yuan, and Lu]{yan2023universal}
Bin Yan, Yi~Jiang, Jiannan Wu, Dong Wang, Ping Luo, Zehuan Yuan, and Huchuan Lu.
\newblock Universal instance perception as object discovery and retrieval.
\newblock In \emph{Proceedings of the IEEE/CVF Conference on Computer Vision and Pattern Recognition}, pages 15325--15336, 2023.

\bibitem[Liu et~al.(2024{\natexlab{a}})Liu, Zeng, Ren, Li, Zhang, Yang, Jiang, Li, Yang, Su, et~al.]{liu2024grounding}
Shilong Liu, Zhaoyang Zeng, Tianhe Ren, Feng Li, Hao Zhang, Jie Yang, Qing Jiang, Chunyuan Li, Jianwei Yang, Hang Su, et~al.
\newblock Grounding dino: Marrying dino with grounded pre-training for open-set object detection.
\newblock In \emph{European conference on computer vision}, pages 38--55. Springer, 2024{\natexlab{a}}.

\bibitem[Wang et~al.(2023{\natexlab{a}})Wang, Wang, Lin, Bai, Zhou, Zhou, Wang, and Zhou]{wang2023one}
Peng Wang, Shijie Wang, Junyang Lin, Shuai Bai, Xiaohuan Zhou, Jingren Zhou, Xinggang Wang, and Chang Zhou.
\newblock One-peace: Exploring one general representation model toward unlimited modalities.
\newblock \emph{arXiv preprint arXiv:2305.11172}, 2023{\natexlab{a}}.

\bibitem[Wang et~al.(2022)Wang, Bao, Dong, Bjorck, Peng, Liu, Aggarwal, Mohammed, Singhal, Som, et~al.]{beit3}
Wenhui Wang, Hangbo Bao, Li~Dong, Johan Bjorck, Zhiliang Peng, Qiang Liu, Kriti Aggarwal, Owais~Khan Mohammed, Saksham Singhal, Subhojit Som, et~al.
\newblock Image as a foreign language: Beit pretraining for all vision and vision-language tasks.
\newblock \emph{arXiv:2208.10442}, 2022.

\bibitem[Zhang et~al.()Zhang, Li, Liu, Zhang, Su, Zhu, Ni, and Shum]{zhangdino}
Hao Zhang, Feng Li, Shilong Liu, Lei Zhang, Hang Su, Jun Zhu, Lionel Ni, and Heung-Yeung Shum.
\newblock Dino: Detr with improved denoising anchor boxes for end-to-end object detection.
\newblock In \emph{The Eleventh International Conference on Learning Representations}.

\bibitem[Qiu et~al.(2025)Qiu, Wu, Ke, Bai, and Zhang]{qiu2025refining}
Congpei Qiu, Yanhao Wu, Wei Ke, Xiuxiu Bai, and Tong Zhang.
\newblock Refining {CLIP}'s spatial awareness: A visual-centric perspective.
\newblock In \emph{The Thirteenth International Conference on Learning Representations}, 2025.
\newblock URL \url{https://openreview.net/forum?id=38No4B8sx6}.

\bibitem[Cheng et~al.(2022{\natexlab{a}})Cheng, Misra, Schwing, Kirillov, and Girdhar]{cheng2022masked}
Bowen Cheng, Ishan Misra, Alexander~G Schwing, Alexander Kirillov, and Rohit Girdhar.
\newblock Masked-attention mask transformer for universal image segmentation.
\newblock In \emph{Proceedings of the IEEE/CVF conference on computer vision and pattern recognition}, pages 1290--1299, 2022{\natexlab{a}}.

\bibitem[Lan et~al.(2024)Lan, Chen, Ke, Wang, Feng, and Zhang]{lan2024proxyclip}
Mengcheng Lan, Chaofeng Chen, Yiping Ke, Xinjiang Wang, Litong Feng, and Wayne Zhang.
\newblock Proxyclip: Proxy attention improves clip for open-vocabulary segmentation.
\newblock In \emph{European Conference on Computer Vision}, pages 70--88. Springer, 2024.

\bibitem[Carion et~al.(2025)Carion, Gustafson, Hu, Debnath, Hu, Suris, Ryali, Alwala, Khedr, Huang, et~al.]{carion2025sam}
Nicolas Carion, Laura Gustafson, Yuan-Ting Hu, Shoubhik Debnath, Ronghang Hu, Didac Suris, Chaitanya Ryali, Kalyan~Vasudev Alwala, Haitham Khedr, Andrew Huang, et~al.
\newblock Sam 3: Segment anything with concepts.
\newblock \emph{arXiv preprint arXiv:2511.16719}, 2025.

\bibitem[Liu et~al.(2024{\natexlab{b}})Liu, Bai, Li, Wang, and Tang]{liu2024open}
Yong Liu, Sule Bai, Guanbin Li, Yitong Wang, and Yansong Tang.
\newblock Open-vocabulary segmentation with semantic-assisted calibration.
\newblock In \emph{Proceedings of the IEEE/CVF Conference on Computer Vision and Pattern Recognition}, pages 3491--3500, 2024{\natexlab{b}}.

\bibitem[Zhou et~al.(2022)Zhou, Loy, and Dai]{zhou2022extract}
Chong Zhou, Chen~Change Loy, and Bo~Dai.
\newblock Extract free dense labels from clip.
\newblock In \emph{European conference on computer vision}, pages 696--712. Springer, 2022.

\bibitem[Xu et~al.(2023)Xu, Zhang, Wei, Hu, and Bai]{xu2023side}
Mengde Xu, Zheng Zhang, Fangyun Wei, Han Hu, and Xiang Bai.
\newblock Side adapter network for open-vocabulary semantic segmentation.
\newblock In \emph{Proceedings of the IEEE/CVF Conference on Computer Vision and Pattern Recognition}, pages 2945--2954, 2023.

\bibitem[Ouyang et~al.(2025)Ouyang, Niu, Wang, Chen, and Lin]{ouyang2025region}
Shuyi Ouyang, Ziwei Niu, Hongyi Wang, Yen-Wei Chen, and Lanfen Lin.
\newblock Region-aware anchoring mechanism for efficient referring visual grounding.
\newblock In \emph{Proceedings of the IEEE/CVF International Conference on Computer Vision}, pages 24192--24202, 2025.

\bibitem[Cai et~al.(2025)Cai, Yeh, Xu, Liu, Meyer, Lei, Zhao, Li, Chandra, and Shi]{cai2025depthlm}
Zhipeng Cai, Ching-Feng Yeh, Hu~Xu, Zhuang Liu, Gregory Meyer, Xinjie Lei, Changsheng Zhao, Shang-Wen Li, Vikas Chandra, and Yangyang Shi.
\newblock Depthlm: Metric depth from vision language models.
\newblock \emph{arXiv preprint arXiv:2509.25413}, 2025.

\bibitem[Piccinelli et~al.(2025)Piccinelli, Sakaridis, Yang, Segu, Li, Abbeloos, and Van~Gool]{piccinelli2025unidepthv2}
Luigi Piccinelli, Christos Sakaridis, Yung-Hsu Yang, Mattia Segu, Siyuan Li, Wim Abbeloos, and Luc Van~Gool.
\newblock Unidepthv2: Universal monocular metric depth estimation made simpler.
\newblock \emph{arXiv preprint arXiv:2502.20110}, 2025.

\bibitem[Taghavi et~al.(2024)Taghavi, Langari, and Pandey]{taghavi2024swinmtl}
Pardis Taghavi, Reza Langari, and Gaurav Pandey.
\newblock Swinmtl: A shared architecture for simultaneous depth estimation and semantic segmentation from monocular camera images.
\newblock In \emph{2024 IEEE/RSJ International Conference on Intelligent Robots and Systems (IROS)}, pages 4957--4964. IEEE, 2024.

\bibitem[Wang et~al.(2024{\natexlab{a}})Wang, Xuan, and Zhang]{wang2024locllm}
Dongkai Wang, Shiyu Xuan, and Shiliang Zhang.
\newblock Locllm: Exploiting generalizable human keypoint localization via large language model.
\newblock In \emph{Proceedings of the IEEE/CVF conference on computer vision and pattern recognition}, pages 614--623, 2024{\natexlab{a}}.

\bibitem[Xu et~al.(2022)Xu, Zhang, Zhang, and Tao]{xu2022vitpose}
Yufei Xu, Jing Zhang, Qiming Zhang, and Dacheng Tao.
\newblock Vitpose: Simple vision transformer baselines for human pose estimation.
\newblock \emph{Advances in neural information processing systems}, 35:\penalty0 38571--38584, 2022.

\bibitem[Fang et~al.(2024)Fang, Sun, Wang, Huang, Wang, and Cao]{fang2024eva}
Yuxin Fang, Quan Sun, Xinggang Wang, Tiejun Huang, Xinlong Wang, and Yue Cao.
\newblock Eva-02: A visual representation for neon genesis.
\newblock \emph{Image and Vision Computing}, 149:\penalty0 105171, 2024.

\bibitem[Wortsman et~al.(2022)Wortsman, Ilharco, Gadre, Roelofs, Gontijo-Lopes, Morcos, Namkoong, Farhadi, Carmon, Kornblith, et~al.]{wortsman2022model}
Mitchell Wortsman, Gabriel Ilharco, Samir~Ya Gadre, Rebecca Roelofs, Raphael Gontijo-Lopes, Ari~S Morcos, Hongseok Namkoong, Ali Farhadi, Yair Carmon, Simon Kornblith, et~al.
\newblock Model soups: averaging weights of multiple fine-tuned models improves accuracy without increasing inference time.
\newblock In \emph{International conference on machine learning}, pages 23965--23998. PMLR, 2022.

\bibitem[Beyer et~al.(2024)Beyer, Steiner, Pinto, Kolesnikov, Wang, Salz, Neumann, Alabdulmohsin, Tschannen, Bugliarello, et~al.]{beyer2024paligemma}
Lucas Beyer, Andreas Steiner, Andr{\'e}~Susano Pinto, Alexander Kolesnikov, Xiao Wang, Daniel Salz, Maxim Neumann, Ibrahim Alabdulmohsin, Michael Tschannen, Emanuele Bugliarello, et~al.
\newblock Paligemma: A versatile 3b vlm for transfer.
\newblock \emph{arXiv preprint arXiv:2407.07726}, 2024.

\bibitem[Wu et~al.(2024{\natexlab{b}})Wu, Hu, Wang, Pang, and Soricut]{wu2024omni}
Jialin Wu, Xia Hu, Yaqing Wang, Bo~Pang, and Radu Soricut.
\newblock Omni-smola: Boosting generalist multimodal models with soft mixture of low-rank experts.
\newblock In \emph{Proceedings of the IEEE/CVF Conference on Computer Vision and Pattern Recognition}, pages 14205--14215, 2024{\natexlab{b}}.

\bibitem[Liu et~al.(2023{\natexlab{b}})Liu, Zeng, Ren, Li, Zhang, Yang, yue Li, Yang, Su, Zhu, and Zhang]{grounding_dino}
Shilong Liu, Zhaoyang Zeng, Tianhe Ren, Feng Li, Hao Zhang, Jie Yang, Chun yue Li, Jianwei Yang, Hang Su, Jun-Juan Zhu, and Lei Zhang.
\newblock Grounding dino: Marrying dino with grounded pre-training for open-set object detection.
\newblock \emph{arXiv:2303.05499}, 2023{\natexlab{b}}.

\bibitem[Yu et~al.(2016)Yu, Poirson, Yang, Berg, and Berg]{yu2016modeling}
Licheng Yu, Patrick Poirson, Shan Yang, Alexander~C Berg, and Tamara~L Berg.
\newblock Modeling context in referring expressions.
\newblock In \emph{European conference on computer vision}, pages 69--85. Springer, 2016.

\bibitem[Guo et~al.(2025)Guo, Wu, Zhu, Leng, Shi, Chen, Fan, Wang, Jiang, Wang, et~al.]{guo2025seed1}
Dong Guo, Faming Wu, Feida Zhu, Fuxing Leng, Guang Shi, Haobin Chen, Haoqi Fan, Jian Wang, Jianyu Jiang, Jiawei Wang, et~al.
\newblock Seed1. 5-vl technical report.
\newblock \emph{arXiv preprint arXiv:2505.07062}, 2025.

\bibitem[Lin et~al.(2014)Lin, Maire, Belongie, Hays, Perona, Ramanan, Doll{\'a}r, and Zitnick]{lin2014microsoft}
Tsung-Yi Lin, Michael Maire, Serge Belongie, James Hays, Pietro Perona, Deva Ramanan, Piotr Doll{\'a}r, and C~Lawrence Zitnick.
\newblock Microsoft coco: Common objects in context.
\newblock In \emph{European conference on computer vision}, pages 740--755. Springer, 2014.

\bibitem[Wang et~al.(2024{\natexlab{b}})Wang, Tang, Jiang, Shi, Naeem, Li, Schiele, and Wang]{wang2024git}
Haiyang Wang, Hao Tang, Li~Jiang, Shaoshuai Shi, Muhammad~Ferjad Naeem, Hongsheng Li, Bernt Schiele, and Liwei Wang.
\newblock Git: Towards generalist vision transformer through universal language interface.
\newblock In \emph{European Conference on Computer Vision}, pages 55--73. Springer, 2024{\natexlab{b}}.

\bibitem[Zhou et~al.(2017)Zhou, Zhao, Puig, Fidler, Barriuso, and Torralba]{zhou2017scene}
Bolei Zhou, Hang Zhao, Xavier Puig, Sanja Fidler, Adela Barriuso, and Antonio Torralba.
\newblock Scene parsing through ade20k dataset.
\newblock In \emph{Proceedings of the IEEE conference on computer vision and pattern recognition}, pages 633--641, 2017.

\bibitem[Caesar et~al.(2018)Caesar, Uijlings, and Ferrari]{caesar2018coco}
Holger Caesar, Jasper Uijlings, and Vittorio Ferrari.
\newblock Coco-stuff: Thing and stuff classes in context.
\newblock In \emph{Proceedings of the IEEE conference on computer vision and pattern recognition}, pages 1209--1218, 2018.

\bibitem[Silberman et~al.(2012)Silberman, Hoiem, Kohli, and Fergus]{silberman2012indoor}
Nathan Silberman, Derek Hoiem, Pushmeet Kohli, and Rob Fergus.
\newblock Indoor segmentation and support inference from rgbd images.
\newblock In \emph{European conference on computer vision}, pages 746--760. Springer, 2012.

\bibitem[Guizilini et~al.(2020)Guizilini, Ambrus, Pillai, Raventos, and Gaidon]{guizilini20203d}
Vitor Guizilini, Rares Ambrus, Sudeep Pillai, Allan Raventos, and Adrien Gaidon.
\newblock 3d packing for self-supervised monocular depth estimation.
\newblock In \emph{Proceedings of the IEEE/CVF conference on computer vision and pattern recognition}, pages 2485--2494, 2020.

\bibitem[Andriluka et~al.(2014)Andriluka, Pishchulin, Gehler, and Schiele]{andriluka20142d}
Mykhaylo Andriluka, Leonid Pishchulin, Peter Gehler, and Bernt Schiele.
\newblock 2d human pose estimation: New benchmark and state of the art analysis.
\newblock In \emph{Proceedings of the IEEE Conference on computer Vision and Pattern Recognition}, pages 3686--3693, 2014.

\bibitem[Beyer et~al.(2020)Beyer, H{\'e}naff, Kolesnikov, Zhai, and Oord]{beyer2020we}
Lucas Beyer, Olivier~J H{\'e}naff, Alexander Kolesnikov, Xiaohua Zhai, and A{\"a}ron van~den Oord.
\newblock Are we done with imagenet?
\newblock \emph{arXiv preprint arXiv:2006.07159}, 2020.

\bibitem[Paiss et~al.(2023)Paiss, Ephrat, Tov, Zada, Mosseri, Irani, and Dekel]{paiss2023teaching}
Roni Paiss, Ariel Ephrat, Omer Tov, Shiran Zada, Inbar Mosseri, Michal Irani, and Tali Dekel.
\newblock Teaching clip to count to ten.
\newblock In \emph{Proceedings of the IEEE/CVF International Conference on Computer Vision}, pages 3170--3180, 2023.

\bibitem[Acharya et~al.(2019)Acharya, Kafle, and Kanan]{acharya2019tallyqa}
Manoj Acharya, Kushal Kafle, and Christopher Kanan.
\newblock Tallyqa: Answering complex counting questions.
\newblock In \emph{Proceedings of the AAAI conference on artificial intelligence}, volume~33, pages 8076--8084, 2019.

\bibitem[Liu et~al.(2023{\natexlab{c}})Liu, Duan, Zhang, Li, Zhang, Zhao, Yuan, Wang, He, Liu, et~al.]{liu2023mmbench}
Yuan Liu, Haodong Duan, Yuanhan Zhang, Bo~Li, Songyang Zhang, Wangbo Zhao, Yike Yuan, Jiaqi Wang, Conghui He, Ziwei Liu, et~al.
\newblock Mmbench: Is your multi-modal model an all-around player?
\newblock \emph{arXiv preprint arXiv:2307.06281}, 2023{\natexlab{c}}.

\bibitem[Chen et~al.(2024)Chen, Li, Dong, Zhang, Zang, Chen, Duan, Wang, Qiao, Lin, et~al.]{chen2024we}
Lin Chen, Jinsong Li, Xiaoyi Dong, Pan Zhang, Yuhang Zang, Zehui Chen, Haodong Duan, Jiaqi Wang, Yu~Qiao, Dahua Lin, et~al.
\newblock Are we on the right way for evaluating large vision-language models?
\newblock \emph{arXiv:2403.20330}, 2024.

\bibitem[Fu et~al.(2023)Fu, Chen, Shen, Qin, Zhang, Lin, Qiu, Lin, Yang, Zheng, et~al.]{fu2023mme}
Chaoyou Fu, Peixian Chen, Yunhang Shen, Yulei Qin, Mengdan Zhang, Xu~Lin, Zhenyu Qiu, Wei Lin, Jinrui Yang, Xiawu Zheng, et~al.
\newblock Mme: A comprehensive evaluation benchmark for multimodal large language models.
\newblock \emph{arXiv:2306.13394}, 2023.

\bibitem[Zhu et~al.(2025)Zhu, Dong, Wang, Li, Deng, Wang, Hong, Geng, Niu, Huang, et~al.]{zhu2025cvbench}
Nannan Zhu, Yonghao Dong, Teng Wang, Xueqian Li, Shengjun Deng, Yijia Wang, Zheng Hong, Tiantian Geng, Guo Niu, Hanyan Huang, et~al.
\newblock Cvbench: Benchmarking cross-video synergies for complex multimodal reasoning.
\newblock \emph{arXiv preprint arXiv:2508.19542}, 2025.

\bibitem[Lu et~al.(2022)Lu, Mishra, Xia, Qiu, Chang, Zhu, Tafjord, Clark, and Kalyan]{lu2022learn_scienceqa}
Pan Lu, Swaroop Mishra, Tanglin Xia, Liang Qiu, Kai-Wei Chang, Song-Chun Zhu, Oyvind Tafjord, Peter Clark, and Ashwin Kalyan.
\newblock Learn to explain: Multimodal reasoning via thought chains for science question answering.
\newblock In \emph{NeurIPS}, 2022.

\bibitem[Li et~al.(2023{\natexlab{a}})Li, Wang, Wang, Ge, Ge, and Shan]{li2023seedbench}
Bohao Li, Rui Wang, Guangzhi Wang, Yuying Ge, Yixiao Ge, and Ying Shan.
\newblock Seed-bench: Benchmarking multimodal llms with generative comprehension.
\newblock \emph{arXiv:2307.16125}, 2023{\natexlab{a}}.

\bibitem[Li et~al.(2024{\natexlab{b}})Li, Ge, Ge, Wang, Wang, Zhang, and Shan]{li2024seed}
Bohao Li, Yuying Ge, Yixiao Ge, Guangzhi Wang, Rui Wang, Ruimao Zhang, and Ying Shan.
\newblock Seed-bench: Benchmarking multimodal large language models.
\newblock In \emph{Proceedings of the IEEE/CVF Conference on Computer Vision and Pattern Recognition}, pages 13299--13308, 2024{\natexlab{b}}.

\bibitem[Li et~al.(2024{\natexlab{c}})Li, Ge, Chen, Ge, Zhang, and Shan]{li2024seed2plus}
Bohao Li, Yuying Ge, Yi~Chen, Yixiao Ge, Ruimao Zhang, and Ying Shan.
\newblock Seed-bench-2-plus: Benchmarking multimodal large language models with text-rich visual comprehension.
\newblock \emph{arXiv preprint arXiv:2404.16790}, 2024{\natexlab{c}}.

\bibitem[Yu et~al.(2024)Yu, Yang, Li, Wang, Lin, Liu, Wang, and Wang]{yu2024mm}
Weihao Yu, Zhengyuan Yang, Linjie Li, Jianfeng Wang, Kevin Lin, Zicheng Liu, Xinchao Wang, and Lijuan Wang.
\newblock Mm-vet: Evaluating large multimodal models for integrated capabilities.
\newblock In \emph{ICML}, 2024.

\bibitem[Rahmanzadehgervi et~al.(2025)Rahmanzadehgervi, Bolton, Taesiri, and Nguyen]{rahmanzadehgervi2025visionlanguagemodelsblind}
Pooyan Rahmanzadehgervi, Logan Bolton, Mohammad~Reza Taesiri, and Anh~Totti Nguyen.
\newblock Vision language models are blind: Failing to translate detailed visual features into words, 2025.
\newblock URL \url{https://arxiv.org/abs/2407.06581}.

\bibitem[Xu et~al.(2025)Xu, Wang, Wang, Chen, Zhou, Yang, Lu, Li, Wang, Zhu, et~al.]{xu2025visulogicbenchmarkevaluatingvisual}
Weiye Xu, Jiahao Wang, Weiyun Wang, Zhe Chen, Wengang Zhou, Aijun Yang, Lewei Lu, Houqiang Li, Xiaohua Wang, Xizhou Zhu, et~al.
\newblock Visulogic: A benchmark for evaluating visual reasoning in multi-modal large language models, 2025.
\newblock URL \url{https://arxiv.org/abs/2504.15279}.

\bibitem[Yue et~al.(2024{\natexlab{a}})Yue, Ni, Zhang, Zheng, Liu, Zhang, et~al.]{yue2024mmmu}
Xiang Yue, Yuansheng Ni, Kai Zhang, Tianyu Zheng, Ruoqi Liu, Ge~Zhang, et~al.
\newblock Mmmu: A massive multi-discipline multimodal understanding and reasoning benchmark for expert agi.
\newblock In \emph{Proceedings of the IEEE/CVF Conference on Computer Vision and Pattern Recognition}, pages 9556--9567, 2024{\natexlab{a}}.

\bibitem[Yue et~al.(2024{\natexlab{b}})Yue, Zheng, Ni, Wang, Zhang, Tong, Sun, Yin, Yu, Zhang, et~al.]{mmmupro}
Xiang Yue, Tianyu Zheng, Yuansheng Ni, Yubo Wang, Kai Zhang, Shengbang Tong, Yuxuan Sun, Ming Yin, Botao Yu, Ge~Zhang, et~al.
\newblock Mmmu-pro: A more robust multi-discipline multimodal understanding benchmark.
\newblock \emph{arXiv preprint arXiv:2409.02813}, 2024{\natexlab{b}}.

\bibitem[Zhang et~al.(2024{\natexlab{b}})Zhang, Du, Chen, Liang, Luo, Zheng, Zhu, Cheng, Xu, Guo, et~al.]{zhang2024cmmmu}
Ge~Zhang, Xinrun Du, Bei Chen, Yiming Liang, Tongxu Luo, Tianyu Zheng, Kang Zhu, Yuyang Cheng, Chunpu Xu, Shuyue Guo, et~al.
\newblock Cmmmu: A chinese massive multi-discipline multimodal understanding benchmark.
\newblock \emph{arXiv preprint arXiv:2401.11944}, 2024{\natexlab{b}}.

\bibitem[Lu et~al.(2024)Lu, Bansal, Xia, Liu, Li, Hajishirzi, Cheng, Chang, Galley, and Gao]{mathvista}
Pan Lu, Hritik Bansal, Tony Xia, Jiacheng Liu, Chunyuan Li, Hannaneh Hajishirzi, Hao Cheng, Kai{-}Wei Chang, Michel Galley, and Jianfeng Gao.
\newblock Mathvista: Evaluating mathematical reasoning of foundation models in visual contexts.
\newblock In \emph{ICLR}, 2024.

\bibitem[Zhang et~al.(2024{\natexlab{c}})Zhang, Jiang, Zhang, Lin, Guo, Qiu, Zhou, Lu, Chang, Qiao, et~al.]{zhang2024mathverse}
Renrui Zhang, Dongzhi Jiang, Yichi Zhang, Haokun Lin, Ziyu Guo, Pengshuo Qiu, Aojun Zhou, Pan Lu, Kai-Wei Chang, Yu~Qiao, et~al.
\newblock Mathverse: Does your multi-modal llm truly see the diagrams in visual math problems?
\newblock In \emph{European Conference on Computer Vision}, pages 169--186. Springer, 2024{\natexlab{c}}.

\bibitem[Xiao et~al.(2024{\natexlab{a}})Xiao, Sun, Liu, and Wang]{xiao2024logicvista}
Yijia Xiao, Edward Sun, Tianyu Liu, and Wei Wang.
\newblock Logicvista: Multimodal llm logical reasoning benchmark in visual contexts.
\newblock \emph{arXiv preprint arXiv:2407.04973}, 2024{\natexlab{a}}.

\bibitem[Guan et~al.(2023)Guan, Liu, Wu, Xian, Li, Liu, Wang, Chen, Huang, Yacoob, Manocha, and Zhou]{hallusion_bench}
Tianrui Guan, Fuxiao Liu, Xiyang Wu, Ruiqi Xian, Zongxia Li, Xiaoyu Liu, Xijun Wang, Lichang Chen, Furong Huang, Yaser Yacoob, Dinesh Manocha, and Tianyi Zhou.
\newblock Hallusionbench: An advanced diagnostic suite for entangled language hallucination \& visual illusion in large vision-language models, 2023.

\bibitem[Li et~al.(2023{\natexlab{b}})Li, Du, Zhou, Wang, Zhao, and Wen]{li2023evaluatingpope}
Yifan Li, Yifan Du, Kun Zhou, Jinpeng Wang, Wayne~Xin Zhao, and Ji-Rong Wen.
\newblock Evaluating object hallucination in large vision-language models.
\newblock \emph{arXiv preprint arXiv:2305.10355}, 2023{\natexlab{b}}.

\bibitem[Wang et~al.(2023{\natexlab{b}})Wang, Shi, Li, Wang, Huang, Xing, Chen, Li, Zhu, Cao, et~al.]{wang2023allseeing}
Weiyun Wang, Min Shi, Qingyun Li, Wenhai Wang, Zhenhang Huang, Linjie Xing, Zhe Chen, Hao Li, Xizhou Zhu, Zhiguo Cao, et~al.
\newblock The all-seeing project: Towards panoptic visual recognition and understanding of the open world.
\newblock \emph{arXiv preprint arXiv:2308.01907}, 2023{\natexlab{b}}.

\bibitem[Wang et~al.(2024{\natexlab{c}})Wang, Ren, Luo, Li, Yan, Chen, Wang, Li, Lu, Zhu, et~al.]{wang2024allseeing_v2}
Weiyun Wang, Yiming Ren, Haowen Luo, Tiantong Li, Chenxiang Yan, Zhe Chen, Wenhai Wang, Qingyun Li, Lewei Lu, Xizhou Zhu, et~al.
\newblock The all-seeing project v2: Towards general relation comprehension of the open world.
\newblock \emph{arXiv preprint arXiv:2402.19474}, 2024{\natexlab{c}}.

\bibitem[Kembhavi et~al.(2016)Kembhavi, Salvato, Kolve, Seo, Hajishirzi, and Farhadi]{Kembhavi2016AI2D}
Aniruddha Kembhavi, Michael Salvato, Eric Kolve, Minjoon Seo, Hannaneh Hajishirzi, and Ali Farhadi.
\newblock A diagram is worth a dozen images.
\newblock \emph{ArXiv}, abs/1603.07396, 2016.

\bibitem[Mathew et~al.(2021{\natexlab{a}})Mathew, Bagal, Tito, Karatzas, Valveny, and Jawahar]{Mathew2021InfographicVQA}
Minesh Mathew, Viraj Bagal, Rub{\`e}n~P{\'e}rez Tito, Dimosthenis Karatzas, Ernest Valveny, and C.V. Jawahar.
\newblock Infographicvqa.
\newblock \emph{2022 IEEE/CVF Winter Conference on Applications of Computer Vision (WACV)}, pages 2582--2591, 2021{\natexlab{a}}.

\bibitem[Singh et~al.(2019)Singh, Natarajan, Shah, Jiang, Chen, Batra, Parikh, and Rohrbach]{textvqa}
Amanpreet Singh, Vivek Natarajan, Meet Shah, Yu~Jiang, Xinlei Chen, Dhruv Batra, Devi Parikh, and Marcus Rohrbach.
\newblock Towards vqa models that can read.
\newblock In \emph{CVPR}, 2019.

\bibitem[Mathew et~al.(2021{\natexlab{b}})Mathew, Karatzas, and Jawahar]{docvqa}
Minesh Mathew, Dimosthenis Karatzas, and CV~Jawahar.
\newblock Docvqa: A dataset for vqa on document images.
\newblock In \emph{WACV}, 2021{\natexlab{b}}.

\bibitem[Masry et~al.(2022)Masry, Long, Tan, Joty, and Hoque]{masry2022chartqa}
Ahmed Masry, Do~Xuan Long, Jia~Qing Tan, Shafiq Joty, and Enamul Hoque.
\newblock Chartqa: A benchmark for question answering about charts with visual and logical reasoning.
\newblock \emph{arXiv:2203.10244}, 2022.

\bibitem[Liu et~al.(2023{\natexlab{d}})Liu, Li, Huang, Yang, Yu, Li, Yin, lin Liu, Jin, and Bai]{liu2024ocrbenchhiddenmysteryocr}
Yuliang Liu, Zhang Li, Mingxin Huang, Biao Yang, Wenwen Yu, Chunyuan Li, Xucheng Yin, Cheng lin Liu, Lianwen Jin, and Xiang Bai.
\newblock Ocrbench: On the hidden mystery of ocr in large multimodal models.
\newblock \emph{arXiv:2305.07895}, 2023{\natexlab{d}}.

\bibitem[Wang et~al.(2024{\natexlab{d}})Wang, Xia, He, Chen, Liu, Zhu, Liang, Wu, Liu, Malladi, Chevalier, Arora, and Chen]{wang2024charxiv}
Zirui Wang, Mengzhou Xia, Luxi He, Howard Chen, Yitao Liu, Richard Zhu, Kaiqu Liang, Xindi Wu, Haotian Liu, Sadhika Malladi, Alexis Chevalier, Sanjeev Arora, and Danqi Chen.
\newblock Charxiv: Charting gaps in realistic chart understanding in multimodal llms.
\newblock \emph{arXiv preprint arXiv:2406.18521}, 2024{\natexlab{d}}.

\bibitem[Fu et~al.(2024)Fu, Hu, Li, Feng, Wang, Lin, Roth, Smith, Ma, and Krishna]{fu2024blink}
Xingyu Fu, Yushi Hu, Bangzheng Li, Yu~Feng, Haoyu Wang, Xudong Lin, Dan Roth, Noah~A Smith, Wei-Chiu Ma, and Ranjay Krishna.
\newblock Blink: Multimodal large language models can see but not perceive.
\newblock In \emph{European Conference on Computer Vision}, pages 148--166. Springer, 2024.

\bibitem[{xAI}(2024)]{realworldqa2024}
{xAI}.
\newblock Realworldqa: A benchmark for real-world spatial understanding.
\newblock \url{https://huggingface.co/datasets/xai-org/RealworldQA}, 2024.
\newblock Accessed: 2025-04-26.

\bibitem[Zhang et~al.(2024{\natexlab{d}})Zhang, Zhang, Tian, Fu, Zhang, Wu, Li, Wang, Wen, Zhang, et~al.]{mme-realworld}
Yi-Fan Zhang, Huanyu Zhang, Haochen Tian, Chaoyou Fu, Shuangqing Zhang, Junfei Wu, Feng Li, Kun Wang, Qingsong Wen, Zhang Zhang, et~al.
\newblock Mme-realworld: Could your multimodal llm challenge high-resolution real-world scenarios that are difficult for humans?
\newblock \emph{arXiv preprint arXiv:2408.13257}, 2024{\natexlab{d}}.

\bibitem[Li et~al.(2025{\natexlab{b}})Li, Meng, Lin, Luo, Tian, Ma, Huang, and Chua]{li2025screenspot}
Kaixin Li, Ziyang Meng, Hongzhan Lin, Ziyang Luo, Yuchen Tian, Jing Ma, Zhiyong Huang, and Tat-Seng Chua.
\newblock Screenspot-pro: Gui grounding for professional high-resolution computer use.
\newblock In \emph{Proceedings of the 33rd ACM International Conference on Multimedia}, pages 8778--8786, 2025{\natexlab{b}}.

\bibitem[Zhou et~al.(2023)Zhou, Lu, Mishra, Brahma, Basu, Luan, Zhou, and Hou]{ifeval}
Jeffrey Zhou, Tianjian Lu, Swaroop Mishra, Siddhartha Brahma, Sujoy Basu, Yi~Luan, Denny Zhou, and Le~Hou.
\newblock Instruction-following evaluation for large language models.
\newblock \emph{CoRR}, abs/2311.07911, 2023.

\bibitem[Gema et~al.(2024)Gema, Leang, Hong, Devoto, Mancino, Saxena, He, Zhao, Du, Madani, et~al.]{mmluredux}
Aryo~Pradipta Gema, Joshua Ong~Jun Leang, Giwon Hong, Alessio Devoto, Alberto Carlo~Maria Mancino, Rohit Saxena, Xuanli He, Yu~Zhao, Xiaotang Du, Mohammad Reza~Ghasemi Madani, et~al.
\newblock Are we done with mmlu?
\newblock \emph{CoRR}, abs/2406.04127, 2024.

\bibitem[Rein et~al.(2023)Rein, Hou, Stickland, Petty, Pang, Dirani, Michael, and Bowman]{gpqa}
David Rein, Betty~Li Hou, Asa~Cooper Stickland, Jackson Petty, Richard~Yuanzhe Pang, Julien Dirani, Julian Michael, and Samuel~R. Bowman.
\newblock {GPQA}: A graduate-level {Google}-proof {Q}{\&}{A} benchmark.
\newblock \emph{CoRR}, abs/2311.12022, 2023.

\bibitem[Cordts et~al.(2016)Cordts, Omran, Ramos, Rehfeld, Enzweiler, Benenson, Franke, Roth, and Schiele]{cordts2016cityscapes}
Marius Cordts, Mohamed Omran, Sebastian Ramos, Timo Rehfeld, Markus Enzweiler, Rodrigo Benenson, Uwe Franke, Stefan Roth, and Bernt Schiele.
\newblock The cityscapes dataset for semantic urban scene understanding.
\newblock In \emph{Proceedings of the IEEE conference on computer vision and pattern recognition}, pages 3213--3223, 2016.

\bibitem[Mottaghi et~al.(2014)Mottaghi, Chen, Liu, Cho, Lee, Fidler, Urtasun, and Yuille]{mottaghi2014role}
Roozbeh Mottaghi, Xianjie Chen, Xiaobai Liu, Nam-Gyu Cho, Seong-Whan Lee, Sanja Fidler, Raquel Urtasun, and Alan Yuille.
\newblock The role of context for object detection and semantic segmentation in the wild.
\newblock In \emph{Proceedings of the IEEE conference on computer vision and pattern recognition}, pages 891--898, 2014.

\bibitem[Liu et~al.(2023{\natexlab{e}})Liu, Duan, Yuanhan~Zhang, Zhang, Zhao, Yuan, Wang, He, Liu, Chen, and Lin]{MMBench}
Yuan Liu, Haodong Duan, Bo~Li Yuanhan~Zhang, Songyang Zhang, Wangbo Zhao, Yike Yuan, Jiaqi Wang, Conghui He, Ziwei Liu, Kai Chen, and Dahua Lin.
\newblock Mmbench: Is your multi-modal model an all-around player?
\newblock \emph{arXiv:2307.06281}, 2023{\natexlab{e}}.

\bibitem[Liu et~al.(2024{\natexlab{c}})Liu, Li, Huang, Yang, Yu, Li, Yin, Liu, Jin, and Bai]{Liu_2024_OCRBench}
Yuliang Liu, Zhang Li, Mingxin Huang, Biao Yang, Wenwen Yu, Chunyuan Li, Xu-Cheng Yin, Cheng-Lin Liu, Lianwen Jin, and Xiang Bai.
\newblock Ocrbench: on the hidden mystery of ocr in large multimodal models.
\newblock \emph{Science China Information Sciences}, 67\penalty0 (12), December 2024{\natexlab{c}}.
\newblock ISSN 1869-1919.
\newblock \doi{10.1007/s11432-024-4235-6}.
\newblock URL \url{http://dx.doi.org/10.1007/s11432-024-4235-6}.

\bibitem[Yue et~al.(2023)Yue, Ni, Zhang, Zheng, Liu, Zhang, et~al.]{yue2023mmmu}
Xiang Yue, Yuansheng Ni, Kai Zhang, Tianyu Zheng, Ruoqi Liu, Ge~Zhang, et~al.
\newblock Mmmu: A massive multi-discipline multimodal understanding and reasoning benchmark for expert agi.
\newblock \emph{arXiv:2311.16502}, 2023.

\bibitem[Xie et~al.(2021)Xie, Wang, Yu, Anandkumar, Alvarez, and Luo]{xie2021segformer}
Enze Xie, Wenhai Wang, Zhiding Yu, Anima Anandkumar, Jose~M Alvarez, and Ping Luo.
\newblock Segformer: Simple and efficient design for semantic segmentation with transformers.
\newblock In \emph{Advances in Neural Information Processing Systems}, volume~34, pages 12077--12090, 2021.

\bibitem[Cheng et~al.(2022{\natexlab{b}})Cheng, Misra, Schwing, Kirillov, and Girdhar]{cheng2021mask2former}
Bowen Cheng, Ishan Misra, Alexander~G. Schwing, Alexander Kirillov, and Rohit Girdhar.
\newblock Masked-attention mask transformer for universal image segmentation.
\newblock 2022{\natexlab{b}}.

\bibitem[Zhao et~al.(2023)Zhao, Rao, Liu, Liu, Zhou, and Lu]{zhao2023unleashing}
Wenliang Zhao, Yongming Rao, Zuyan Liu, Benlin Liu, Jie Zhou, and Jiwen Lu.
\newblock Unleashing text-to-image diffusion models for visual perception.
\newblock \emph{ICCV}, 2023.

\bibitem[Li et~al.(2025{\natexlab{c}})Li, Wang, Duan, Zhang, and Li]{LI2025111409}
Yi~Li, Hualiang Wang, Yiqun Duan, Jiheng Zhang, and Xiaomeng Li.
\newblock A closer look at the explainability of contrastive language-image pre-training.
\newblock \emph{Pattern Recognition}, 162:\penalty0 111409, 2025{\natexlab{c}}.
\newblock ISSN 0031-3203.
\newblock \doi{https://doi.org/10.1016/j.patcog.2025.111409}.
\newblock URL \url{https://www.sciencedirect.com/science/article/pii/S003132032500069X}.

\bibitem[Kim et~al.(2025)Kim, Ju, Han, Yang, and Hwang]{kim2025distilling}
Chanyoung Kim, Dayun Ju, Woojung Han, Ming-Hsuan Yang, and Seong~Jae Hwang.
\newblock Distilling spectral graph for object-context aware open-vocabulary semantic segmentation.
\newblock In \emph{Proceedings of the Computer Vision and Pattern Recognition Conference}, pages 15033--15042, 2025.

\bibitem[Zou et~al.(2023)Zou, Dou, Yang, Gan, Li, Li, Dai, Behl, Wang, Yuan, et~al.]{zou2023generalized}
Xueyan Zou, Zi-Yi Dou, Jianwei Yang, Zhe Gan, Linjie Li, Chunyuan Li, Xiyang Dai, Harkirat Behl, Jianfeng Wang, Lu~Yuan, et~al.
\newblock Generalized decoding for pixel, image, and language.
\newblock In \emph{Proceedings of the IEEE/CVF conference on computer vision and pattern recognition}, pages 15116--15127, 2023.

\bibitem[Mizrahi et~al.(2023)Mizrahi, Bachmann, Kar, Yeo, Gao, Dehghan, and Zamir]{mizrahi20234m}
David Mizrahi, Roman Bachmann, Oguzhan Kar, Teresa Yeo, Mingfei Gao, Afshin Dehghan, and Amir Zamir.
\newblock 4m: Massively multimodal masked modeling.
\newblock \emph{Advances in Neural Information Processing Systems}, 36:\penalty0 58363--58408, 2023.

\bibitem[Wang et~al.(2023{\natexlab{c}})Wang, Bao, Dong, Bjorck, Peng, Liu, Aggarwal, Mohammed, Singhal, Som, et~al.]{wang2023image}
Wenhui Wang, Hangbo Bao, Li~Dong, Johan Bjorck, Zhiliang Peng, Qiang Liu, Kriti Aggarwal, Owais~Khan Mohammed, Saksham Singhal, Subhojit Som, et~al.
\newblock Image as a foreign language: Beit pretraining for vision and vision-language tasks.
\newblock In \emph{Proceedings of the IEEE/CVF Conference on Computer Vision and Pattern Recognition}, pages 19175--19186, 2023{\natexlab{c}}.

\bibitem[Pramanick et~al.(2024)Pramanick, Han, Hou, Nag, Lim, Ballas, Wang, Chellappa, and Almahairi]{pramanick2024jack}
Shraman Pramanick, Guangxing Han, Rui Hou, Sayan Nag, Ser-Nam Lim, Nicolas Ballas, Qifan Wang, Rama Chellappa, and Amjad Almahairi.
\newblock Jack of all tasks master of many: Designing general-purpose coarse-to-fine vision-language model.
\newblock In \emph{Proceedings of the IEEE/CVF Conference on Computer Vision and Pattern Recognition}, pages 14076--14088, 2024.

\bibitem[Zhang et~al.(2025)Zhang, Zhong, and Han]{zhang2025mr}
Chang-Bin Zhang, Yujie Zhong, and Kai Han.
\newblock Mr. detr: Instructive multi-route training for detection transformers.
\newblock In \emph{Proceedings of the Computer Vision and Pattern Recognition Conference}, pages 9933--9943, 2025.

\bibitem[Kamath et~al.(2021)Kamath, Singh, LeCun, Synnaeve, Misra, and Carion]{kamath2021mdetr}
Aishwarya Kamath, Mannat Singh, Yann LeCun, Gabriel Synnaeve, Ishan Misra, and Nicolas Carion.
\newblock Mdetr-modulated detection for end-to-end multi-modal understanding.
\newblock In \emph{Proceedings of the IEEE/CVF international conference on computer vision}, pages 1780--1790, 2021.

\bibitem[Xiao et~al.(2024{\natexlab{b}})Xiao, Wu, Xu, Dai, Hu, Lu, Zeng, Liu, and Yuan]{xiao2024florence}
Bin Xiao, Haiping Wu, Weijian Xu, Xiyang Dai, Houdong Hu, Yumao Lu, Michael Zeng, Ce~Liu, and Lu~Yuan.
\newblock Florence-2: Advancing a unified representation for a variety of vision tasks.
\newblock In \emph{Proceedings of the IEEE/CVF Conference on Computer Vision and Pattern Recognition}, pages 4818--4829, 2024{\natexlab{b}}.

\bibitem[Zhan et~al.(2024)Zhan, Zhu, Chen, Yang, Tang, and Wang]{zhan2024griffon}
Yufei Zhan, Yousong Zhu, Zhiyang Chen, Fan Yang, Ming Tang, and Jinqiao Wang.
\newblock Griffon: Spelling out all object locations at any granularity with large language models.
\newblock In \emph{European Conference on Computer Vision}, pages 405--422. Springer, 2024.

\end{thebibliography}
